\documentclass[pdflatex,sn-basic]{sn-jnl}% Basic Springer Nature Reference Style/Chemistry Reference Style

\usepackage{graphicx}%
\usepackage{multirow}%
\usepackage{amsmath,amssymb,amsfonts}%
\usepackage{amsthm}%
\usepackage{mathrsfs}%
\usepackage[title]{appendix}%
\usepackage{xcolor}%
\usepackage{textcomp}%
\usepackage{manyfoot}%
\usepackage{booktabs}%
\usepackage{algorithm}%
\usepackage{algorithmicx}%
\usepackage{algpseudocode}%
\usepackage{listings}%
\usepackage{adjustbox}
\usepackage{comment}
\usepackage{pifont}
\usepackage{makecell} 
\usepackage{array} %<================= Added this
\usepackage{longtable}

\newcommand{\xmark}{\ding{55}} % X mark

\theoremstyle{thmstyleone}%
\theoremstyle{thmstyletwo}%

\theoremstyle{thmstylethree}%

\begin{document}

\title[Article Title]{A Comprehensive Perspective on Explainable AI across the Machine
Learning Workflow}


\author*[1]{\fnm{George} \sur{Paterakis}}\email{gpaterakis@jadbio.com}

\author[2]{\fnm{Andrea} \sur{Castellani}}\email{andrea.castellani@honda-ri.de}
\equalcont{These authors contributed equally to this work.}

\author[1]{\fnm{George} \sur{Papoutsoglou}}\email{gpap@jadbio.com}
\equalcont{These authors contributed equally to this work.}
 
\author[2]{\fnm{Tobias} \sur{Rodemann}}\email{tobias.rodemann@honda-ri.de}
\equalcont{These authors contributed equally to this work.}

\author[1,3]{\fnm{Ioannis} \sur{Tsamardinos}}\email{tsamard@jadbio.com}
\equalcont{These authors contributed equally to this work.}

\affil*[1]{\orgname{JADBio Gnosis DA S.A.}, \orgaddress{\street{N.Plastira 100}, \city{Heraklion}, \postcode{70013}, \state{Crete}, \country{Greece}}}

\affil[2]{\orgname{Honda Research Institute Europe}, \orgaddress{\street{Carl-Legien-Strasse 30}, \city{Offenbach am Main}, \postcode{63073}, \state{Hessia}, \country{Germany}}}

\affil[3]{\orgdiv{Computer Science}, \orgname{University of Crete}, \orgaddress{\street{Voutes}, \city{Heraklion}, \postcode{70013}, \state{Crete}, \country{Greece}}}
\abstract{
%  150 to 250 words
Artificial intelligence is reshaping science and industry, yet many users still regard its models as opaque “black boxes”. 
Conventional explainable artificial-intelligence methods clarify individual predictions but overlook the upstream decisions and downstream quality checks that determine whether insights can be trusted. 
In this work, we present Holistic Explainable Artificial Intelligence (HXAI), a user-centric framework that embeds explanation into every stage of the data-analysis workflow and tailors those explanations to users.
HXAI unifies six components (data, analysis set-up, learning process, model output, model quality, communication channel) into a single taxonomy and aligns each component with the needs of domain experts, data analysts and data scientists.
A 112-item question bank covers these needs; our survey of contemporary tools highlights critical coverage gaps. 
Grounded in theories of human explanation, principles from human–computer interaction and findings from empirical user studies, HXAI identifies the characteristics that make explanations clear, actionable and cognitively manageable. 
A comprehensive taxonomy operationalises these insights, reducing terminological ambiguity and enabling rigorous coverage analysis of existing toolchains. 
We further demonstrate how AI agents that embed large-language models can orchestrate diverse explanation techniques, translating technical artifacts into stakeholder-specific narratives that bridge the gap between AI developers and domain experts.
Departing from traditional surveys or perspective articles, this work melds concepts from multiple disciplines, lessons from real-world projects and a critical synthesis of the literature to advance a novel, end-to-end viewpoint on transparency, trustworthiness and responsible AI deployment.
}



\maketitle
\section{Introduction}\label{sec1}

Artificial intelligence (AI) is rapidly transforming a wide range of industries, including healthcare, finance, manufacturing, and others \citep{Secinaro2021, bahoo2024artificial, gao2024artificial, lai2024large}. Despite these advances, many state-of-the-art AI systems remain highly complex and difficult to interpret, often viewed as  “black-boxes”, even by data experts \citep{Mythos-Lipton}. This lack of transparency undermines trust and poses a significant barrier to broader adoption, particularly in high-stakes or regulated environments where explainability and reliability is essential \citep{caruana, 10.1093/haschl/qxae123}.

Efforts to foster trust have received increasing attention in recent years. A central focus of these efforts has been the development of explainable AI (XAI) techniques, which aim to provide greater insight into model behavior and the underlying decision-making processes \citep{AIPlanner,ABUSITTA2024124710, Darpa_Gunning}. While XAI techniques have made progress to enhance transparency and understanding, there is still no standard way to define or evaluate explainability. As a result, most methods focus primarily on explaining model outputs, overlooking the broader data analysis workflow thus, limiting their scope and effectiveness. Additionally, many of these approaches are post-hoc in nature, and, most importantly, often fail to produce explanations that are accessible or meaningful to non-expert stakeholders \citep{BewareAI, Galinkin, Advattack,alvarezmelis2018robustness}

Bridging this gap between AI system complexity and user interpretability is an active area of research, with efforts emerging from both technical and non-technical fronts. From the technical perspective, approaches such as automated machine learning (AutoML) seek to democratize AI development by enabling non-experts to design and deploy models without requiring deep expertise in algorithm selection, tuning, or optimization \citep{hutter2019automated, xanthopoulos2020putting}. From the non-expert or decision-maker's perspective, large language models (LLMs), are being trained to enable natural language human-computer interaction (HCI), thereby reducing reliance on technical intermediaries \citep{zhao2024surveylargelanguagemodels, minaee2024largelanguagemodelssurvey}. Nevertheless, making AutoML outputs interpretable to non-experts remains an open research problem, as current solutions often assume a baseline level of machine learning proficiency \citep{humancenteredautoml, pfisterer2019towards}. 
Similarly, LLMs face several limitations that hinder their ability to deliver high-quality explanations, such as limited access to external knowledge beyond their training data and utilization of external tools to refine explanations to a user’s specific needs \citep{cheng2024dated}. Together, these challenges highlight the need for further research to enhance the transparency, interactivity, and personalization of AI explanations for greater trust and usability across diverse user groups. 

To address these challenges, we propose a paradigm shift in the field of XAI by introducing \textbf{Holistic Explainable AI (HXAI)}. On the one hand, HXAI builds upon the foundational vision of AutoML in order to  widen the scope of XAI to encompass the entire data analysis workflow. Such extension, beyond model output, sets the ground for transforming HXAI into a truly end-to-end democratized AI process. Then, on the other hand, HXAI extends traditional XAI to support users across all levels of expertise by promoting transparency, reducing ambiguity, and providing clear definitions of the elements necessary to achieve trustworthiness. In essence, HXAI is a comprehensive research direction aimed at explaining every stage of an analysis  to a broad audience, taking into account varying levels of needs and expertise.

To further support a user-centered perspective, we propose a comprehensive question bank that captures the varying needs of different stakeholders. This serves both to identify gaps in existing XAI approaches and to guide the development of HXAI tools. 
This work is distinctive in that it does not conform to the traditional structure of a review or perspective article; rather, it offers a novel, multidisciplinary viewpoint on explainability, drawing on personal insights, practical experience, and critical engagement with the literature.

\noindent Our main contributions in this work are:

\begin{enumerate}
\item We introduce Holistic Explainable AI (HXAI), a new direction that widens and scales the perspective of traditional XAI towards providing explainability, transparency, and trustworthiness across all steps and all stakeholders of a data analysis process.
\item  We provide a comprehensive taxonomy for the core components of HXAI and establish clear definitions to reduce ambiguity in the field.
\item We incorporate insights from human sciences, HCI, and user studies to identify the characteristics of effective, user-friendly explanations.
\item We propose leveraging AI agents to bridge the gap between experts and non-experts, promoting broader AI adoption through enhanced explainability and trust.
\end{enumerate}

\begin{figure}[tb]
    \centering
    \includegraphics[width=1\textwidth]{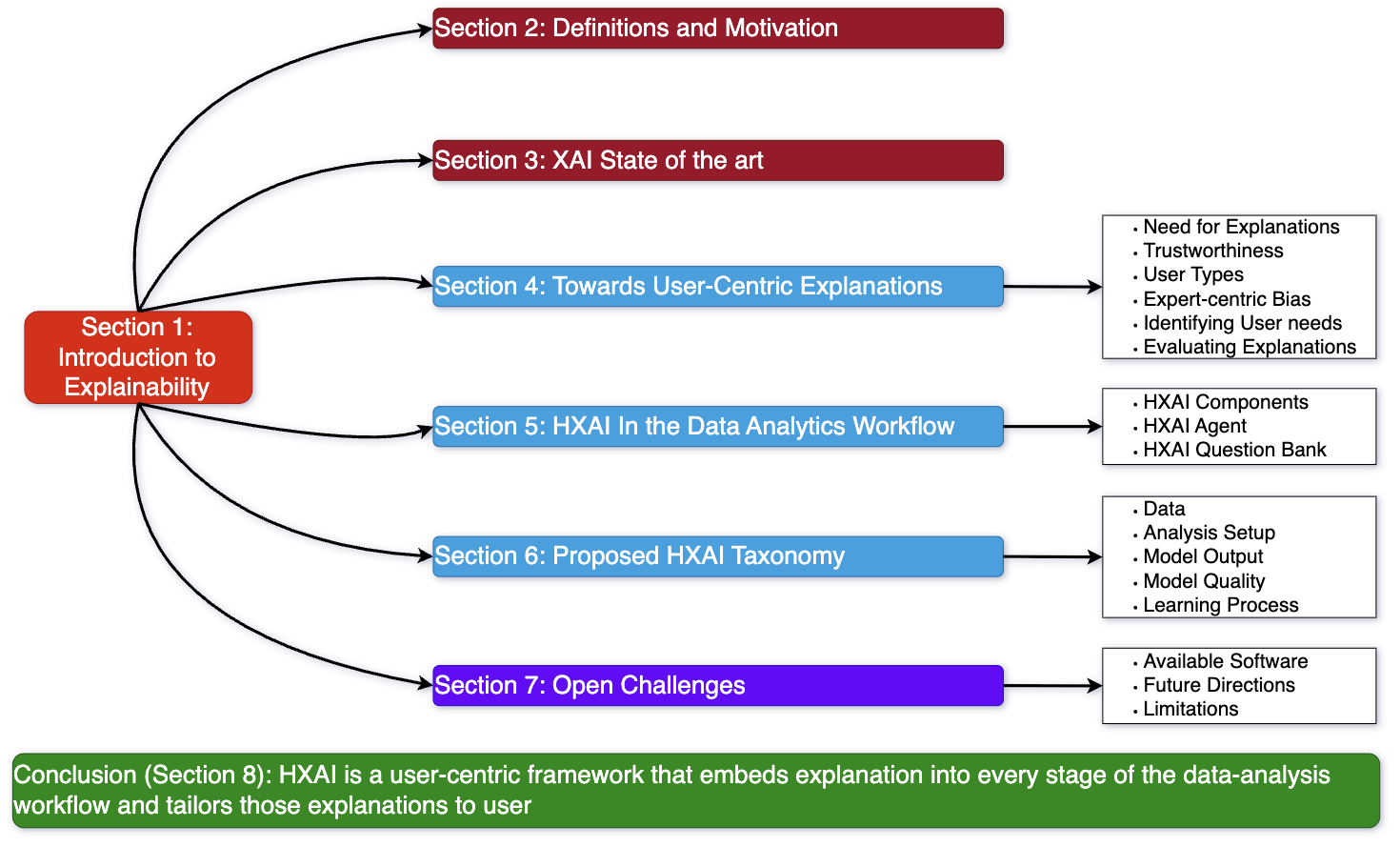} 
    \caption{Outline of the manuscript. Section 1 introduces the concept of explainability and outlines the structure of the paper.
    %which is why arrows extend from it to all other sections, highlighting its role as a conceptual anchor. 
    Sections 2 and 3 (dark red) provide background and survey work in XAI, establishing the necessary foundations.
    Sections 4, 5, and 6 (blue) form the core of our contribution, presenting the proposed HXAI framework. 
    Section 7 (purple) identifies open challenges in HXAI development. 
    Finally, Section 8 (green) concludes the manuscript.
    }
    \label{fig:overview} % Reference label
\end{figure}

\noindent This paper is organized as shown in Figure \ref{fig:overview}, to help readers navigate more easily.
Section 2 provides motivation and definitions for the field of Holistic Explainable AI (HXAI). 
In Section 3, we review the existing literature of XAI and showcase the current limitations. 
Section 4 starts with our contributions by setting the current requirements and desiderata for user-centric explanations, drawing from literature in social sciences, human-computer interaction, and user studies. 
Section 5 describes how HXAI can be embedded into current data analysis workflows, and how an HXAI agent can bridge the gap between different types of users. 
In Section 6, we describe a taxonomy for the HXAI. 
In Section 7, (a) we review current state-of-the-art software, and (b) discuss future research directions for HXAI, and (c) outline the limitations of the current work.
Finally, Section 8 concludes the paper. 

% place figure only top or bottom of the page, not in middle of text

\section{Motivation and Definitions}

\subsection{Motivation}

For AI to achieve widespread acceptance, it must be both accurate and justifiable in its operations. For example, a simple "yes" or "no" response to whether a transaction is fraudulent is inadequate \citep{Mill2023}. Stakeholders need insight into why a transaction was flagged as fraudulent, including the factors that influenced the decision and the confidence of the prediction. This level of interpretability is essential for fostering trust and ensuring that AI systems can be effectively integrated into high-stakes environments.

More generally, all applied-AI projects involve building a solution, evaluating it, and make decisions based on its outputs. Typically, each of these actions is executed by a different stakeholder who possesses markedly different skill sets. For example, data engineers may interpret performance metrics, hyper-parameter logs and feature-importance plots with ease, but they lack the deep contextual knowledge that guides operational or policy decisions. Conversely, domain experts—road-safety managers, clinicians, energy traders—bring invaluable situational insight yet often have little familiarity with the statistical assumptions or optimization heuristics embedded in modern AutoML pipelines.

This asymmetry manifests most acutely at the moment of hand-off. Engineers may deliver a ranked list of influential variables or a plot displaying the network of their partial-dependencies, believing they have provided a transparent rationale. Decision-makers, however, want to evaluate the same artefacts through a causal and regulatory lens: Why does variable X appear more influential than variable Y? How would the prediction change under a feasible intervention? Can the data sources be trusted across all operating conditions? When these questions remain unanswered, or when the semantics of the output are not accurately communicated, even technically sound models can be misinterpreted, lead to wrong decisions, and eventually be dismissed as opaque “black boxes”, eroding confidence and delaying adoption.

The episode underscores a broader methodological challenge: current XAI techniques are optimized for different parts of the analytics workflow, not for cross-disciplinary dialogue. Effective explanations must therefore (i) match the granularity of information to stakeholder expertise, (ii) translate statistical relationships into actionable, counterfactual narratives, and (iii) expose data provenance alongside model-quality diagnostics. Addressing this gap is the central objective of the HXAI framework proposed in this work.

\subsection{HXAI framework}

HXAI is a unified framework that integrates both technical explainability, aimed at data experts, and user-centered explainability that supports understanding among non-expert stakeholders. To operationalize such a framework, it is essential to identify and differentiate key stakeholder groups based on their level of AI competence. 
These include data scientists, machine learning engineers, (data scientist users), data analysts and business intelligence professionals (data analyst users), and domain experts or decision-makers (domain expert users) with limited technical background, who rely on the outputs of AI systems to inform critical decisions. 
In addition, it requires to explore a multidisciplinary approach to explainability, through a collaboration of human sciences, HCI and AI research fields \citep{miller2019explanation,LONGO2024102301}. Such approach can allow for the extraction of fundamental criteria behind effective explanations.

\begin{figure}[htb!]
    \centering
    \includegraphics[width=1\textwidth]{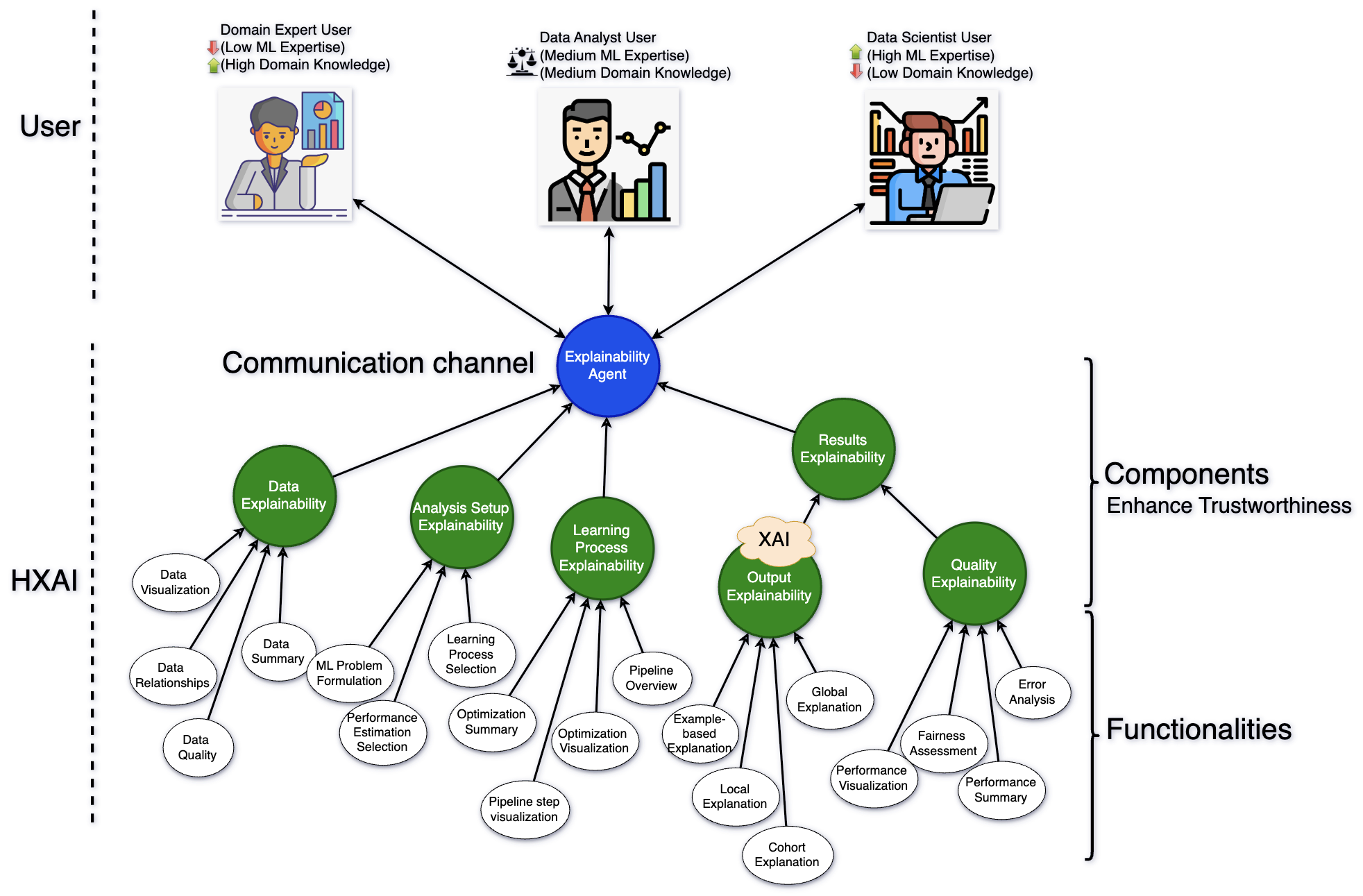} 
    \caption{The proposed HXAI approach. HXAI comprises of six different components which are unified to provide explanations to users with varying AI expertise. Current XAI literature mainly refers to the Output Explainability component as highlighted. HXAI combines information from social science and HCI studies to provide users-needs high quality and relevant explanations through an AI agent.}
    \label{fig:HXAI-Introduction} % Reference label
\end{figure}

Figure \ref{fig:HXAI-Introduction} presents the proposed HXAI architecture. At its core, it comprises of six key explainability components aligned with the stages of the data analysis process: (1) data, (2) analysis setup, (3) learning process, (4) model output, (5) model quality, and (6) the communication channel. This framework enables the integration of explanations across different steps to deliver unified, end-to-end interpretability. 
An LLM-based AI agent (HXAI agent) then tailors these explanations to diverse user needs with natural language output. 
AI agents have recently emerged as an evolution of LLMs \citep{Agents}. These systems enhance the capabilities of the base LLMs by incorporating planning, external tool usage, and contextual augmentation, allowing for more accurate and dynamic responses. AI agents leverage LLMs as their "brain" while integrating external tools such as web search, code interpreters, and databases. These tools supplement the agent’s reasoning process by providing real-time information, executing computations, and retrieving stored knowledge, significantly improving their ability to generate relevant explanations \citep{wang2023interactive}.

\subsection{Definitions}
\label{sec:definition}

In this section, we establish formal definitions for the HXAI field to eliminate ambiguities, addressing a persistent challenge in XAI, where "explainability" and "interpretability" are often used inconsistently \citep{linardatos}. This lack of clarity has been a central topic in numerous XAI studies \citep{linardatos, GUIDOTTI, Adadi, Mythos-Lipton, gilpin}. To avoid similar issues in HXAI, we adopt formal definitions from relevant literature, making a clear distinction between "interpretability" and "explainability." Below, we present the most essential HXAI definitions, with additional definitions available in Section \ref{app:def} for completeness.

\noindent \textbf{Definition 1} \textbf{\textit{(Holistic Explainable Artificial Intelligence (HXAI))}. The research area focused on explaining every aspect of the ML process, including data, analysis setup, learning process, model results, and tailoring explanations to user expertise.}

\noindent \textbf{Definition 2} \textit{(Accountability)}. The ability to determine whether a decision was made according to specific standards \citep{kroll2015accountable,doshi2017accountability}. Accountability allows holding someone responsible if those standards are not met.

\noindent \textbf{Definition 3} \textit{(Fairness)}. The ability of an AI system to make unbiased predictions, without favoring specific populations in the data \citep{ALI2023101805}.

\noindent \textbf{Definition 4} \textit{(Responsibility)}. A model is deemed responsible if it incorporates and upholds the values, morals, and ethics of society \citep{rojat2021explainable}. Responsible AI requires the participation of all stakeholders and members of society \citep{10.5555/3383507}.

\noindent \textbf{Definition 5} \textit{(Robustness)}. The ability of an AI system to work correctly in cases of uncertainty, such as small input perturbations \citep{ALI2023101805}.  

\noindent \textbf{Definition 6} \textit{(Reproducibility)}. When a system systematically obtains similar results in specific conditions \citep{rojat2021explainable}.

\noindent \textbf{Definition 7} \textit{(Stability)}. The ability of an AI system to be robust and reproducible.

\noindent \textbf{Definition 8} \textit{(Completeness)}. The extent to which a model's behavior is described accurately \citep{gilpin}. 

\noindent \textbf{Definition 9} \textit{(Interpretability)}. The ability to explain or present a model's behavior in terms that are easy for humans to understand \citep{Velez}. This term is also used interchangeably with "Comprehensibility" according to \citep{GUIDOTTI, Schwalbe2024}. Interpretability is associated with understanding "how" the model made a decision \citep{ALI2023101805}. 

\noindent \textbf{Definition 10} \textit{(Explainability)}. The process of providing an explanation as a bridge between a model and a human, ensuring the explanation is both accurate (complete) and easy to understand (interpretable) \citep{GUIDOTTI, gilpin}. Explainability is linked to the ability to understand "why" the model made a decision \citep{ALI2023101805}.

\noindent \textbf{Definition 11} \textit{(Interactivity)}. The ability of the explainable system to interact with the user. This interaction can happen in the form of dialogue and user questions.

\noindent \textbf{Definition 12} \textit{(Holisticness)}. An XAI methodology is said to be holistic when it explains the whole process that led to the results.

\noindent \textbf{Definition 13} \textit{(Human-centered)}. A system is human-centered when it takes into account the user expertise, needs, and goals.

\noindent \textbf{Definition 14} \textit{(Trustworthiness)}. The ability of the system to be trusted by the user.

\noindent \textbf{Definition 15} \textit{(Explainable Artificial Intelligence (XAI))}. The research area dedicated to explaining an AI model’s predictions \citep{Darpa_Gunning}.

\section{XAI State of the Art}
\label{sect:stateofart}

In this section, we provide an overview of the current state of AI explainability. We begin by reviewing recent surveys and reviews on XAI, followed by an analysis of current gaps in the literature and a discussion of how HXAI addresses them. 

\subsection{XAI Literature Review}
There have been numerous surveys published in the field of XAI recently \citep{ALI2023101805,ABUSITTA2024124710,Schwalbe2024,Dwivedi,linardatos,MONTAVON,Adadi,GUIDOTTI,gilpin}. The majority concentrates on explaining model predictions while only a few address data explainability \citep{Dwivedi,ALI2023101805,Minh2022,burkart} and even fewer results explainability, covering aspects such as robustness and fairness \citep{linardatos}. This narrow focus highlights the broader issue in the XAI literature: explainability is often limited to specific stages of the analysis workflow, overlooking critical components such as data, analysis setup, and model quality. Table \ref{tbl:related} summarizes the scope of these prior surveys in comparison to our work. A brief overview of the most relevant surveys and reviews follows.

\citep{ALI2023101805} proposed creating explanations by focusing on the following four axes (a) data explainability, (b) model explainability, (c) post-hoc explainability, and (d) evaluation of explanations. For each of the axes, they provided a set of questions and a taxonomy. Finally, they conducted a case study to show-case the application of their XAI taxonomy. In a survey by \citep{ABUSITTA2024124710}, recent findings in the field of XAI were presented, along with a newly proposed taxonomy of XAI methods. The study primarily focuses on data-driven machine learning models, identifying four key challenges in XAI and providing guidelines for addressing them. 
Similarly, \citep{Schwalbe2024} conducted a structured literature review and a meta-study on XAI, proposing a "complete" taxonomy of XAI methods by synthesizing taxonomies from previous surveys. Their taxonomy integrates three key aspects: problem definition, explanator, and evaluation metrics. 
Additionally, \citep{Dwivedi} presented the use cases of XAI, distinguishing between two key phases: the understanding phase and the explaining phase. The authors also discussed the stakeholders involved in each phase and introduced a taxonomy based on model and data explainability. Finally, they provided an overview of available software toolkits for XAI. 
\citep{linardatos} proposed a taxonomy based on different perspectives of interpretability methods. They identified four key aspects: whether the interpretability method is model-agnostic, the type of data it applies to, the purpose of interpretability, and its scope (i.e., single-instance explanations vs. general model behavior). 
\citep{MERSHA2024128111} reviewed over 200 articles on XAI literature, addressing previous limitations and gaps. They also discussed advantages and disadvantages of each XAI methods in depth.
A taxonomy and definition of XAI focused on Deep Neural Networks was also provided by \citep{MONTAVON}. 
Similarly, \citep{danilevsky-etal-2020-survey} focused on XAI methods for Natural Language Processing (NLP), exploring various aspects of explanation methods, including natural language explanations. Additionally, they reviewed existing evaluation methodologies and identified gaps in the literature.
\citep{burkart} presented five distinct approaches to explainability within the supervised learning domain. Their work identifies diverse user groups and proposes assessment methods for XAI. Through illustrative use cases, they clarify the application of the various explainability techniques. Furthermore, they highlight the critical role of data quality.
\citep{Adadi} introduces key concepts of XAI, providing definitions for fundamental terms. Their work outlines four primary motivations for XAI: justification, control, improvement, and discovery. Additionally, they present a taxonomy of XAI methods based on a comprehensive literature review. 
\citep{GUIDOTTI} identifies four key black-box problems in XAI: model explanation, outcome explanation, model inspection, and the transparent design problem. For each, they provide definitions, a literature review, and a taxonomy based on four specific aspects: the type of black-box problem, the type of explanator, the type of data, and the black-box model. 
Recently, \citep{Minh2022} categorized XAI methods in three levels of explainability, pre-modelling, interpretable model and post-modelling. Furthermore, they delve into the open challenges of XAI.
Finally, \citep{gilpin} provided a set of XAI definitions, a taxonomy, and proposed best practices for interpretability in AI. Their survey, which focuses on explaining Deep Learning methods, categorizes XAI approaches into three main groups: Processing, Representation, and Explanation Producing.

\begin{table}[!htbp]
    \centering
    \caption{This table presents the topics covered in prior research papers on XAI vs ours. "M." denotes model. "Perf." and "Pred.", performance and prediction respectively. Learning Process is denoted by "L.Proc". "Q.Bank" represents Question Bank. \label{data-tbl}}
    \label{tbl:related}
    \renewcommand{\arraystretch}{1.5} % Adjust row spacing
    \setlength{\tabcolsep}{1pt}       % Adjust column spacing
    \begin{tabular*}{\textwidth}{@{\extracolsep\fill}lccccccccc}
    \toprule
        & \multicolumn{5}{c}{Explainability} & \multicolumn{3}{c}{} \\
    \cmidrule(lr){2-6}
    & & & \multicolumn{2}{c}{Model Results} & & & & \\
    \cmidrule(lr){4-5}
    Paper & Data & Analysis & M. Perf. & M. Pred. & L. Proc. & User & Tools & Q.Bank \\
   
    \midrule
    \citep{ALI2023101805} 
    & $\blacksquare$ 
    & 
    & 
    & $\blacksquare$ 
    & 
    & $\blacksquare$
    & $\blacksquare$
    & 20
    & \\
    \citep{ABUSITTA2024124710} 
    & 
    & 
    & 
    & $\blacksquare$
    & 
    & 
    & 
    & 
    & \\
    \citep{MERSHA2024128111}
    & 
    & 
    & 
    & $\blacksquare$
    & 
    & $\blacksquare$
    & 
    & 
    & \\
    \citep{Schwalbe2024} 
    &
    & 
    & 
    & $\blacksquare$
    & 
    & $\blacksquare$
    & 
    & 
    & \\
    \citep{Dwivedi} 
    & $\blacksquare$
    & 
    & 
    & $\blacksquare$
    & 
    & $\blacksquare$
    & $\blacksquare$
    & 
    &\\
    \citep{linardatos} 
    & 
    & 
    & $\blacksquare$
    & $\blacksquare$
    & 
    & 
    & $\blacksquare$
    & 
    &\\
    \citep{MONTAVON} & & & & $\blacksquare$& & & & &\\
    \citep{danilevsky-etal-2020-survey}& & & & $\blacksquare$& & & & &\\
    \citep{burkart}&$\blacksquare$ & & &$\blacksquare$ & &$\blacksquare$  & & &\\
    \citep{Adadi} 
    & 
    & 
    & 
    & $\blacksquare$
    & 
    & 
    & $\blacksquare$
    & 
    &\\
    \citep{GUIDOTTI} 
    & 
    & 
    & 
    & $\blacksquare$
    & 
    & 
    & 
    & 
    &\\
    \citep{Minh2022}
    & $\blacksquare$
    & 
    & 
    & $\blacksquare$
    & 
    & 
    & $\blacksquare$
    & 
    &\\
    \citep{gilpin} 
    & 
    & 
    & 
    & $\blacksquare$
    & 
    & $\blacksquare$
    & 
    & 
    &\\
    \hline
    Ours 
     & $\blacksquare$
     & $\blacksquare$
     & $\blacksquare$
     & $\blacksquare$
     & $\blacksquare$
     & $\blacksquare$
     & $\blacksquare$
     & $112$
    \\
    \bottomrule
    \end{tabular*}
\end{table}

\subsection{Limitations in Current Studies}

Table \ref{tbl:related} presents a comparative analysis of our work and previous reviews. While past research has primarily focused on explaining model output, more recent studies have introduced data explainability methods. However, several critical aspects remain largely unaddressed, including explanations of model performance, insights into the learning process, and the rationale behind the analysis setup.

These gaps result in a fragmented understanding of both the model and its outcomes. Since each of these components is crucial for a comprehensive interpretation of prediction models, we propose below an extended taxonomy that incorporates all such aspects. Furthermore, we analyze the needs of users with varying levels of expertise, leveraging also feedback from our in-house teams and external collaborations. Despite the fact that six studies review XAI stakeholders and highlight the importance of effective explanations, only two of them provide an in-depth analysis of what constitutes a good explanation \citep{ALI2023101805,burkart}.

Additionally, only one review offers a question bank. We fill the gap, by proposing and presenting an extensive question bank of over 100 questions across all HXAI components. This unique resource can serve as a foundation for interactive XAI systems, as well as a benchmark for evaluating them in user-studies. 

To this direction, we extend the list of tools by including \textbf{both} XAI and AutoML tools, along with an overview of their explainability capabilities assessed through our proposed question bank. This analysis reveals current gaps in the available tools, both within the XAI and AutoML domains. We note that this list is not limited to open-source tools, but extends to closed-source ones. Our approach aims to bridge existing gaps and present a more unified approach (HXAI) for addressing the challenges of AI explainability.

\section{Towards User-Centric Explanations}

This section shifts the focus to the human element at the core of explainability, exploring the critical interplay between AI systems and their end-users. We first examine why explanations are needed, diving into the motivations driving the push for explainability, such as learning \citep{Hassija2024}, model debugging \citep{kulesza2015principles,Adadi}, and governance \citep{MERSHA2024128111}. Next, we categorize the explainee types, which is the user-base of XAI, showcasing how their needs and goals vary. This leads to a discussion of the expert-centric bias in XAI, a common pitfall where explanations are tailored primarily for those with ML knowledge, often overlooking the non-experts \citep{gilpin,du2019techniques,langer2021we}. We then address users’ needs for explanations, identifying key requirements, and explore how this gap can be bridged.

\subsection{The need for explanations}

In general, explanations are needed for learning, for making prediction, for finding meaning, or for creating a shared understanding, that is changing the beliefs of other people \citep{miller2019explanation}. 
The need for explanations in ML is caused by a variety of reasons. 
First, a user may seek explanations, as the system's output may not be as expected \citep{wang2019designing}. 
Second, explanations can offer support in decision making \citep{zhang2020effect}. 
Explanations are needed for black-box methods to gain the trust of users, and be used in critical areas such as the medical field \citep{veliz2021we}. 
Similarly, explanations have been proposed for justification of results, control of flaws and errors, and discovery of new knowledge \citep{Adadi,Hassija2024}. 
Explanations have been proposed for model debugging and improvement, ranging from fairness issues \citep{NIPS2016_9d268236}, to identification of important features, and error analysis \citep{Adadi,MERSHA2024128111}. In particular, explainability has been proposed as a means to enhance key attributes of machine learning systems, including privacy, usability, robustness, and more \citep{Velez}.
Finally, explainability can help model governance, by ensuring that model decisions are ethical and compliant with laws or regulations \citep{MERSHA2024128111}.

\subsection{Trustworthiness as the primary objective}

Trustworthiness has been identified as the primary objective of explainability in AI by several previous studies \citep{rojat2021explainable,ALI2023101805}. Accordingly, we identify four different dimensions of trustworthiness -explainability, stability, responsibility, and human-centered - and the relationships among the HXAI definitions previously introduced in Section~\ref{sec:definition}.

\begin{figure}[htb!]
    \centering
    \includegraphics[width=0.95\textwidth]{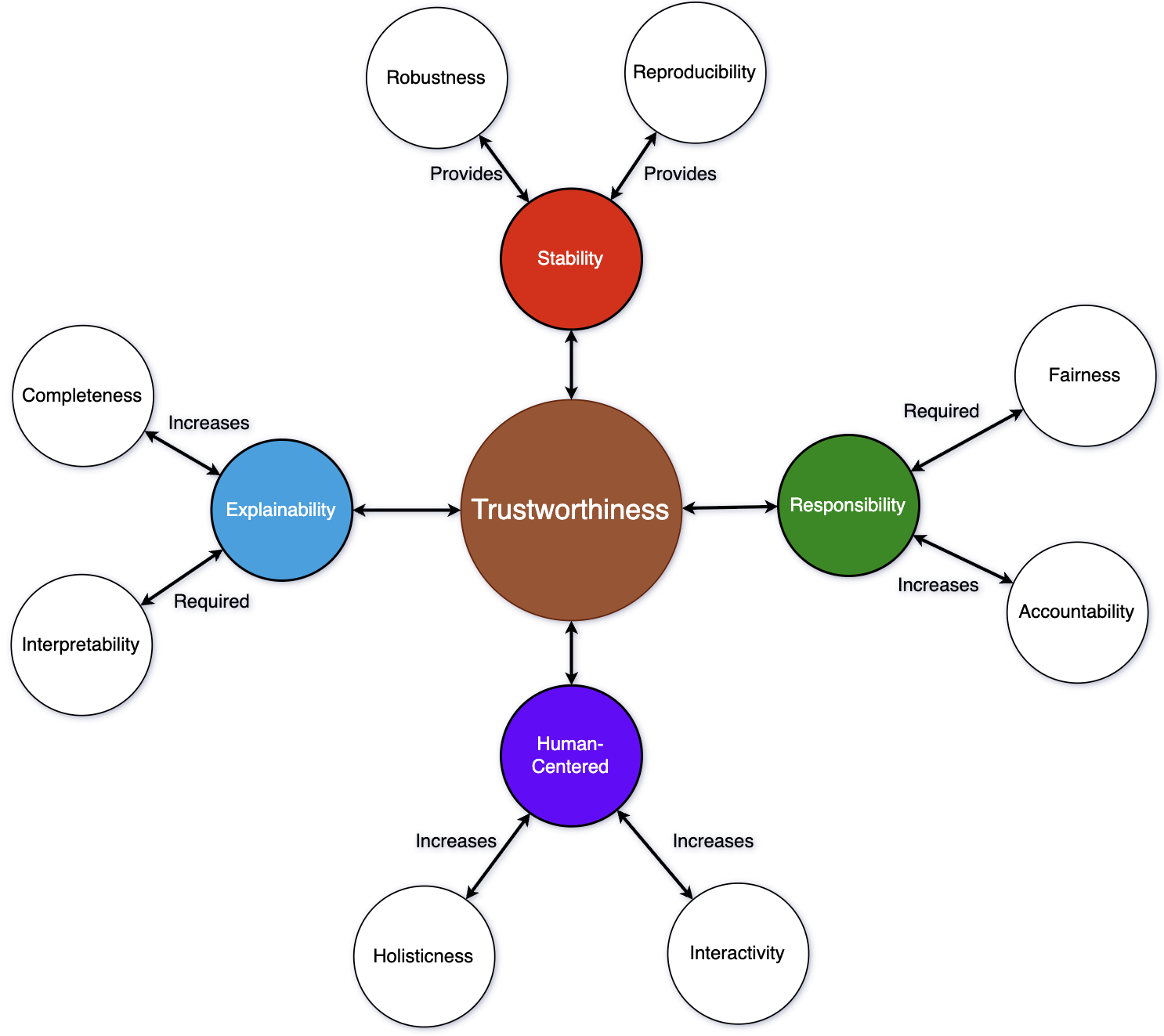} 
    \caption{The graph presents the Trustworthiness along with the different dimensions and concepts that are required to achieve trust of AI systems.}
    \label{fig:trustworthy} % Reference label
\end{figure}

As Figure~\ref{fig:trustworthy} illustrates, to induce trustworthiness we first need explainability which, in turn, is increased by completeness, and requires interpretability. 
Robustness and reproducibility provide stability on the other hand. 
Third, responsibility is concerned with issues such as fairness and accountability as seen in responsible AI literature \citep{10.5555/3383507,BARREDOARRIETA202082}. 
Fairness is required for responsibility, while accountability increases responsibility. 
The final characteristic is the human-centered design, which involves holisticness and interactivity. Both holisticness and interactivity increase the human-centered design. Throughout the remainder of the paper, we present HXAI as a framework for achieving trustworthiness in AI.

To achieve trustworthiness in HXAI, it is essential to first identify the relevant user profiles and understand their specific needs.
Meeting the broader desiderata of trustworthiness requires satisfying multiple criteria through the design of effective explanations, ones that enable users to comprehend and trust AI driven decisions. In the following sections, we examine this challenge from multiple perspectives to explore how explanations can be made more effective and foster user trust.

\subsection{HXAI User types}

The types of explainees (stakeholders) have been previously discussed in research. Building on the works of \citep{langer2021we,BARREDOARRIETA202082,tobias-users}, we introduce three types of users (explainees) who interact with an explainable system. Traditionally, stakeholders have been classified according to their roles rather than their expertise. However, we argue that most users fall into one of the three categories we present in the following section. Each of these user types has a distinct level of data analysis expertise and varying needs for explanations. The degree of expertise directly influences the type of explanations required. Therefore, each user type necessitates a tailored approach to explainability \citep{weller2019transparency}. For example, non-experts maybe interested in a single prediction or a group of predictions, while ML experts maybe more interested in the model's general behavior \citep{rojat2021explainable}. A short presentation of each type of explainee follows. 

\noindent \textbf{Domain Experts}: These users possess deep knowledge in their specific field but have little to no expertise in machine learning (ML). Their primary concern is whether the model's outputs can be trusted and align with their domain knowledge \citep{weller2019transparency}. Domain experts need to have confidence in the system, which is strongly linked to trust \citep{langer2021we}. It's important for the user to have a balanced trust in the system, otherwise under/over-trusting the system could lead to misuse \citep{parasuraman1997humans}. For example, a domain expert might ask, "Is the model good?" To address this, explanations should be simple and focused on high-level metrics or visualizations that are easy to interpret, such as accuracy scores or feature importance charts. These allow the user to have confidence and trust in the system, while not causing cognitive overload.

\noindent \textbf{Data Analysts}: These users serve as a bridge between AI systems and stakeholders, translating AI outcomes into actionable insights. While they have a solid understanding of ML, their primary focus is on understanding and justifying the model's results to domain experts. They are not interested in improving the system, but they may look into issues such as fairness, to make sure the model is ethical, and error analysis, to find potential boundaries or cases where the model shouldn't be trusted. For example, to answer a domain expert’s question, they might ask subsequent questions like, "Has the model over-fitted the data?", "What is the model uncertainty?", "Is the model fair?". Responses should emphasize mid-level diagnostics, such as performance on validation data or indicators of group bias, presented in an accessible manner without deep technical jargon. 

\noindent \textbf{Data Scientists}: As data analysis experts, these users are deeply involved in developing, training, and refining models \citep{weller2019transparency}. Their questions are often highly specific and technical, reflecting their need to understand the model's inner workings. Their main goals are to verify the system working as expected, how to improve the performance of the system, and finding potential errors in the system \citep{langer2021we}. For instance, they might ask, "Is the model calibrated?", "Is it prone to adversarial attacks?", or "How does it perform across various metrics?" Explanations for data scientists should provide detailed and granular insights, including calibration curves, adversarial robustness analysis, global feature importance, and error diagnostics, to help debug and optimize the model.

\begin{figure}[ht!]
    \centering
    \includegraphics[width=0.9\textwidth]{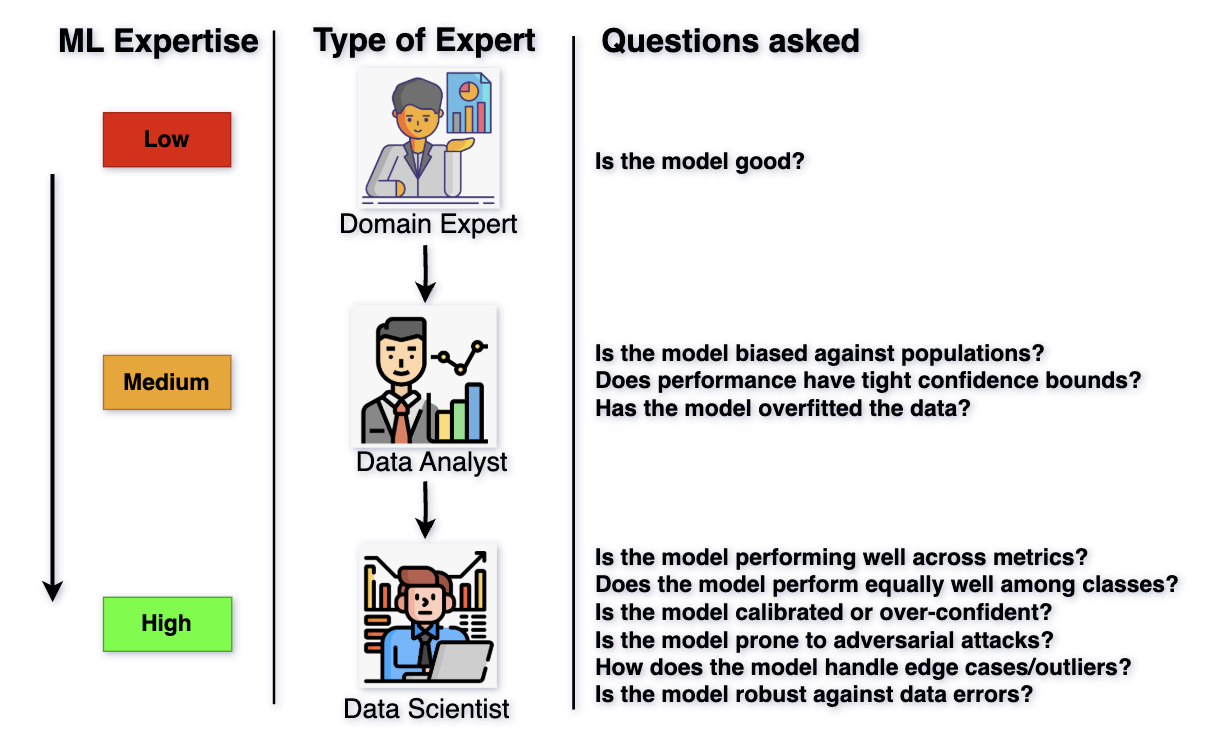} 
    \caption{The figure shows how different expertise in ML leads to different types of questions that need to be answered.}
    \label{fig:HXAI-Quest-user} % Reference label
\end{figure}

Having identified these HXAI user profiles, it becomes evident that a one-size-fits-all approach to explainability is inadequate. Each user group engages with AI systems differently, based on their varying levels of prior knowledge, objectives, and cognitive capacities, which must be considered when designing and presenting explanations. Nevertheless, much of the current focus in XAI remains centered primarily on a single user group, the data scientists.

% \subsection{Expert-Centric Bias in XAI}

% Current Landscape: Overview of how current XAI solutions serve ML experts (technical visualizations, complex metrics, etc.).
% Implications: Discuss why this focus creates a disconnect with non-expert users, affecting trust, usability, and broader adoption.
% Motivating Questions: What are the challenges faced by laymen when interacting with XAI tools?
% How can we redesign XAI interfaces to be inclusive?

\subsection{Identifying User needs}
\label{relatedwork:UserFriendlyExplanations}

Data Scientists have been developing XAI methods for themselves, and not for the actual users \citep{miller2017explainable}. Current explanation tools, are designed for, and understood by ML experts \citep{du2019techniques,langer2021we,gilpin}. These developments have led to a gap between the developers of XAI tools and the actual users, which we call "expert-centric bias". XAI was proposed to make AI more explainable, however the gap remains as the XAI results are not understood by non-experts. To bridge this gap, a cross-disciplinary collaboration is proposed between researchers and practitioners in explainable AI and those in the social and behavioral sciences \citep{miller2017explainable}. Understanding the desiderata of each stakeholder is fundamental to bridge the gap \citep{langer2021we}. 

In this section, we address the expert-centric bias in what defines a good explanation by exploring previous research on user needs \citep{miller2017explainable}. As in \citep{LONGO2024102301}, we draw multidisciplinary insights from social sciences, human-computer interaction (HCI), user studies, and XAI evaluation studies. Furthermore, we propose three strategies for delivering effective explanations \citep{langer2021we}. First, multiple explanations should be provided to the user, offering diverse information and presentation modes. Second, interactive explanations are essential, starting with high level information, and allowing users to progressively get details on specific parts of the workflow. Third, a trained stakeholder is proposed to bridge the gap and provide further explanations to non-experts. In HXAI, this stakeholder will be the Data Analyst.

\subsubsection{Insights from Human Sciences}

Recently, researchers have suggested that XAI should prioritize studying users' mental models rather than solely developing new explanation methods \citep{Páez2019}. Mental models are internal representations that individuals construct to understand, explain, and predict the behavior of systems or phenomena \citep{kulesza2013too}. In the context of explainable AI, mental models help users grasp how an AI system operates \citep{mohseni2021multidisciplinary}. Better mental models lead to higher task performance when humans collaborate with AI (Human-AI team) \citep{bansal2019updates}. Without a clear understanding of users' needs and expertise, the effectiveness of these models remains uncertain.

Other series of essays, outlined the theoretical foundations of explanations, emphasizing causal reasoning as the core mechanism behind decision-making, sense-making, and mental model construction \citep{hoffman2017explaining, hoffman2017explaining2}. Their findings suggest that people initially rely on simple explanations, progressively moving into more detailed ones. Building on these theoretical insights, \citep{miller2019explanation} reviewed research from the social sciences (including philosophy, psychology, and cognitive science) to present user-friendly explanations. Explanations were characterized as (1) contrastive, explaining why one decision was made instead of another, (2) selective, highlighting only the most relevant information for a decision, (3) credible, aligning with general knowledge and user expectations, and (4) conversational, structured as a dialogue between the explanator and the explainee. Causal explanations have also been deemed as conversational \citep{Hilton1990ConversationalPA}. Causal explanations identify the reasons behind a relationship and should be, first, relevant to "why" questions, and, second, true.

Along these lines, \citep{lim2009assessing} explored different types of explanations, highlighting the role of "why" and "why not" questions in enhancing understanding. To ensure user trust they suggested the availability of high-quality information for explanations to be effective. The types of explanations, their limitations and functions have also been studied by \citep{keil2006explanation}. In their view, explanations can expose gaps in understanding, which humans usually over-estimated. Explanations come in different forms and form a variety of functions. Accordingly, \citep{lombrozo2006structure} examined the structure and function of explanations. Structurally, explanations consist of the explanandum (what is being explained) and the explanans (the explanation itself). Functionally, explanations serve as mechanisms for generalization to novel cases, facilitating understanding and satisfaction, guiding reasoning, and supporting prediction and control of future events.

\subsubsection{Insights from Human-Computer Interaction}

Research in human-computer interaction (HCI) focuses on end-user needs such as trust and understanding of machine generated explanations \citep{mohseni2021multidisciplinary, Kong2024}. Human-computer trust has been explored in \citep{muir1987trust} where authors emphasize the importance of \textit{trust calibration}. That is, allow the user to identify when the model’s decision is correct, and when it’s wrong. To achieve this, they suggest several key strategies. First, is to train users through multiple system interactions to improve their understanding. Second, to continuously update user expectations regarding the AI system’s competence. Third, to design AI systems that enhance the user decision making rather than  replacing it. Forth, to iteratively improve the trust based on ongoing experience and feedback.

Trust calibration is also central to AI-assisted human decision-making \citep{bansal2019updates,Bansal_Nushi_Kamar_Lasecki_Weld_Horvitz_2019,zhang2020effect}. This form of human-AI collaboration is preferred in high-stakes scenarios over fully automated AI decisions. The primary goal is to help users develop a mental model that understands the AI's error boundaries \citep{bansal2019updates}, enabling them to either accept or override the system's decisions. Improving the mental model has also been the focus of \citep{eiband2018bringing} who implemented a stage-based framework. The process included two steps answering to the following: (a) what to explain and (b) how to explain it. 

On the other hand, a variety of multidisciplinary approaches incorporating HCI and XAI have been developed. For example, assessing "who" the user is, by technical and social approaches, and "Why" an explanation is needed has been the focus of the recently proposed Human-Centered Explainable AI (HCXAI) \citep{10.1007/978-3-030-60117-1_33, ehsan2021operationalizing}. Similarly, YAI (why AI) has been proposed as a theoretical approach to user-centered explanations \citep{Sovrano2024}. It is based on two key pillars: illocution, which addresses questions such as how, what, and why; and pragmatism, which tailors explanations to the user’s background, needs, and goals. In their work, the authors propose that user-centered explanations are individual goal-driven paths within a vast, unbounded space of possible explanations. The properties of these paths, such as their length and direction, depend on the specific type of user the explanation is designed for. 

Another human-centered approach, Theory Driven XAI, has been designed with the user in mind \citep{wang2019designing}. This framework focuses on four criteria. First, how people reason and explain. Second, how XAI generates explanation, Third, how people actually reason, including potential errors. Finally, how XAI can support reasoning, and mitigate those errors. To verify the proposed framework, a user study was conducted, 14 domain experts were recruited to examine explanations in medical diagnosis. Users sought support for alternative hypothesis generation, which can be facilitated through contrastive and counterfactual methods. Other identified needs included detailed rather than shallow explanations, the integration of multiple explanation methods, access to raw data to enhance trust, and feature interaction plots to illustrate data relationships.

%In the field of HCI, a new approach to explainability, named eXplainable AI for Designers (XAID), has been proposed \citep{zhu2018explainable} to support game designers. XAID focused on specific user needs and tasks, instead of focusing on developing new XAI methods. Authors described Explainability as \textit{"the ability to answer why questions"}.

\subsubsection{Insights from User Studies}

There have been several user studies conducted to identify key criteria for explanations tailored to non-experts. These typically measure one of three user qualities: (a) user's task performance when provided with explanations, (b) trust calibration, meaning knowing when to trust the model and when not to, and (c) self-reported understanding and trust in the AI. 

Understanding the whole prediction process leading to the presented results has helped 17 non-expert participants better understand a recommendation system model \citep{kulesza2013too}. Users also reported higher trust in the system when they were provided with highly detailed explanations and understood the entire process \citep{VANDERWAA2021103404}. Specifically, rule-based and example-based explanations were found to be persuasive, often increasing users' reliance on the AI predictions, even when those predictions were incorrect. However, findings also showed that neither explanation method significantly improved users’ task performance. In a large user study involving over 1,500 participants, individuals who were presented with an explanation on each individual recommendation made by a model were more likely to accept the AI’s decisions, even when those decisions were incorrect \citep{bansal2021does}. Overall, the combined AI-human performance surpassed that of the AI alone. However, individual explanations did not significantly enhance task performance compared to simply showing the AI’s confidence scores (i.e., predicted probabilities). Confidence scores (classification probabilities) were also employed in a study with 72 participants by \citep{zhang2020effect}, who found that users' trust increased in line with the AI's scores. However, this trust didn't also lead to improved accuracy in the final decisions (task performance). In a second experiment, they tested the effect of single-prediction explanations on model trust vs the knowledge of only confidence scores. The effect of single-prediction explanations was found to be insignificant compared to the baseline.

Regarding the size of the explanations, a user study involving 100 participants, found that larger explanation sizes, variable repetition, and an increased number of cognitive chunks \textit{negatively} impacted users' response times and satisfaction \citep{narayanan2018humans}.  Similarly, when users' expectations were violated, explanations with a moderate level of detail enhanced their trust in the system's decisions \citep{kizilcec2016much}. Specifically, it was found that highly detailed explanations could negatively impact trust, highlighting the need for balance. However, when the system's decisions aligned with users' expectations, explanations had little effect on trust. Furthermore, meaningless explanations can hinder the user's ability to perceive the AI system's performance as seen in a user study with 60 participants \citep{nourani}. Lastly, a user study involving 25 participants identified three key actions to improve human mental models: (a) reducing task dimensionality by eliminating irrelevant features, (b) minimizing the randomness of model errors, and (c) simplifying the complexity of the decision boundary \citep{Bansal_Nushi_Kamar_Lasecki_Weld_Horvitz_2019}. 

From a non-experts' perspective, a study indicated that these users are not happy with current explanations format (static) and would prefer interactive explanations through natural language conversations \citep{lakkaraju2022rethinking}. Non-Experts emphasized the need (a) to know the correctness of an explanation, (b) have the ability to ask custom questions, and (c) access explanations for specific data sub-groups. This aligns with social sciences research that characterize explanations are conversational \citep{miller2019explanation}. Interestingly, a user study involving 36 participants with diverse expertise assessed the explainability of an AutoML system \citep{zoller2023xautoml}. Domain (non) experts prioritized global surrogate models, whereas ML experts focused on analyzing constructed pipelines and intermediate outputs from pipeline pre-processors. When it came to understanding the learning process, only ML experts benefited. Overall, participants preferred having more explanations rather than too few and showed a strong inclination toward visual explanations.

%Feature-based and example-based explanations have been shown to increase trust in the model predictions, especially when model predictions are correct \citep{lai2019human}.

Finally, the Subjective Information Processing Awareness (SIPA) scale, based on six items from Situational Awareness, has been proposed in \citep{schrills2023users} to measure three key aspects: (a) transparency, (b) understanding of the current situation, and (c) projection of future states. On such basis, a user study involving 70 participants was conducted, evaluating ten hypotheses. The results indicate that relevant information becomes beneficial only after multiple interactions with the system. However, users who received more relevant information spent more time on the task. Interestingly, there was no significant difference in user's task performance between users with low and high amounts of information. The authors attribute this finding to information overload, linked to interactivity, and caution that XAI may not be effective in mitigating errors. 

Even though insights from user-studies are helpful, a recent survey of user studies on XAI found that their results are often inconsistent across key evaluation metrics such as trust, usability, understanding, and human-AI task performance \citep{10316181}. The authors attribute this inconsistency primarily to the use of varying evaluation methodologies across studies.

\subsubsection{Characteristics of Effective Explanations}
\label{sec:effexpl}

Based on the above insights from social sciences, human-computer interaction (HCI), and user studies, effective explanations in Explainable AI (XAI) should be designed with the following key characteristics:

\noindent \textbf{Contrastive, Causal \& Truthful} – Explanations should clarify why one decision was made instead of another and provide causal reasoning that aligns with users’ knowledge and expectations. They should answer "why" questions accurately while maintaining credibility and relevance \citep{miller2019explanation, Hilton1990ConversationalPA, hoffman2017explaining, hoffman2017explaining2}.  

\noindent \textbf{User-Centered} – Explanations should be tailored to the user’s background, needs, and goals. \citep{Sovrano2024, miller2019explanation, ehsan2021operationalizing, schrills2023users}.  

\noindent \textbf{Information-Relevant \& Avoiding Overload} – Users benefit from relevant explanations, but excessive information can cause cognitive overload, reducing efficiency without improving accuracy \citep{schrills2023users}. A balance between too much and too little detail is necessary to maintain user trust and usability \citep{nourani,narayanan2018humans}.

\noindent \textbf{Mental Model Enhancement} – Effective explanations should help users build accurate mental models of AI systems, improving their ability to predict, interpret, and collaborate with the model. A well-developed mental model leads to better task performance and informed decision-making \citep{eiband2018bringing, mohseni2021multidisciplinary}.  

\noindent \textbf{Trust Calibration} – Explanations should help users calibrate their trust in AI by understanding when to accept or override its decisions. This ensures users neither over-rely on AI nor dismiss its outputs without proper consideration \citep{kulesza2013too, mohseni2021multidisciplinary, kizilcec2016much}.  

\noindent \textbf{Supports Human-AI Collaboration} – Explanations should assist users in decision-making rather than replacing human judgment, particularly in high-stakes scenarios. AI should complement human expertise by enhancing decision-making rather than making autonomous decisions \citep{bansal2019updates, Bansal_Nushi_Kamar_Lasecki_Weld_Horvitz_2019, zhang2020effect}.  

\noindent \textbf{Conversational \& Interactive} – Explanations should be structured as dialogues, allowing users to ask follow-up questions and explore different levels of detail. Users prefer interactive formats over static ones, particularly through natural language conversations \citep{miller2019explanation, lakkaraju2022rethinking, Hilton1990ConversationalPA}.  

\noindent \textbf{Iterative} – Explanations should support repeated interactions, as trust in AI systems develops over time through continuous engagement and refinement \citep{muir1987trust, schrills2023users}.  

\noindent \textbf{Visual} – Users, particularly non-experts, favor visual explanations over purely textual ones, as they improve comprehension, engagement, and ease of interpretation \citep{zoller2023xautoml}.  

\noindent \textbf{Holistic} – Explanations should provide a comprehensive understanding of the AI system, covering its overall functionality \citep{zoller2023xautoml, kulesza2013too}.

\subsubsection{Effective Explanations lead to trustworthiness}

\begin{figure}[ht!]
    \centering
    \includegraphics[width=0.8\textwidth]{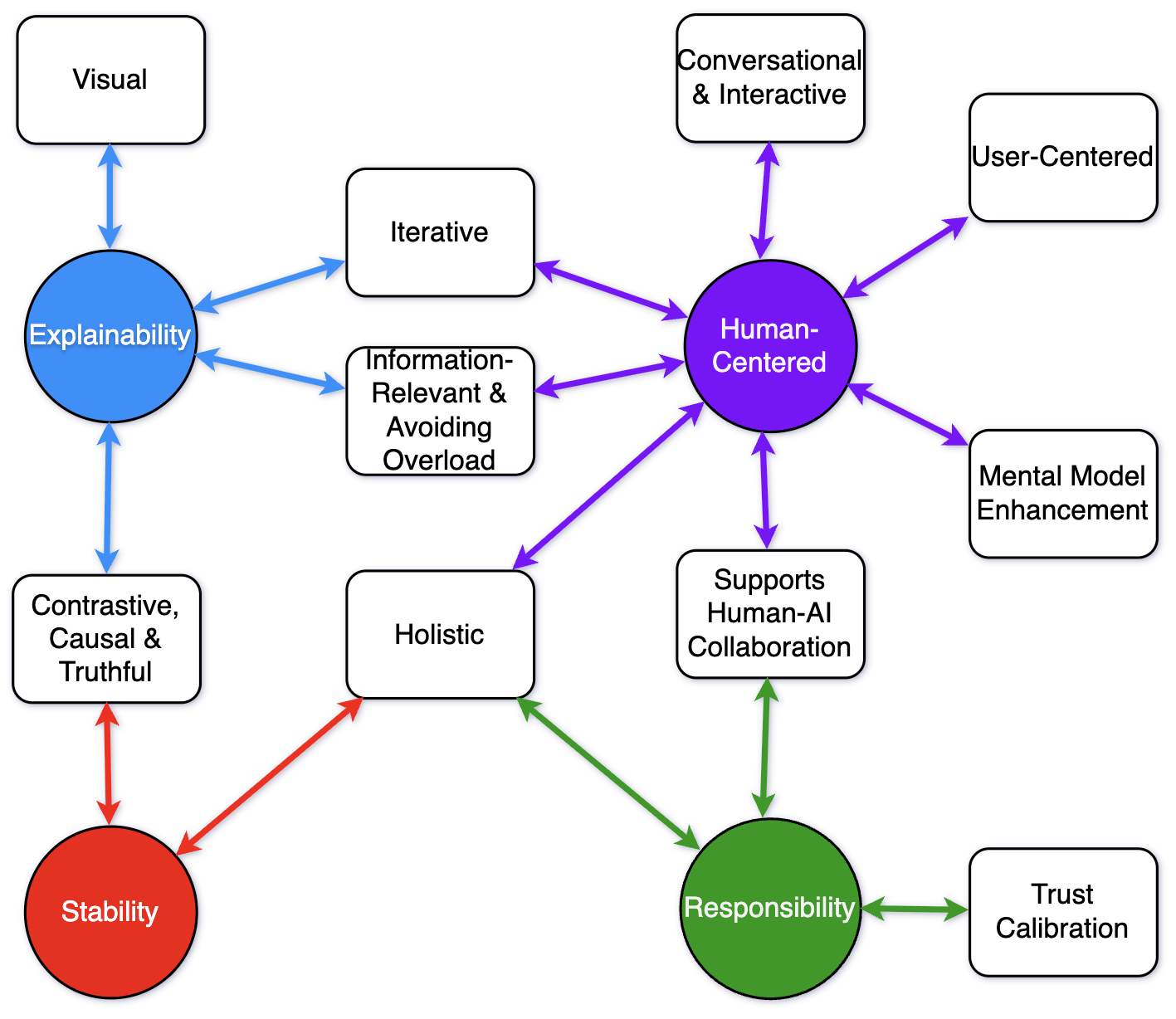} 
    \caption{Mapping of Trustworthiness dimension to supporting Effective Explanation Qualities in HXAI. The white rectangles represent Effectiveness Explanation Qualities, while the colored circles represent trustworthiness dimensions.}
    \label{fig:HXAI-Explanation-Trust} % Reference label
\end{figure}

% Table \ref{tab:Trust-Explanation}
In this section, we map the presented effective explanations to the criteria proposed for trustworthiness in HXAI. As shown in Figure \ref{fig:HXAI-Explanation-Trust}, each of the trustworthiness dimensions (colored-circles) is linked to several effective explanation qualities (white-rectangles). 

Explainability can be strengthened by explanations that are contrastive, causal, and truthful. Additionally, it may be improved with visual and iterative qualities to minimize cognitive overload. Furthermore, explainability can be enhanced by improving the user's mental model. 

Accordingly, human-centeredness can be supported by user-centered designs, that is conversational and has interactive formats. Additionally, explanations should enhance users’ understanding without overwhelming them. Iterative and holistic explanations are also characteristics that can help human-centeredness, enabling users to explore AI decisions and explanations progressively. 

On the other hand, stability in explanations may be fostered through robustness checks using contrastive methods in explanations and the use of holistic representations that explain multiple parts of the ML system. Finally, responsibility can be improved through explanation that calibrate trust, that is helping users align confidence appropriately with model certainty, explanations that support Human-AI collaboration, and that are holistic meaning increased accountability and responsibility through transparency in all steps of the ML system.

\begin{comment}
    
\begin{table}[htb]
\small
\centering
\renewcommand{\arraystretch}{2.5} % Adjust row spacing
\caption{Mapping of Trustworthiness Aspects to Supporting Effective Explanation Qualities in HXAI.}
\label{tab:Trust-Explanation}
\begin{tabular}{ll} % Two columns
\toprule
\textbf{Trustworthiness} & \textbf{Effective Explanation Qualities} \\
\midrule

\makecell[l]{Explainability} 
& \makecell[l]{Contrastive, Causal \& Truthful\\ Information-Relevant \& Avoiding Overload\\ Visual\\ Iterative} \\
\midrule
\makecell[l]{Human-Centered} 
& \makecell[l]{User-Centered\\ Mental Model Enhancement\\ Information-Relevant \& Avoiding Overload\\ Supports Human-AI Collaboration\\ Conversational \& Interactive\\ Iterative\\ Holistic} \\
\midrule
\makecell[l]{Stability} 
& \makecell[l]{Contrastive, Causal \& Truthful\\ Holistic} \\
\midrule
\makecell[l]{Responsibility} 
& \makecell[l]{Trust Calibration\\ Supports Human-AI Collaboration\\ Holistic} \\

\bottomrule
\end{tabular}

\end{table}
\end{comment}

\subsection{Evaluating Explanations}

Another open challenge in XAI research is the evaluation of explanations. Currently, there is no universally accepted metric or methodology for assessing explanation quality. We refer the reader to \citep{app12199423,mohseni2021multidisciplinary} for a comprehensive review of XAI evaluation methods. An overview of XAI evaluation approaches follows.

\citep{rojat2021explainable} introduces two broad categories of evaluation approaches. The first is qualitative evaluation, which relies on expert assessments of explanations. The second is quantitative evaluation, which is based on specific, measurable criteria. They argue that qualitative evaluations may be more suitable for non-experts, whereas quantitative evaluations tend to be more appropriate for individuals with ML expertise.

A similar distinction exists in cognitive psychology, where two types of measurements are commonly used, the subjective and the objective measurements. The former, which are indirect, rely on participants' self-reports. The latter, assess participants' performance on a given task \citep{hsiao2021roadmap}. Notably, these two types of measurements are not always aligned, highlighting the complexity of evaluating explanations.

\citep{hoffman2018metrics} proposes several evaluation measures for XAI systems. First step involves assessing the goodness of an explanation during its generation through structured checklists. Another measure considers user satisfaction, measuring the extent to which users feel they understand the underlying AI system. As users engage with explanations, they develop a mental model of the system, making user understanding of the model, another key evaluation factor. Additionally, evaluating a user’s performance on a specific task can provide further insights, as task performance is influenced by satisfaction with the explanation, trust in the system, and the user’s mental model.

Building on this framework, \citep{hsiao2021roadmap} extend the existing evaluation measures by introducing additional factors such as user curiosity, trust, and system interactivity. They provide guidelines on different measurement types and methodologies, further distinguishing between subjective and objective approaches. This distinction reinforces the importance of considering both perspectives when assessing explainability in AI systems.

Similarly, \citep{MERSHA2024128111} has classified evaluation methods into human-centered and computer-centered. Human-centered evaluation is based on comprehensibility, trust, user satisfaction, and decision-making support. Computer-centered evaluation is focusing on metrics such as fidelity, consistency, robustness, efficiency of an XAI method.

Finally, \citep{doshi2018considerations} presented three evaluation approaches. First, Application-grounded which include real world tasks involving real-humans, where the evaluation is based on how explanations help humans perform a task. Second, human-grounded evaluation which includes real-humans but simplified tasks. This evaluation includes laypeople, and usually measures general attributes of explainability such as understanding under time-constraints. The third approach doesn't include humans, and is based on proxy tasks. The problem of this approach is what should be the selected proxy, after that it's simply an optimization problem.

To address the open challenge of explanation evaluation, we observe that current approaches span a range of evaluations from human-centered (qualitative, subjective, task-based) to computer-centered (quantitative, objective, proxy-based), each with distinct strengths and limitations. The key takeaway is that no single evaluation method suffices for effective assessment often requires combining subjective user experience (e.g., trust, satisfaction) with objective performance metrics (e.g., fidelity, robustness). In the context of HXAI, we propose a hybrid evaluation strategy aligned with four dimensions: employing task performance and model fidelity for system-level explainability, while integrating user studies and cognitive alignment assessments for human-centered understanding. This approach addresses the current fragmentation by unifying evaluation through both technical and human-centric lenses, ultimately enabling personalized and context-aware explanation assessment.

\section{Holistic Explainable AI in the Data Analytics Workflow}
\label{sec:HXAI-Intro}

HXAI is a unified approach that incorporates several XAI aspects covered mostly in isolation in the past, aiming to provide in-depth explanations that are adapted to the user's skill. In this section, we (a) introduce and illustrate how the six HXAI components align with the typical data analysis workflow, introducing the HXAI~Agent, who aggregates the explanatory information and provides a user-centered explanation based on the user's profile. Finally, we present a comprehensive question bank to support HXAI implementation and benchmarking.

\begin{figure}[tb]
    \centering
    \includegraphics[width=1\textwidth]{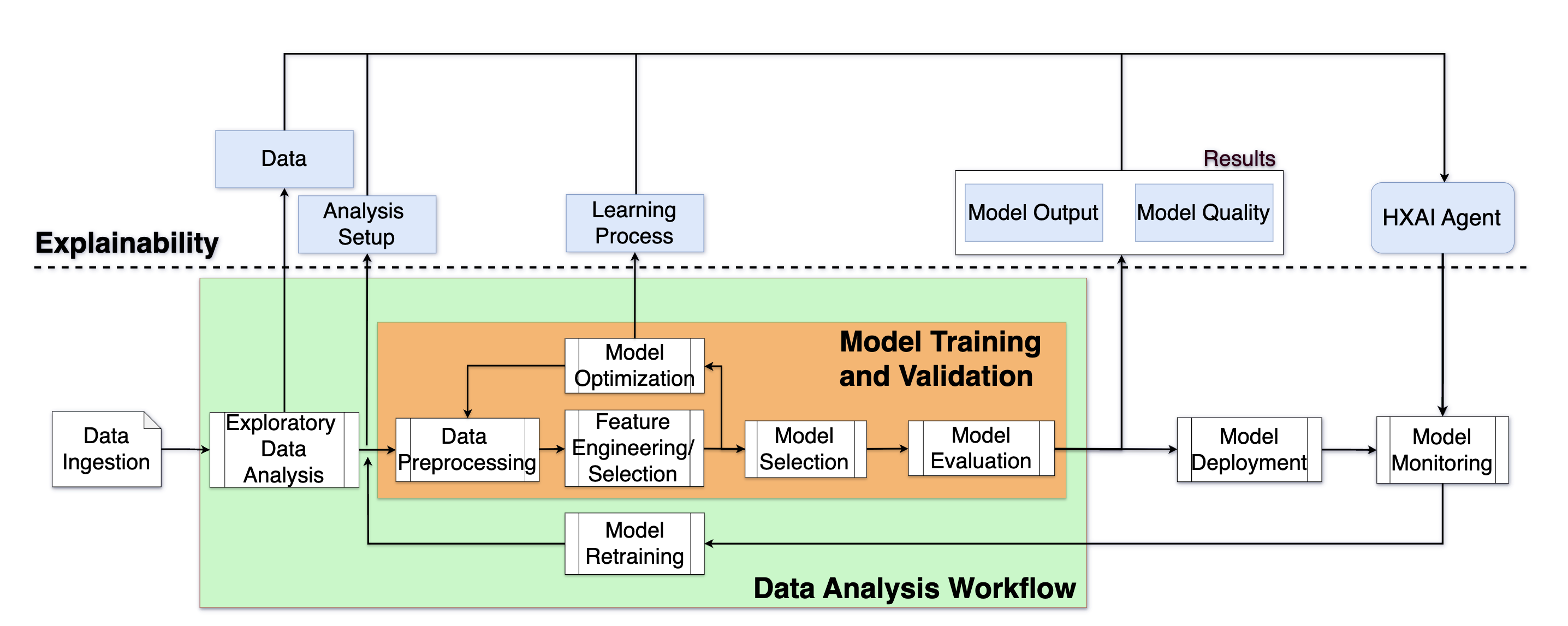} 
    \caption{The HXAI framework aligns well with the traditional ML analysis pipeline. Core components of HXAI are shown in sky blue. The green rectangle indicates the parts of the data analysis process, such as Exploratory Data Analysis (EDA), Model Training \& Validation (orange rectangle), and Model Retraining. }
    %White rectangles show individual components in real-world commercial ML pipelines including serving ( Deployment + Monitoring ).
    %HXAI agent is taking the information from the blue rectangles (lines omitted for simplicity) and tailors the explanation to the user.}
    \label{fig:HXAI-Pipe} % Reference label
\end{figure}

\subsection{HXAI Components in the Data Analysis Workflow}

Figure \ref{fig:HXAI-Pipe} depicts the components of a Data Analysis workflow from data ingestion to model deployment and monitoring. We assume that all methods within the Explainability and Data Analysis Workflow are mathematically sound and technically correct. Furthermore, we consider that all necessary task-specific and methodological requirements are fully met, allowing us to build upon a solid and reliable foundation.
At the core of the workflow is the model training and validation process. HXAI components, highlighted in sky blue, fit into different steps of the workflow. Finally, a central component that orchestrates the communication of the explanations to the different types of users is considered, namely the HXAI agent. This component is responsible for aggregating information from all preceding HXAI components, and serving explanations to the user by accounting for his needs, based on the explainability qualities discussed in the previous section. A key advantage of this approach is its compatibility with any data analysis process, making it highly adaptable across different scenarios. As illustrated in Figure \ref{fig:HXAI-Pipe}, the data analysis workflow (green box) is independent of the HXAI components (blue boxes), ensuring modularity and flexibility.

In more detail, the first component of HXAI, \textbf{Data Explainability}, enhances Exploratory Data Analysis (EDA) by emphasizing a deeper understanding of the data. This is achieved through data summarization, pre-analysis data quality assessment, data relationship analysis, and data visualization. For instance, this component computes key statistics for continuous variables, such as minimum, maximum, mean, and standard deviation, and visualizes distributions using histograms. Uni-variate analysis helps detect outliers, which can be highlighted in visual or textual formats. Additionally, by examining relationships within the data, it can identify highly correlated features and similar samples.

Before initiating model training, the \textbf{Analysis Setup Explainability} component ensures transparency in the machine learning problem formulation, the performance estimation process needed to execute, and the learning process that follows. This component helps users understand what is the specific problem being addressed, the optimization analysis required, and the rationale behind the selection of the specific learning process. 
For example, consider a user analyzing breast cancer data. The analysis setup component clarifies that the prediction task is to determine whether a patient has cancer or not. Consequently, a classification task is put forward and that the F1-Score is chosen as the optimization metric due to the the data structure and the importance of correctly identifying positive cases. Additionally, if the dataset is small, the component informs the user that a repeated cross-validation protocol followed by a performance correction method are needed and will be employed in order to ensure unbiased estimation of the model performance.

Thereupon, during Model Training \& Validation, another key component of explainability emerges: \textbf{Learning Process Explainability}. This component provides insights into how model optimization progressed, both in real-time and post-analysis, offering a clearer understanding of the training process.
It can visualize the machine learning pipeline, track how hyper-parameters were explored across iterations, and display performance metric curves over time. By presenting these insights through intuitive visualizations, users can better interpret the model's learning ability, identify potential issues during model training, and make informed adjustments to improve the predictive model's performance.

Finally, once the model generation process is complete, the \textbf{Results Explainability} component comes into play. This component incorporates traditional XAI methods to clarify both the model’s predictions and its predictive performance.
Results Explainability can be explored in two key areas:

\noindent \textbf{Model Outputs}: This component leverages XAI to explain predictions at different levels: individual predictions, cohort/sub-group predictions, and overall data-level predictions (general model behavior).

\noindent \textbf{Model Quality}: This component focuses on performance metrics, including performance curves, trade-offs, and even auxiliary metrics such as fairness metrics. A crucial functionality of this component is error analysis, which identifies where and why the model makes incorrect predictions.

These insights are useful in isolation to understand parts of the data analysis workflow. However, by combining these insights, users can gain a deeper understanding of both how the model arrives at its decisions and how well it performs across different scenarios.

\subsection{The HXAI Agent}

The HXAI framework was designed to meet the needs of a diverse range of users by providing effective explanations and fostering trust in AI systems. We defined in Section \ref{sec:effexpl} what effective explanations require by adopting a multi-disciplinary approach. HXAI fulfills these requirements across multiple HXAI components, such as data, setup, learning process, model output, and model quality. Additionally, HXAI is promoting key dimensions of Trustworthy AI, such as stability, responsibility, and human-centeredness (see Figure \ref{fig:trustworthy} for an overview of Trustworthiness).

\begin{figure}[htb!]
    \centering
    \includegraphics[width=0.9\textwidth]{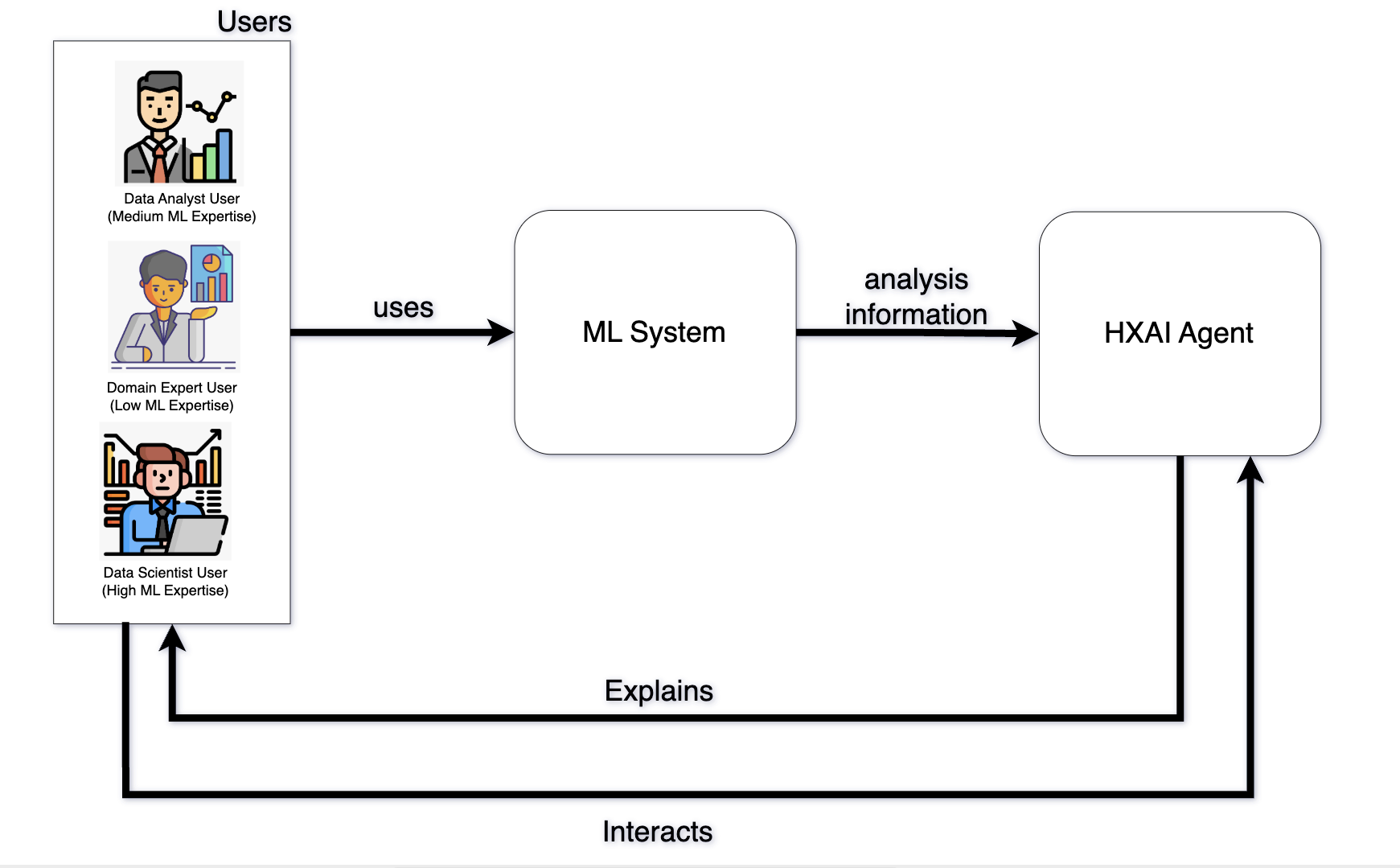} 
    \caption{High-level overview of the HXAI Agent functionality within any data analysis workflow. A user can initiate the analysis process. During and after the analysis execution the system transfers all available information to the agent. The agent then interacts with the user, delivering clear explanations and insights throughout the analysis.}
    \label{fig:high-level} % Reference label
\end{figure}

However, traditional explainability methods often fall short in terms of personalization, interactivity, and adaptability. We propose an LLM-powered AI Agent framework for HXAI, referred to as the \textbf{HXAI Agent}. This component acts as the interface (communication-channel) between the technical components of HXAI and its end-users, delivering explanations in an accessible and user-centered manner. Figure \ref{fig:high-level} shows a high level overview of the HXAI Agent integration. 

By harnessing the generative and adaptive capabilities of LLMs, the HXAI Agent produces coherent, personalized explanations in natural language that are tailored to individual user needs. 
Beyond static response generation, the agent extends LLM functionality through dynamic tool integration, enabling context-aware, interactive, and user-tailored explanations.

In the following sections, we examine the HXAI Agent in depth. First, we analyze how the agent aggregates relevant information across the preceding HXAI components and performs a type of root-cause analysis to improve system transparency. Second, we assess the suitability of LLMs as the main reasoning engine of the component. Third, we investigate how AI agents can augment the capabilities of LLMs, particularly in the context of delivering effective and contextualized explanations. Finally, we evaluate how the HXAI Agent supports trustworthiness by examining: (a) the specific explanatory needs of users that it addresses, (b) its characteristics, and (c) the HXAI components it integrates to ensure user-aligned explanations.

\subsubsection{Information Aggregation}

Each HXAI component is responsible to explain an aspect of the data analysis workflow. Nevertheless, in many cases an explanation requires knowledge from multiple components. For example, in a classification task where the distribution of the observations in different classes is skewed, the choice of an appropriate, imbalance-invariant, performance optimization metric is very important. To solve this problem, information from the Data Explainability and the Analysis Setup Explainability components needs to be combined by the HXAI agent. 

Additionally, aggregating information from different HXAI components, allows the agent to conduct some type of root-cause analysis, identify the underlying causes of potential errors in the workflow, and provide justified and explainable improvements. The agent offers a unique unified explanation framework that enhances \textbf{holistic} transparency and understanding of data analysis workflow, ensuring the development of more reliable, trustworthy, and effective AI systems.

\subsubsection{LLMs as the communication channel}
\label{relatedwork:LLM}

The most critical component of the HXAI agent is the LLM. LLMs are gaining traction in the field of XAI for enhancing interpretability and transparency in AI systems \citep{ABUSITTA2024124710, wu2024usable}. A lot of recent work has suggested that LLMs can act as a communication channel between Domain Experts and AI systems.

For instance, \citep{tornede2023automl} proposes the use of LLMs to explain the AutoML process by leveraging multi-modal LLMs that incorporate training data and feature-importance plots. Similarly, \citep{singh2024rethinking} suggests employing LLMs for data explanations or even as potential replacements for traditional XAI methods. 

Several LLM-powered explanation systems have already been developed \citep{lubos2024llm, yue2023democratizing, ali2023huntgpt}, focusing primarily on explaining model outputs \citep{fredes2024using, baral2024adaptive}. Initial user studies reveal that LLM-generated explanations are often preferred over simple baselines or the absence of explanations altogether \citep{mavrepis2024xai, lubos2024llm, barua2024concept, Slack2023}.
GPT-4 strong explainability performance is attributed to its reasoning capabilities \citep{bordt2024data, aksu2024xforecast}. Finally, TalkToModel offers the closest approach to HXAI by providing data, model, and performance explanations \citep{Slack2023}.

The power of LLMs lies in their ability to dynamically adapt their outputs to individual users and engage in natural, conversational interactions. These are key characteristics of effective explanations. LLMs can tailor multiple explanations to different users, adjust the level of detail to avoid cognitive overload, and support interactive dialogues that enhance mental models of users. Given these capabilities, LLMs have the potential to act as effective substitutes for data analysts in explainability tasks.

\subsubsection{HXAI Agentic Capabilities}
\label{sec:HXAIAgent}
One additional step suggested involves integrating LLM with functionalities and tools that enable them to generate persuasive and correct explanations, without undermining trust and reliability in critical applications \citep{ajwani2024llm}. 
For instance, techniques such as Retrieval Augmented Generation (RAG) have shown promise in enriching LLMs with external context \citep{gao2024retrievalaugmentedgenerationlargelanguage}, by improving the factual grounding of their outputs. 
These enhancements allow LLMs to deliver explanations that go beyond mere suggestions but rather ground truth answers \citep{bordt2024data}. 
When equipped with such capabilities, LLMs evolve into what are often referred to as AI agents. AI agents offer several advancements, including enhanced explainability. The proposed agentic framework extends the HXAI approach in multiple ways:

\noindent \textbf{Real-time information retrieval} By leveraging external tools such as web search, AI agents can stay updated with the latest advancements in the field, dynamically augmenting their knowledge with up-to-date literature on new algorithms and techniques \citep{wang2023interactivenaturallanguageprocessing}. This dynamic augmentation eliminates the need for retraining and allows the system to remain relevant as ML methodologies evolve. For example, if the ML process involves an AutoML system, the AI agent can autonomously access and use the latest documentation.

\noindent \textbf{History of interactions} User's ability improves as they interact more with HXAI systems \citep{muir1987trust, schrills2023users}. By tracking previous exchanges with users through a database, the system can refine its responses over time, offering increasingly personalized and context-aware explanations. This historical awareness allows AI agents to better understand user preferences, knowledge levels, and specific needs, ultimately improving the overall user experience \citep{Chen2024}. Ultimately, these capabilities allow the user to gain trust in the system faster, requiring less interactions. However, evaluating the personalization abilities of AI agents is still an open problem \citep{salemi2023lamp}. 

\noindent \textbf{Dynamic Extraction of Explanations} Beyond static explanations, AI agents can execute real-time reasoning and code-based interpretability methods. Recent advancements have shown that reasoning and coding agents can match the performance of elite human programmers in competitive settings \citep{openai2025competitiveprogramminglargereasoning}. By integrating an AI agent with a code interpreter, users can request on-the-fly execution of HXAI techniques, such as feature importance visualizations. This enables interactive explanations, where users can refine queries, discover deeper aspects of the system's behavior, and receive dynamically generated reports tailored to their needs.

\subsubsection{How HXAI bridges the gap}

Following Figure \ref{fig:trustworthy}, the HXAI architecture is designed to induce Trustworthiness, by aligning user needs to its four dimensions: Stability, Responsibility, Explainability, and Human-centeredness. Table \ref{tab:HXAI-Effective-AIGents} shows how key explanatory requirements of users are met by HXAI, how these map to specific dimensions of trustworthiness, and the corresponding HXAI component that is responsible for addressing them.

\begin{table}[htb]
\small
\centering
\renewcommand{\arraystretch}{3} % Adjust row spacing
\caption{Mapping of HXAI Components to Effective Explanation Qualities and Trustworthiness dimensions. The HXAI Agent provides adaptive and integrated support across the entire ML workflow.}
\label{tab:HXAI-Effective-AIGents}
\begin{tabular}{lll} % Three columns
\toprule
\textbf{Effective explanation qualities} & \textbf{Trustworthiness Dimension} & \textbf{HXAI Component} \\
\midrule

\makecell[l]{Contrastive, Causal\\ \& Truthful} 
& \makecell[l]{Explainability \\ Stability} 
& \makecell[l]{Model Output} \\

\makecell[l]{Trust Calibration} 
& \makecell[l]{Responsibility} 
& \makecell[l]{Model Quality} \\

\makecell[l]{User-Centered} 
& \makecell[l]{Human-Centered} 
& \makecell[l]{HXAI Agent} \\

\makecell[l]{Information-Relevant\\ \& Avoiding\\ Overload} 
& \makecell[l]{Explainability \\ Human-Centered} 
& \makecell[l]{HXAI Agent} \\

\makecell[l]{Supports Human-AI\\ Collaboration} 
& \makecell[l]{Responsibility \\ Human-Centered} 
& \makecell[l]{HXAI Agent} \\

\makecell[l]{Conversational\\ \& Interactive} 
& \makecell[l]{Explainability \\ Human-Centered} 
& \makecell[l]{HXAI Agent} \\

\makecell[l]{Iterative} 
& \makecell[l]{Explainability \\ Human-Centered} 
& \makecell[l]{HXAI Agent} \\

\makecell[l]{Mental Model\\ Enhancement} 
& \makecell[l]{Explainability \\ Human-Centered } 
& \makecell[l]{All} \\

\makecell[l]{Visual} 
& \makecell[l]{Explainability} 
& \makecell[l]{All} \\

\makecell[l]{Holistic} 
& \makecell[l]{Human-Centered\\ Responsibility \\ Stability} 
& \makecell[l]{All} \\

\bottomrule
\end{tabular}

\end{table}

First, HXAI promotes responsibility by offering root-cause analysis and highlighting potential weaknesses in the data analysis process. The user is informed of the analysis configuration before the analysis begins. Additionally, the user is informed about the progression of the learning process, and a variety of metrics that inform the user in-depth about the performance of the ML pipeline. These actions are grounded in the proven principles and research of AutoML and machine learning, which form the foundation upon which the HXAI Agent operates. These capabilities enable users to diagnose errors, understand underlying problems, and make informed corrections, thereby enhancing decision-making across the entire workflow.

Second, explainability is strengthened through the use of contrastive and causal methods, integrated within the Model Output Explainability component. Moreover, the HXAI Agent enhances user comprehension by aggregating information across all HXAI components and selectively presenting only the most relevant insights, thereby preventing information overload. Through continuous interaction with the Agent, via conversational interfaces and iterative exploration, users progressively develop a deeper understanding of the system’s processes and outputs. Additionally, visualizations across all HXAI components further contribute to improved explainability.

Third, Stability, which includes robustness and reproducibility, is achieved by HXAI through its foundation on state-of-the-art XAI methods. Robustness is primarily provided through the Model Quality Explainability component by conveying the reliability of the model’s predictions and the level of uncertainty associated with them. Reproducibility, on the other hand, is improved through transparency around key decisions in the analysis pipeline, including the choice of validation method, evaluation metric, pipeline construction, and hyper-parameter selection. In this way, it increases stability ultimately enhancing trust in the AI system.

Forth, regarding the Human-centeredness, the HXAI Agent naturally fulfills several key criteria for effective explanations, as outlined in Table \ref{tab:HXAI-Effective-AIGents}. It plays a central role in making explainability more interactive, personalized, and user-friendly. By using conversational interfaces, the HXAI Agent can tailor its responses to the user's level of expertise and specific questions, helping to reduce information overload and improve the user’s understanding of how the AI system works. Its support for iterative and holistic explanations ensures calibration of user trust . Through ongoing dialogue, the HXAI Agent enables users to gradually build a clearer mental model of the system. By drawing from all components of the HXAI framework and presenting only what is most relevant to each individual, it creates a user-centered experience that promotes trust in the machine learning workflow. The HXAI Agent also supports responsibility by offering root cause analysis and pointing out possible weaknesses in the ML process. These features help users identify what went wrong and how to fix it, improving decision-making throughout the workflow.

\subsection{HXAI Question Bank}

Building on the foundational work of \citep{QuestionBank} and \citep{ALI2023101805}, a practical way to illustrate how HXAI can help users develop a mental model of the system, and thereby induce their trust, is through a structured question bank. 

An overview of such question bank that aligns with the six core components of HXAI is presented in Table \ref{tab:short-questions}. Specifically, for each functionality of the HXAI components, a representative question is provided that highlights the broad and often ambiguous concerns from one of the user group: Data Analyst (DA), Domain Expert (DE), and Data Scientist (DS).

Each of these questions can be answered through visual and textual explanations, complemented by overviews, summaries, and comparative analyses. Then, for each question, the corresponding methods capable of addressing them, the target users who benefit from the answers, and the specific contributions of these answers to enhancing trustworthiness, is indicated. 
A more comprehensive version of this bank, including detailed, low-level questions can be found in Section \ref{app:qb} of the Appendix.

\begin{table}[htb!]
\footnotesize
\centering
\renewcommand{\arraystretch}{0.5} % Adjust row spacing
\renewcommand{\aboverulesep}{0.3mm} % Adjust below midrule  
\renewcommand{\belowrulesep}{0.3mm} % Adjust eabove midrule
\begin{tabular}{lllll} 
\toprule
\textbf{Component} & \textbf{Functionality} & \textbf{Question} & \textbf{Method} & \textbf{User} \\
\midrule
\multirow{4}{*}{\textbf{Data}} 
& \makecell[l]{Visualization} 
& \makecell[l]{How does the \\ data look like?} 
& \makecell[l]{Histograms \\ Scatter plots \\ t-SNE} 
& \makecell[l]{All}  \\
\cmidrule{2-5}

& \makecell[l]{Summary} 
& \makecell[l]{What are the \\ data characteristics?} 
& \makecell[l]{Meta-Features \\ \citep{RIVOLLI2022108101}} 
& \makecell[l]{DA \\ DS} \\
\cmidrule{2-5}

& \makecell[l]{Relationships} 
& \makecell[l]{How do the \\ data connect to \\ each other?} 
& \makecell[l]{Correlation heatmaps \\ K-means clusters \\ Causal graphs} 
& \makecell[l]{All} \\
\cmidrule{2-5}

& \makecell[l]{Quality} 
& \makecell[l]{Are the data \\ good enough?} 
& \makecell[l]{Outlier detection \\ Data Deduplication} 
& \makecell[l]{DA \\ DS}  \\

\midrule

\multirow{2}{*}{\makecell[l]{\textbf{Analysis} \\\textbf{Setup}}} 
& \makecell[l]{Problem \\ Formulation} 
& \makecell[l]{What is the output?} 
& \makecell[l]{Present the task \\ and the expected \\ model output} 
& \makecell[l]{DE} \\
\cmidrule{2-5}

& \makecell[l]{Optimization} 
& \makecell[l]{How is the model \\ quality measured?} 
& \makecell[l]{Present the selected metric \\ the rationale of selection \\ and explanation of it.} 
& \makecell[l]{All}  \\

\midrule

\multirow{3}{*}{\makecell[l]{\textbf{Model} \\\textbf{Output}}} 
& \makecell[l]{Post-Hoc \\ Example-based} 
& \makecell[l]{What small, \\  plausible changes   to an \\ instance would   lead to  \\ a different prediction?} 
& \makecell[l]{Counterfactuals \\ \citep{wachter2017counterfactual} \\ \citep{mothilal2020explaining}} 
& \makecell[l]{DE \\ DA} \\
\cmidrule{2-5}

& \makecell[l]{Post-Hoc \\ Local-Scope} 
& \makecell[l]{What changes \\ can I make without \\ changing the prediction?} 
& \makecell[l]{Anchors \\ \citep{ribeiro2018anchors}} 
& \makecell[l]{DE \\ DA}  \\
\cmidrule{2-5}

& \makecell[l]{Post-Hoc \\ Global-Scope} 
& \makecell[l]{What is a simple \\ visualization of my model?} 
& \makecell[l]{Surrogate models \\ \citep{bastani2017interpreting} \\ \citep{lakkaraju2017interpretable}} 
& \makecell[l]{DE \\ DA}\\

\midrule

\multirow{4}{*}{\makecell[l]{\textbf{Model} \\\textbf{Quality}}} 
& \makecell[l]{Performance \\ Visualization} 
& \makecell[l]{How does the\\ performance look like?} 
& \makecell[l]{ROC \\ PR Curve \\ Calibration Curve} 
& \makecell[l]{DA\\DS} \\
\cmidrule{2-5}

& \makecell[l]{Error \\ Analysis} 
& \makecell[l]{Where does \\ the model \\ make mistakes?} 
& \makecell[l]{Data Slicing \\ \citep{plumb2022towards} \\ Confusion Matrix} 
& \makecell[l]{DA\\DS} \\
\cmidrule{2-5}

& \makecell[l]{Fairness \\ Assessment} 
& \makecell[l]{Is the model fair?} 
& \makecell[l]{Fairness Metrics \\ \citep{aghaei2019learning} \\ \citep{NIPS2016_9d268236} \\ \citep{dwork2012fairness} \\ \citep{berk2017convex}} 
& \makecell[l]{All} \\
\cmidrule{2-5}

& \makecell[l]{Performance \\ Summary} 
& \makecell[l]{Is the model accurate?} 
& \makecell[l]{Dashboard\\ with metrics} 
& \makecell[l]{All}  \\

\midrule

\multirow{4}{*}{\makecell[l]{\textbf{Learning} \\\textbf{Process}}} 
& \makecell[l]{Optimization \\ Visualization} 
& \makecell[l]{How does the model learn?} 
& \makecell[l]{Learning Curve\\ \citep{park2020hypertendril} \\ F-Anova \\ \citep{hutter2014efficient}} 
& \makecell[l]{DS} \\
\cmidrule{2-5}

& \makecell[l]{Optimization \\ Summary} 
& \makecell[l]{What are the \\ statistics of model \\ learning process?} 
& \makecell[l]{Status Heatmap\\ \citep{sass2022deepcave} \\ Dashboard \\ Text} 
& \makecell[l]{DA\\DS}  \\
\cmidrule{2-5}

& \makecell[l]{Intermediate \\ Visualization} 
& \makecell[l]{How does intermediate\\ steps work?} 
& \makecell[l]{Table View \\ Distribution plots } 
& \makecell[l]{DS}  \\
\cmidrule{2-5}

& \makecell[l]{Pipeline \\ Visualization} 
& \makecell[l]{How does the \\ ML process look like?} 
& \makecell[l]{SSG \\ \citep{zoller2023xautoml}, CPCP \\ \citep{weidele2020autoaiviz} } 
& \makecell[l]{DS}  \\

\midrule
\midrule

\multirow{4}{*}{\makecell[l]{\textbf{HXAI} \\\textbf{Agent}}} 
& \makecell[l]{Data} 
& \makecell[l]{Are Data errors handled?} 
& \makecell[l]{SMOTE \\ \citep{chawla2002smote} \\ AI Explanations} 
& \makecell[l]{DS}  \\
\cmidrule{2-5}

& \makecell[l]{Analysis \\ Setup} 
& \makecell[l]{Is the analysis correctly setup?} 
& \makecell[l]{AI Explanations} 
& \makecell[l]{DA\\DS}  \\
\cmidrule{2-5}

& \makecell[l]{Model \\ Quality} 
& \makecell[l]{What are the model's errors?} 
& \makecell[l]{AI Explanations} 
& \makecell[l]{DA\\DS} \\
\cmidrule{2-5}

& \makecell[l]{Learning \\ Process} 
& \makecell[l]{Can the Learning Process \\ be improved?} 
& \makecell[l]{AI Explanations} 
& \makecell[l]{DS}  \\

\bottomrule
\end{tabular}
\caption{High-level explainability questions across each explainability component of HXAI.}
\label{tab:short-questions}
\end{table}

\clearpage

\section{Proposed HXAI Taxonomy}

In this section, we introduce a taxonomy for HXAI. Starting from its distinct components illustrated in Figure \ref{fig:HXAI-Introduction}, a description of the relevant methods and their categorization is presented for each. Our aim is to offer a structured perspective that goes beyond the traditional focus on model outputs commonly emphasized in the XAI literature, as discussed in Section~\ref{sect:stateofart}. 

\subsection{Data Explainability Taxonomy}

In principle, the quality of machine learning models is intrinsically linked to the quality of the data used during training \citep{budach2022effects, 10.1145/3394486.3406477}. Hence, a comprehensive understanding of data distributions, relationships, and potential issues is crucial for developing robust models. High-quality data ensures that models perform as intended, leading to accurate and reliable outcomes. Conversely, poor-quality data can result in unreliable predictions and biased results, undermining the effectiveness of AI applications \citep{polyzotis2019data}.

\begin{figure}[tb!]
    \centering
    \includegraphics[width=0.9\textwidth]{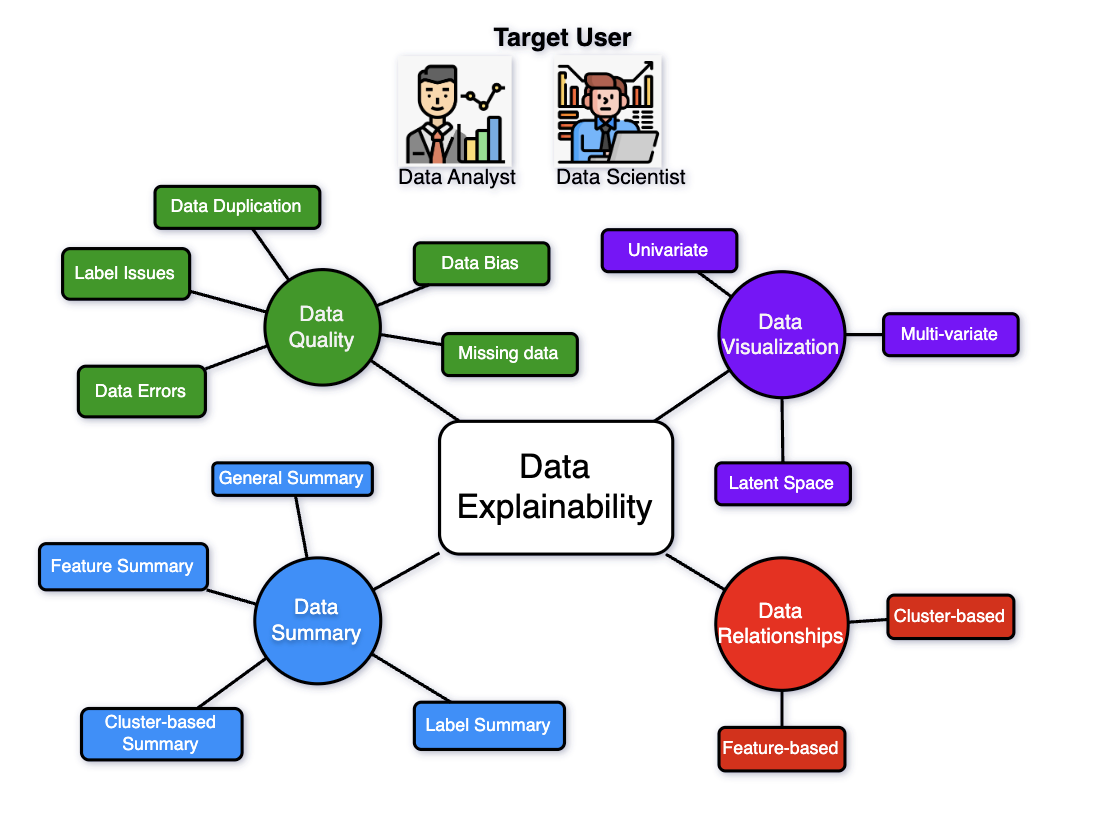} 
    \caption{This figure presents the proposed Data Explainability  Taxonomy.}
    \label{fig:XAI-data} % Reference label
\end{figure}

Contrary to previous reviews on data explainability (see Table \ref{expl-data-tbl}), we propose a detailed taxonomy as seen in Figure \ref{fig:XAI-data}. Data explainability comes before the Analysis Setup process, and provides data understanding to Data Analysts and Data Scientists. Data explainability is divided into four main functionalities as presented in Table \ref{tab:data_explainability_taxonomy}: (a) \textit{Data Visualization}, which focuses on generating visual representations of data; (b) \textit{Data Summary}, which condenses large datasets into manageable overviews; (c) \textit{Data Relationships}, which explores interactions among features, samples, and feature-target pairs; and (d) \textit{Data Quality}, which encompasses methods for assessing and improving data integrity. 
Figure \ref{fig:XAI-data} illustrates the taxonomy for Data Explainability and contains the sub-functionalities of each functionality of Data Explainability.

\begin{table}[htb!]
    \centering
    \caption{Comparison of sub-functionalities covered in prior research on data explainability versus our work.}
    \label{expl-data-tbl}
    \renewcommand{\arraystretch}{1.5} % Adjust row spacing
    \setlength{\tabcolsep}{1pt}       % Adjust column spacing
    \begin{tabular*}{\textwidth}{@{\extracolsep\fill}lcccc}
    \toprule
    & \multicolumn{4}{c}{Data Explainability} \\
    \cmidrule(lr){2-5}
    Paper & Visualization & Summary & Relationships & Quality \\
    \midrule
    \citep{ALI2023101805} 
    & $\blacksquare$ 
    & $\blacksquare$ 
    & 
    &  \\
    \citep{Dwivedi} 
    & $\blacksquare$
    & $\blacksquare$ 
    & 
    & \\
    \citep{burkart}
    & $\blacksquare$ 
    & 
    & 
    & $\blacksquare$ \\
    \citep{Minh2022}
    & 
    & $\blacksquare$
    & 
    & \\
    \hline
    Ours 
    & $\blacksquare$
    & $\blacksquare$
    & $\blacksquare$
    & $\blacksquare$ \\
    \bottomrule
    \end{tabular*}
\end{table}

\subsubsection{Data Summary}

Data summarization provides essential insights into the structure and content of a dataset. While similar to Datasheets \citep{gebru2021datasheets}, which primarily document the data collection process and offer a high-level overview, Data Summary Explainability goes a step further. It focuses on analyzing the actual content of the data through descriptive characteristics known as meta-features \citep{RIVOLLI2022108101}. These meta-features help capture the underlying properties of the dataset to support better understanding, modeling, and decision-making. 

Data Summary can be further organized into four key sub-functionalities:

\noindent \textbf{General Summary}: This sub-functionality includes fundamental data characteristics, referred to as "Simple metafeatures" or "General measures" which aim to provide an overview of the dataset \citep{RIVOLLI2022108101,Castiello}. These measures capture basic properties such as the number of features, instances, target classes and missing values. 
By combining them into ratios, such as the feature-to-instance ratio (\( \frac{\#feat}{\#inst.} \)), it becomes possible to infer dataset traits like dimensionality and potential modeling risks such as overfitting in high-dimensional data \citep{RIVOLLI2022108101}. Additional measures, such as the total number of missing values or the number of features containing missing data, can offer insights into overall data quality.

\noindent \textbf{Feature Summary}: This sub-functionality focuses on summarizing individual features using appropriate statistics, providing information on the data distribution \citep{RIVOLLI2022108101, Castiello}.  A subset of the statistical metafeatures can be adopted for feature summaries. Prominent examples, are central tendency metrics such as mean, geometric mean, standard deviation and variance.

\noindent \textbf{Cluster-Based Summary}: Cluster-based involves summarizing the data in terms of specific data cohorts. For this, the dataset is first partitioned into groups according to feature values, output values or into clusters using an unsupervised method. Summary statistics are, then, calculated for each subgroup/class/cluster. This approach can reveal different patterns within specific segments of the data, such as summaries for features corresponding to distinct values of a categorical variable or target outcome. This is a novel sub-functionality on data summarization, inspired by the problem of data bias and fairness \citep{biasSurvey}.

\noindent \textbf{Label Summary}: Label summarization concentrates on the target variable, detailing its data type and distribution, and imbalance ratio. It uses a subset of "Simple metafeatures" to summarize the information in the label. This summary is important to understand how the outcome is distributed, and if there needs to be a specific handling in the downstream pipeline in case of data imbalance.

\begin{table}[htb!]
\small
\centering
\renewcommand{\arraystretch}{3} % Adjust row spacing
\caption{Data Explainability Taxonomy}
\label{tab:data_explainability_taxonomy}
\begin{tabular}{llll} % Four columns, with wrapping for 2nd, 3rd, 4th
\toprule
\textbf{Functionality} & \textbf{Description} & \textbf{Examples} & \textbf{User} \\
\midrule
\makecell[l]{Data \\ Visualization} 
& \makecell[l]{Graphical representation \\ of data distributions and \\ patterns} 
& \makecell[l]{Histograms \\ Scatter plots \\ t-SNE projections} 
& \makecell[l]{Data Analysts \\ Data Scientists} \\
\midrule

\makecell[l]{Data \\ Summary} 
& \makecell[l]{Condensed overview of \\ dataset characteristics and \\ content} 
& \makecell[l]{Feature counts \\ \# of clusters \\ Imbalance ratios} 
& \makecell[l]{Data Analysts \\ Data Scientists} \\
\midrule

\makecell[l]{Data \\ Relationships} 
& \makecell[l]{Exploration of interactions \\ among features and samples} 
& \makecell[l]{Correlation heatmaps \\ K-means clusters \\ Causal graphs} 
& \makecell[l]{Data Analysts \\ Data Scientists} \\
\midrule

\makecell[l]{Data \\ Quality} 
& \makecell[l]{Assessment \\ of data quality} 
& \makecell[l]{Missing value reports \\ Outlier detection \\ Bias metrics} 
& \makecell[l]{Data Analysts \\ Data Scientists} \\

\bottomrule
\end{tabular}

\end{table}

\subsubsection{Data Visualization}
Visual analysis plays a crucial role in explaining data \citep{Dwivedi}. Visualization allows for faster recognition of patterns and information extraction compared to other formats such as text, numbers, and tables \citep{Gandhi2020}. There is a variety of methods proposed in the past \citep{ajibade2016overview}. Visualization methods can be categorized based on the type of information they present. We identify three main sub-functionalities.

\noindent \textbf{Univariate Visualization}: These techniques focus on a single feature. For continuous features, common plots include Kernel Density Estimate (KDE) plots, and histograms; for categorical variables, count, bar and pie charts are often used.

\noindent \textbf{Multi-variate Visualization}: This sub-functionality encompasses methods that display two or three features simultaneously, allowing for the examination of pairwise or triplet interactions. Examples include scatter plots and heat maps for continuous variables/outcomes, stacked bar plots and clustered bar plots for categorical variables/outcomes, and box plots for numerical-categorical pairs. Additionally, parallel coordinate plots, interaction plots, and color-mapped scatter plots can be employed to illustrate complex interactions \citep{PCPOrder}. Finally, correlation networks can show an overview of the correlation between features in the dataset.

\noindent \textbf{Latent Space Visualization}: When data consist of more than four features, it is often useful to project the data into a lower-dimensional (2D or 3D) space to facilitate interpretation. Prominent methodologies include Principal Component Analysis (PCA), t-distributed Stochastic Neighbor Embedding (t-SNE), Uniform Manifold Approximation and Projection (UMAP), and Autoencoder networks \citep{PCA,tsne,bank2023autoencoders,mcinnes2018umap}. PCA projects the data into a lower-dimension using linear projection, to minimize the reconstruction error. Contrary to that, UMAP, t-sne and autoencoders apply non-linear transformations, allowing them to capture more complex data patterns.

\subsubsection{Data Quality}
Assessing data quality aims to identify issues \textit{before} model training. Data Cleaning, a research field dedicated to detecting and correcting data errors, plays a crucial role in this process \citep{zha2025data}. Studies report that 80\% of the time in a machine learning project is spent on data preparation before training even begins \citep{cote2024data}. While a comprehensive review of Data Cleaning is beyond the scope of this paper, we refer the reader to systematic reviews on the topic \citep{cote2024data}. Although many data quality issues should be handled within the training pipeline to prevent overfitting, an initial quality check provides a quick overview of potential problems. Data quality is often categorized into multiple functionalities \citep{budach2022effects}. In this paper, we propose a Data Quality taxonomy based on five key data issues.

\noindent \textbf{Missing Data}:
This sub-functionality focuses on detecting and addressing missing values, which are prevalent in real-world datasets \citep{Kang2013-ku}. Many machine learning models cannot handle missing data, making it essential to identify and address them. Missing data can occur under three mechanisms: Missing Completely at Random (MCAR), Missing at Random (MAR), and Missing Not at Random (MNAR) \citep{little2019statistical}. Handling strategies vary depending on the type of missingness, yet identifying the exact mechanism remains an open challenge. Little’s test can confirm MCAR, but distinguishing MAR from MNAR is more complex \citep{Little01121988}. Prior research shows that missing data significantly impacts model performance \citep{budach2022effects}. However, simple imputation methods can be surprisingly competitive with more sophisticated techniques, particularly in predictive modeling with complex ML models or AutoML frameworks \citep{10.1145/3643643}.

\noindent \textbf{Data Errors}:
Feature accuracy is another sub-functionality of data quality. Detecting erroneous feature values helps identify potential mislabeling and data inconsistencies. Various methods exist for detecting feature errors, often relying on outlier or anomaly detection techniques. Outliers are data points that significantly deviate from the rest of the dataset \citep{Enderlein1987HawkinsDM}. Model-based approaches such as One-Class Support Vector Machines and Isolation Forests can be used for anomaly detection \citep{liu2008isolation,chen2001one}. Density-based algorithms like Local Outlier Factor (LOF) \citep{knox1998algorithms} and deep-learning methods such as Variational Autoencoders (VAEs) have also been applied to this problem \citep{pmlr-v108-eduardo20a}. PicketNet, a tabular transformer model, has shown promise in distinguishing clean data from corrupted data \citep{liu2022picket}.

\noindent \textbf{Data Bias}:
Fairness is a crucial consideration in data quality. Under-represented groups (data skew) in specific features can introduce bias, leading to unfair model outputs. Detecting these biases before analysis can help mitigate such issues. Data bias is an under-explored topic, it can stem from various sources, including missing data, biased human decisions, and algorithmic biases \citep{pessach2022review}. Notably, some outlier detection methods have been found to exacerbate bias \citep{liu2022picket}. Data fairness issues are mitigated during and post-training by changing the learning process and model selection criteria.

\noindent \textbf{Label Issues}:
This sub-functionality encompasses key label quality issues such as class imbalance and label noise. Even benchmark datasets like ImageNet contain a significant number of mislabeled samples \citep{northcutt2021pervasive}. Label noise can substantially degrade model performance, particularly when more than 20\% of the training data is mislabeled \citep{budach2022effects}. Similarly, class imbalance can introduce bias and should be detected and addressed appropriately in the learning process. Mislabels identification usually require a trained model. Misclassified samples and samples with high uncertainty are potential candidates for label correction \citep{rottmann2023automated,atkinson2020identifying}. Other approaches to find mislabels include influence functions, outlier detection methods, and resampling methods \citep{DBLP:journals/corr/abs-2103-11807,koh2017understanding,10.1111/j.2517-6161.1992.tb01866.x,Bates_2023}. The most usual approach to fixing mislabels is filtering (removing) the mislabeled samples from the training data \citep{brodley1999identifying,kernelbased}.

\noindent \textbf{Data Duplication}:
This sub-functionality focuses on identifying duplicated samples or features and detecting highly correlated features. Addressing data duplication is critical, as it can violate the i.i.d. assumption and negatively impact downstream analysis. Redundant data don't improve ML performance, while increase the complexity of the model \citep{budach2022effects}. Another challenge is detecting label leakage. Features that are highly correlated with the target variable may indicate leakage. However, there are cases where such correlations occur naturally and do not imply leakage. In case of tabular data, simple statistical tests can be used to find similarities. In terms of text data, methods categorized in two approaches exist, token comparison \citep{jin2021deep,wang2020cordel} and latent space comparison \citep{10.1007/s00778-022-00745-1,wang2023sudowoodo}. Both approaches use a similarity function between attributes or records to find matching data \citep{heise2014estimating}. Finally, in image data, the matching happens on the area level or the feature level \citep{encyclopedia5010004,jiang2021review}.

\subsubsection{Data Relationships}
The final functionality of our Data Explainability taxonomy focuses on identifying relationships within the data, which is further divided into two sub-functionalities:

\noindent \textbf{Feature-based Relationships}: This approach examines the interactions between features or between features and the target variable. Techniques such as Pearson and Spearman correlations help detect linear and non-linear relationships, respectively. In addition, feature engineering, such as creating new features by combining or transforming existing ones (e.g., via logarithmic transformation), is considered a part of this analysis. Causal discovery algorithms and automated causal discovery methods \citep{biza2024automatedcausaldiscoverycase, tsamardinos2006max} can further help assess whether observed correlations imply causation. 

\noindent \textbf{Cluster-based Relationships}: Clusteing methods, such as k-means and DBSCAN, analyze relationships at the sample level by identifying similar data points \citep{kmeans, ester1996density}. After clustering, further analysis can be conducted to reveal patterns within each group, such as feature importance and distribution characteristics.

\subsection{Analysis Setup Explainability Taxonomy}

To the best of our knowledge, the concept of Analysis Setup Explainability is first described here. Drawing inspiration from existing commercial AutoML tools, meta-learning, and the principles of AI model cards, it addresses a distinct and under-explored need \citep{mitchell2019model}. For example, AI model cards is a standardized documentation for trained models aiming to improve reasoning around a model’s use, ethical issues, and performance across various settings. Similarly, Analysis Setup Explainability seeks to communicate the reasoning behind the choices, prior to any model being trained in a standardized manner. This added transparency informs users about the selected algorithms, analysis options, and the rationale behind them.

In particular, Analysis Setup Explainability focuses on making transparent the rationale behind how analysis pipelines should be formulated and tested, namely the analysis \textit{configuration} and the configuration selection strategy.  A configuration typically comprises of data preprocessing and data modeling methods, and the parameter values of these methods, referred to as \textit{hyperparameters} \citep{JADBio}. This phase is often the most complex and least automated, typically requiring expert human involvement. The difficulty arises from the need to interpret and integrate both data characteristics (e.g., feature types) and model related metafeatures (e.g., computational constraints) in order to make analysis setup decisions such as which are the appropriate predictive modeling algorithms for a given classification task.

\begin{figure}[htb!]
    \centering
    \includegraphics[width=0.85\textwidth]{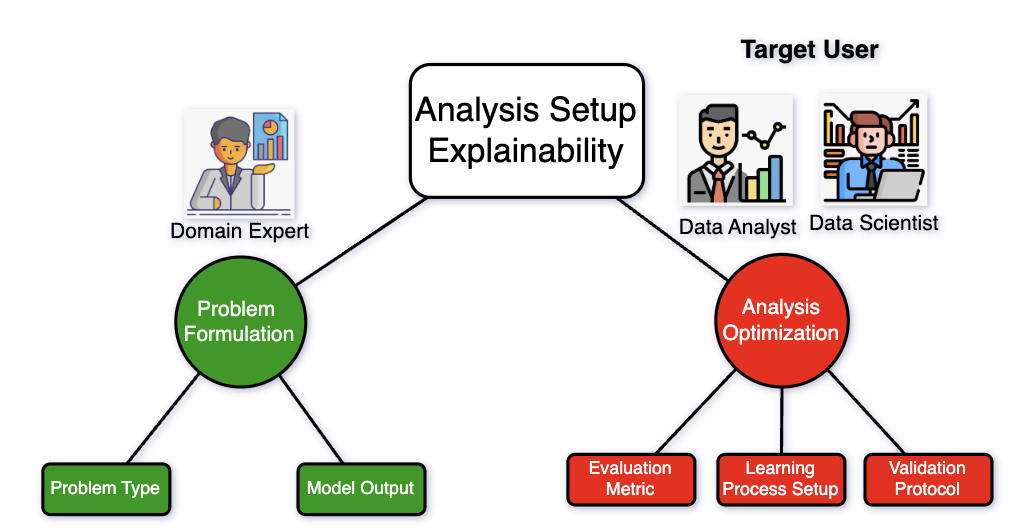} 
    \caption{This figure presents the proposed Analysis Setup Explainability Taxonomy.}
    \label{fig:Analysis-Setup} % Reference label
\end{figure}

We can categorize analysis setup explainability in two functionalities, namely Problem Formulation and Analysis Optimization, as seen in figure \ref{fig:Analysis-Setup}.
Table \ref{tab:analysis_setup_taxonomy} presents an overview of the two functionalities. Specifically, Problem Formulation is straightforward for Data Analysts and Data Scientists and concerns an understanding of the underline task, and the expected model's output. However, Domain Experts, who might lack the necessary expertise, would greatly benefit by understanding how the problem is formulated. Analysis Optimization is important to Data Analysts and Scientists, as it sheds light into the modeling decisions prior to the start of the analysis. Through enhanced understanding of the analysis choices to be made, they can avoid errors, ground the learning process to pre-validated rules, save time, and reduce computational costs associated with re-training.

\noindent \textbf{Problem Formulation} Here, the nature of the addressed problem is explained, specifying the type of analysis task (e.g., classification, regression, or survival analysis). It also clarifies the structure and interpretation of the model's output. Such transparency over problem definition ensures alignment between the user's understanding and the system's objective. This utility is especially targeted to the domain expert that has little-to-none data analysis experience. It allows a preview on what the user is trying to predict, provide alternative options, and what should be expected from the system to output.

\noindent \textbf{Analysis Optimization} This functionality elucidates the decisions made during the training process. It details the chosen validation protocol, including the rationale behind its selection.  Furthermore, it explains the selected evaluation metric, outlining its function and justifying its appropriateness for the task. Finally, it specifies the success criteria and learning process setup, thereby aligning the system's performance goals with the user's objectives.

The selection of the validation protocol, the learning process, the hyper-parameters and the evaluation metric is of paramount importance \citep{bischl2023hyperparameter}. Poor choices in any of these can lead to failed training runs, or worse, a misalignment with business objectives and unrealistic expectations in production environments. Although a wide variety of metrics, validation protocols, and learning processors exist, many of these are not well understood by domain experts and, in some cases, even by data analysts.

\subsubsection{Learning Process Setup}
Selecting an appropriate learning process is complex and non-trivial. It typically requires the expertise of a data scientist or the use of an AutoML approach to ensure a high-quality and well-structured learning pipeline to get high performance on the ML analysis \textit{task} (e.g classification) \citep{JADBio}. 

To address this challenge, meta-learning has recently emerged as a promising solution that leverages historical knowledge from previous ML tasks across diverse algorithms and datasets to provide more accurate selection of analysis settings such as the hyperparameters and predictive algorithms to try \citep{vanschoren2019meta}.
These algorithms can replace traditional rule-based systems and expert-driven recommendations, provide insights into hyperparameter importance, define sensible optimization ranges, and facilitate an effective and efficient optimization search process. Additionally, meta-learning acts as a safeguard, helping domain experts avoid suboptimal configurations during analysis.

First, meta-learning can be based solely on \textbf{model evaluations}. In this approach, \textbf{task-independent recommendations} are employed, meaning they do not require access to any data from the new task, just the rankings of \textit{configurations} on previous evaluations \citep{autosklearn,hutter2019automated,brazdil2003ranking}. Additionally, \textbf{configuration space design} leverages prior evaluations to refine the search space for the optimal hyperparameter values. Surrogate models have been used to determine default hyperparameters for various predictive modeling algorithms and to establish sensible tuning ranges \citep{tunability}. For example, \citep{van2018hyperparameter} applied functional ANOVA (f-ANOVA) to 100 datasets to identify the most influential hyperparameters for three classification models, as well as to derive priors over their value ranges. A heuristic approach that selects the \textit{top k} configurations has also been applied to two machine learning algorithms to determine default hyperparameter values \citep{weerts2020importance}.

Second, meta-learning can be based on \textbf{task properties}. For each task, a set of characteristics, known as meta-features, is extracted \citep{RIVOLLI2022108101}. These meta-features provide informative descriptors of the task. A meta-learner can then be trained using both meta-features and performance evaluations from previous tasks to predict performance on a new task \citep{hutter2019automated}. Various methods exist in this direction, \citep{borboudakis2023meta} reduced the hyper-parameter space using random forests as meta-model. \citep{10.1007/3-540-45372-5_32} built meta-models to predict the performance of ML algorithms \citep{metalearn-predict}.  Alternatively, rather than predicting performance directly, we can utilize \textbf{task similarity}. This involves employing distance-based methods, such as KNN or distance metrics like L1, to measure task similarity based on meta-features \citep{brazdil2003ranking,gomes2012combining}. The identified configurations can then be used to warm-start the optimization process. For a comprehensive review of meta-learning, we refer the reader to \citep{vettoruzzo2024advances,hutter2019automated}.

\begin{table}[htb!]
\small
\centering
\renewcommand{\arraystretch}{3} % Adjust row spacing
\begin{tabular}{llll} % Four columns, with wrapping for 2nd, 3rd, 4th
\toprule
\textbf{Functionality} & \textbf{Description} & \textbf{Examples} & \textbf{User} \\
\midrule
\makecell[l]{Problem \\ Formulation} & \makecell[l]{Explanation of the problem \\ type and model output \\ structure} & \makecell[l]{Classification labels \\ Regression predictions \\ Model Probabilities} & \makecell[l]{Domain experts} \\

\makecell[l]{Analysis \\ Optimization} & \makecell[l]{Details on validation, \\ metrics and learning process} & \makecell[l]{Define metric \\ Select optimizer \\ Select k-fold CV} & \makecell[l]{Data Analysts \\ Data Scientists} \\
\bottomrule
\end{tabular}
\caption{Analysis Setup Explainability Taxonomy}
\label{tab:analysis_setup_taxonomy}
\end{table}

\subsubsection{Validation Protocol}

A validation protocol is essential for estimating a model's performance. One of the most widely used methods is the hold-out approach, though it comes with trade-offs. If the training data are large, the test/validation set will be small, leading to high variance in performance estimates. Conversely, if the training data are small, the model's performance may be underestimated, resulting in a pessimistic evaluation \citep{DontLoseSamples}.

As an alternative, resampling strategies can be employed. These methods repeatedly split the data, and the final performance estimate is obtained by averaging results across splits. The most prominent technique is k-fold cross-validation (k-fold CV). For small datasets, it is often recommended to repeat k-fold CV multiple times to obtain more stable estimates.

The validation protocol setup is typically determined by data analysts and data scientists, following best practices established in scientific literature. However, cross-validation can introduce optimistic bias when numerous machine learning model configurations are tested \citep{DBLP:conf/setn/TsamardinosRL14}. To mitigate this bias, various correction methods have been proposed \citep{tibshirani2009bias,statnikov2005comprehensive,BBC-CV,pmlr-v256-paraschakis24a}.

Additionally, the choice of resampling strategy depends on the nature of the outcome variable. For instance, in analyses involving imbalanced datasets, stratified resampling is beneficial as it ensures that class frequencies remain consistent between training and test sets. Overall, selecting an appropriate validation protocol is a non-trivial process that should be handled by experts. AutoMLs tools automate this process, and abstract it away from the user. For example, JADBio offers AI Decision Support System to select the appropriate resampling protocol for the task and dataset at hand. Similarly, Auto-sklearn 2.0 has introduced an automated approach for selecting validation protocols through policy selection, a form of meta-learning based on meta-features \citep{autosklearn}.

\subsubsection{Evaluation Metric}

In machine learning, a variety of evaluation metrics are used to assess model performance. Different tasks, such as classification and regression, require different metrics. Additionally, certain tasks may necessitate specific metrics due to the nature of the problem and the trade-offs between classes. Selecting the appropriate metric is crucial, as it directly determines the best-performing model. The choice of metric should align with the task at hand, considering both the business objective and the complexities of the data.

Although accuracy is the most commonly used evaluation metric, it is not suitable for tasks with significant class imbalance \citep{hossin2015review}. In such cases, alternative metrics such as the F-score and geometric mean provide more meaningful assessments. More generally, the Receiver Operating Characteristic Area Under the Curve (ROC-AUC) is often recommended as default evaluation metric for classification tasks \citep{10.1007/978-3-031-35314-7_2}.

There are not widely accepted guidelines for selecting the appropriate metric for a given task, and this decision is typically the responsibility of data analysts and data scientists, who ensure that the chosen metric aligns with the overarching business objectives. In that case, the use of domain-specific metrics that are the standards in their fields, could be proposed by the explainability tool such as the Matthews correlation coefficient used by the Food and Drug Administration agency of USA \citep{Chicco2020}.

\subsection{Results Explainability Taxonomy}

In this section, we present two taxonomies for the Results Explainability component one for its model output component and one for the model quality. The field of XAI is mainly concerned with model outputs, while the model quality is tackled in isolation by several fields such as fairness and error analysis. Model outputs explain the rationale behind the model's decisions, while model quality explains the model's performance.

\subsubsection{Model Output Explainability Taxonomy}

\begin{figure}[ht!]
    \centering
    \includegraphics[width=\textwidth]{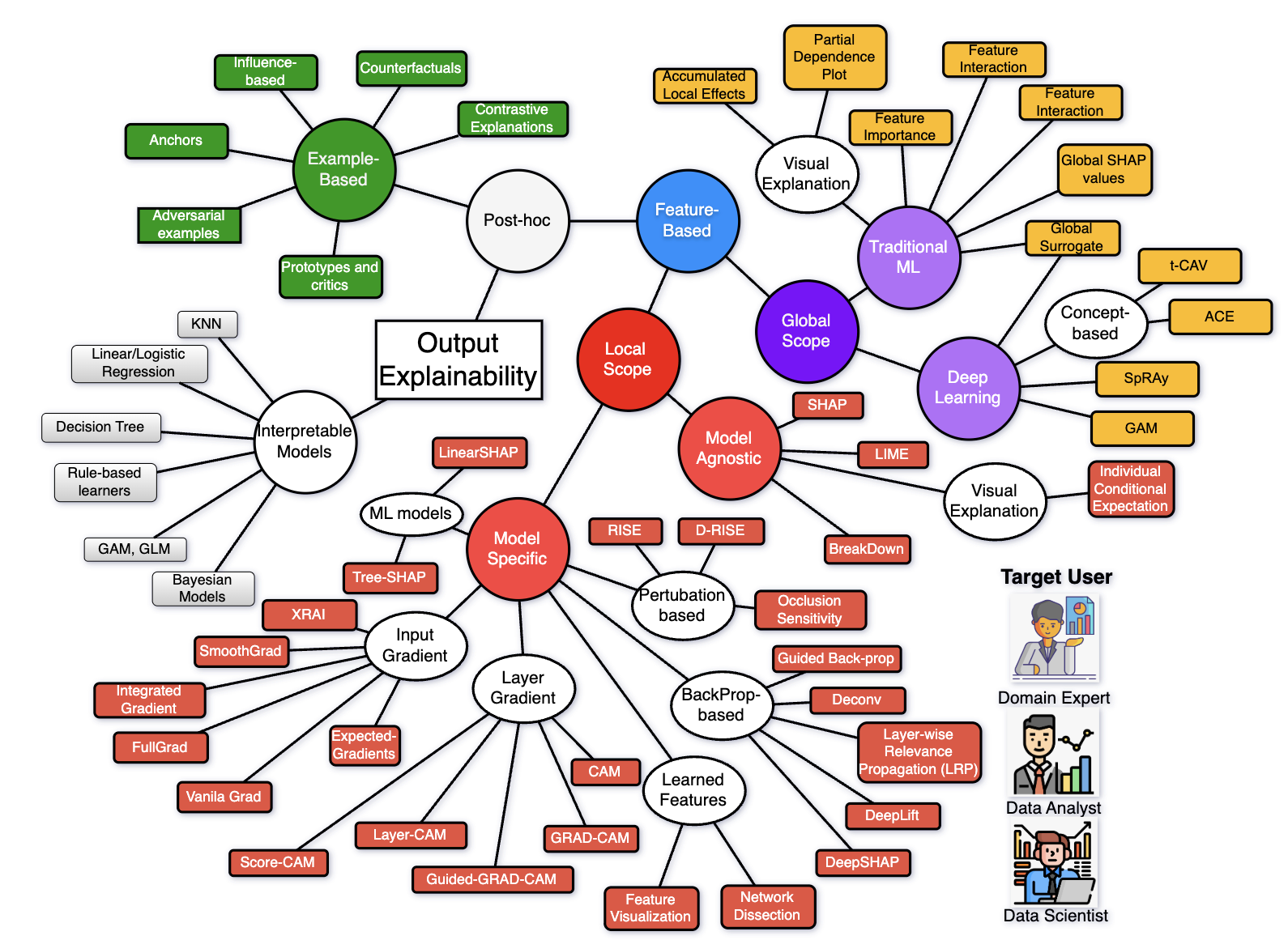} 
    \caption{This figure presents the proposed XAI Taxonomy.}
    \label{fig:XAI-Taxonomy} % Reference label
\end{figure}

Formulating an XAI taxonomy from the relevant literature involves tackling several challenges including misleading nomenclature \citep{speith2022review}. Nevertheless, an actionable and comprehensive taxonomy in the form of a graph that enables intuitive navigation based on specific research needs was developed and is illustrated Figure \ref{fig:XAI-Taxonomy}. In more detail, Table \ref{tab:model_output_taxonomy} presents the two main approaches to Output Explainability: interpretable models and post-hoc methods applied to black-box models \citep{BARREDOARRIETA202082}; their high-level functionalities, their short description, methods examples, and user personas that each pertains to. In brief, Interpretable models are of interest to Domain Experts and Data Analysts, as they are easily understood. However the lack of accuracy in complex problems leads to post-hoc explanations of black-box models. Post-hoc Example-based explanations are typically used to explain models through specific samples, and are of interest to each user category. Post-hoc Local-Scope questions can allow insights into specific predictions. Finally Post-hoc Global Scope methods provide a general overview of the model's decisions across the whole dataset which benefits all the user categories. These functionalities and sub-functionalities of methods and approaches are presented in more detail next.

\begin{table}[htb]
\small
\centering
\renewcommand{\arraystretch}{3} % Adjust row spacing
\begin{tabular}{llll} % Four columns, with wrapping for 2nd, 3rd, 4th
\toprule
\textbf{Functionality} & \textbf{Description} & \textbf{Examples} & \textbf{User} \\
\midrule
\makecell[l]{Interpretable \\ Models} & \makecell[l]{Transparent models \\ designed for direct \\ understanding} & \makecell[l]{Linear Regression \\ Decision Trees \\ RuleFit} & \makecell[l]{Data Analysts \\ Domain experts} \\

\makecell[l]{Post-Hoc \\ Example-Based} & \makecell[l]{Explanations via sample \\ alterations or similar \\ samples} & \makecell[l]{Counterfactuals \\ Anchors \\ Prototypes} & \makecell[l]{All} \\

\makecell[l]{Post-Hoc \\ Local Scope} & \makecell[l]{Feature importance for \\ individual predictions, \\ specific or agnostic} & \makecell[l]{SHAP \\ LIME \\ GradCAM} & \makecell[l]{All} \\

\makecell[l]{Post-Hoc \\ Global Scope} & \makecell[l]{Overall model behavior \\ across the dataset, for \\ ML or deep learning} & \makecell[l]{PDP \\ TCAV \\ Global Surrogate} & \makecell[l]{All} \\
\bottomrule
\end{tabular}
\caption{Model Output Explainability Taxonomy}
\label{tab:model_output_taxonomy}
\end{table}

\textbf{Interpretable models} are typically classic statistical and machine learning methods, such as Linear/Logistic Regression, Generalized Linear Models (GLMs), Generalized Additive Models (GAMs), k-Nearest Neighbors (KNN), Bayesian Models, Decision Trees, and rule-based learners like RuleFit. The majority of XAI literature is focusing on post-hoc explainability, as black-box models often outperform their interpretable counterparts \citep{Darpa_Gunning, BARREDOARRIETA202082}. Post-hoc methods can be broadly categorized into example-based and feature-based approaches. 

\textbf{Example-based methods} explain a model’s behavior by altering samples or identifying similar samples to provide insights into its decisions \citep{Dwivedi, molnar2022}. Typical methods in this direction include Counterfactual Explanations, Scoped Rules (Anchors), Prototypes and Criticisms, Adversarial examples, Influential Instances, and the Contrastive Explanations Method (CEM). 
Counterfactual Explanations explain a prediction by identifying the minimum change to feature values required to change the prediction \citep{wachter2017counterfactual}. We note that counterfactual explanations don't suffice for causation, and are different from causal counterfactuals \citep{Baron2023}.
Scoped Rules (Anchors) find if-then rules \citep{ribeiro2018anchors}. 
Prototypes and Criticisms involve representative data points of a data distribution (prototypes), and non-representative instances (criticisms) \citep{kim2016examples}. 
Adversarial methods try to trick the model into making incorrect predictions \citep{szegedy2013intriguing}. 
Influential Instances identify how much influence a training sample had on a prediction \citep{koh2017understanding,krishnan2017palm}. 
CEM explains the model decision by contrasting what should and should not be present for the prediction \citep{dhurandhar2018explanations}.

\textbf{Feature-based methods} focus on quantifying the importance of features in the model's predictions \citep{9050829}. These methods operate at different scopes: the local scope, which explains individual predictions, and the global scope, which examines the model's behavior across the entire dataset \citep{molnar2022}. Local methods can be further divided into model-specific approaches, which are tailored to specific types of models such as deep neural networks, and model-agnostic approaches, which can be applied to any model by analyzing its inputs and outputs \citep{Schwalbe2024}. 

\textbf{Model-specific methods} with local scope can be categorized into six groups:

\noindent \textbf{Input Gradients} leverage the gradient of the model’s output with respect to its input features to assess feature importance. Among these methods, Vanilla Grad was the first proposed explainability technique based on the gradient of input features \citep{simonyan2013deep}, though it suffers from the gradient saturation problem. Integrated Grad improves upon Vanilla Grad by addressing the gradient saturation issue, explaining the difference between the model's prediction and a reference sample \citep{sundararajan2017axiomatic}. SmoothGrad offers an enhancement to Vanilla Grad by reducing noise in gradient explanations \citep{smilkov2017smoothgrad}. FullGrad combines input gradients with feature-level gradients\citep{srinivas2019full}. XRAI, a region gradient-based method built on Integrated Gradients, first over-segments an image, applies integrated gradients to each segment to obtain importance scores, and then merges smaller areas into larger segments based on these scores \citep{kapishnikov2019xrai}. Finally, Expected Gradients, an extension of Integrated Gradients, uses batch training to regularize expected gradients, further refining the approach \citep{erion2021improving}.

\noindent \textbf{Layer Gradient} methods focus on interpreting the contributions of specific layers, targeting feature maps to produce visual explanations. Among these, Class Activation Mapping (CAM) utilizes the final convolutional layer of a network to localize regions of interest in the input image \citep{zhou2016learning}. GradCAM, an enhanced version of CAM, leverages gradient information from the last convolutional layer to compute feature importance \citep{selvaraju2017grad}. Guided Grad-CAM combines Grad-CAM with guided backpropagation to produce fine-grained explanations \citep{selvaraju2017grad}. Layer-CAM rethinks the relationships between the feature maps and their corresponding gradients \citep{jiang2021layercam}. Finally, Score-CAM eliminates the dependence on gradients by directly using activation maps and their contributions to the model’s output through a forward pass \citep{wang2020score}.

\noindent \textbf{Backpropagation-based} methods explain model outputs by analyzing how information moves backward through the neural network during backpropagation. Among these, Guided Backpropagation visualizes the importance of input pixels to a prediction by combining backpropagation and negative gradient filtering \citep{springenberg2014striving}. Deconvolution reverses the convolutional neural network operations to map the feature activations to the input space \citep{zeiler2014visualizing}. Layer-wise Relevance Propagation (LRP) decomposes the output prediction of a neural network into contributions from its input features \citep{LRP}. DeepLift computes importance scores by explaining the difference of the output from some baseline output, in terms of the differences in their inputs \citep{li2021deep}. Finally, DeepSHAP enhances the DeepLIFT algorithm by integrating SHAP values \citep{scott2017unified}.

\noindent \textbf{Input perturbation} methods assess the importance of input features by perturbing the input and observing the impact on the output. Among these, Randomized Input Sampling for Explanation (RISE) randomly masks parts of an image and aggregates the effect of the masking on predictions \citep{petsiuk2018rise}. D-RISE, an extension of RISE, dynamically adapts the mask size and shape based on the model’s focus \citep{petsiuk2021black}. Occlusion sensitivity systematically occludes parts of the input and observes how the prediction changes \citep{zeiler2014visualizing}.

\noindent \textbf{Learned features} in convolutional neural networks involve the learning of abstract features and concepts from raw image pixels. Feature Visualization visualizes these learned features by activation maximization through finding appropriate input patterns \citep{olah2017feature}. Network Dissection attempts to map human concepts to what the neural network has learned \citep{bau2017network}.

\noindent \textbf{ML-Model} methods include SHAP variations specifically designed for classic ML algorithms. TreeSHAP is a variant of SHAP tailored for tree-based models like random forests or gradient boosting machines \citep{lundberg2020local}. LinearSHAP is a SHAP variant developed for linear models \citep{scott2017unified}.

In the case of \textbf{model-agnostic methods with local scope}, four prominent methods exist that are applicable to all kinds of models. Local surrogate models (LIME) explain a prediction by approximating the black box model through an interpretable surrogate model using perturbations \citep{LIME}. Shapley values (SHAP) serve as an attribution method that assigns the prediction to individual features \citep{scott2017unified}. BreakDown directly calculates variable attributions for a selected observation using a greedy approach without relying on surrogate models \citep{breakdown}. Individual Conditional Expectation (ICE) acts as a visualization tool that describes how changing a feature's value alters the model's prediction \citep{ICE}.

\textbf{Global methods} can provide explanations for sub-groups within the dataset. Global scoped methods are typically divided into those designed for traditional machine learning models and those focused on deep learning-based models. 

\noindent \textbf{Global scoped methods for classic ML models} include several approaches. Partial Dependence Plots (PDP) visualize the average effect of a feature on the model's predictions \citep{friedman2001greedy}. Accumulated Local Effects (ALE) visualize the effect of a feature by aggregating its local effects, serving as an improvement over PDP by handling feature dependencies \citep{apley2020visualizing}. Permutation Feature Importance measures the importance of features by assessing the decrease in the model's performance when the feature’s values are randomly permuted \citep{fisher2019all}. Feature Interaction methods measure the combined effect of features on the model’s predictions, with examples like the H-statistic \citep{friedman2008predictive}. Global Surrogate methods, which are interpretable (e.g., decision trees), attempt to approximate the black-box model by training on its predictions \citep{bastani2017interpreting,lakkaraju2017interpretable}. Finally, Global SHAP Values identify the contribution of features by

\noindent  \textbf{Global scoped methods for deep learning models} encompass several techniques. Detecting Concepts involves identifying any user-defined concept in the latent space of a model, prominent methods include Testing with Concept Activation Vectors (TCAV) \citep{kim2018interpretability} and Automated Concept-based Explanation (ACE) \citep{ghorbani2019towards}. 
Spectral Relevance Analysis (SpRAy) performs spectral clustering on local LRP explanations \citep{Lapuschkin2019}. 
Global Attribution Mapping (GAM) groups similar local explanations to form subpopulations feature importance \citep{ibrahim2019global}. 
Global Surrogate methods approximate the predictions of the neural network through an interpretable model, prominent methods in this direction include DeepRed \citep{zilke2016deepred}, ANN-DT \citep{schmitz1999ann}, and TREPAN \citep{craven1995extracting}.

\subsubsection{Model Quality Explainability Taxonomy}

\begin{figure}[ht!]
    \centering
    \includegraphics[width=0.8\textwidth]{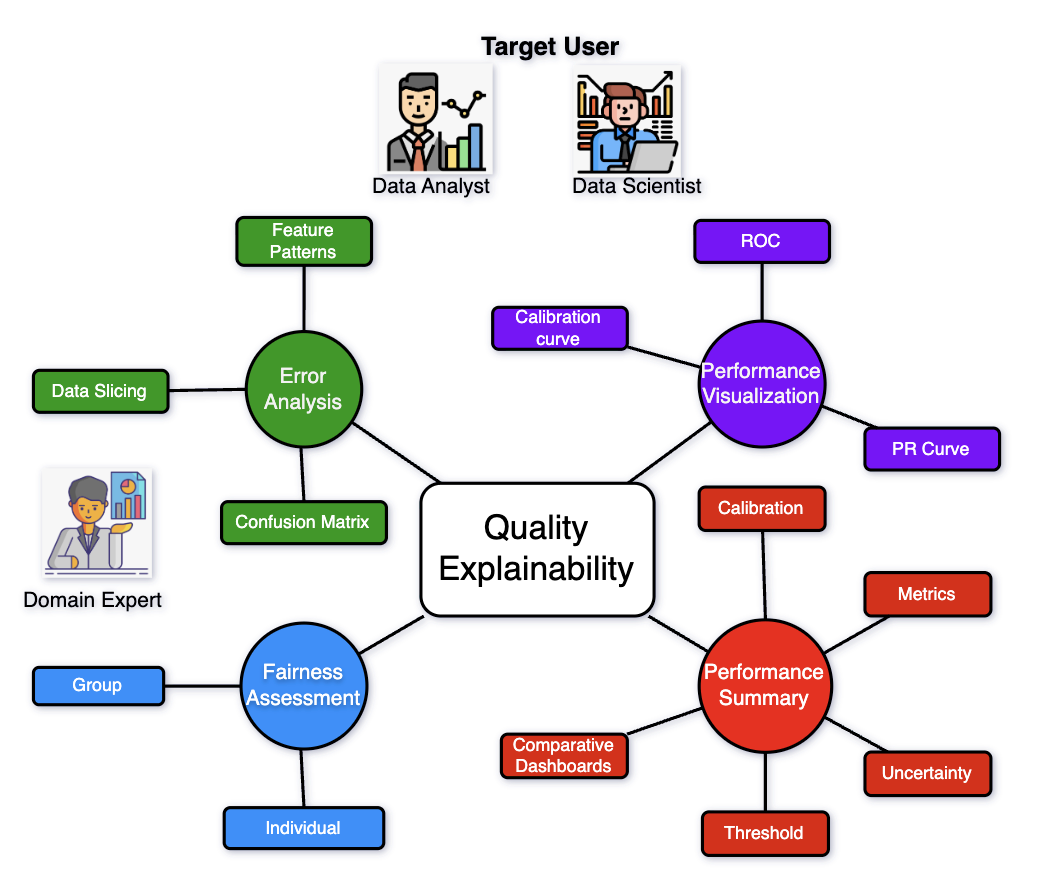} 
    \caption{This figure presents the proposed Model Quality Explainability  Taxonomy.}
    \label{fig:quality} % Reference label
\end{figure}

Understanding and interpreting model quality is essential for building trust and ensuring reliable decision-making in machine learning systems. To address this, we introduce a  taxonomy that organizes model quality explainability into four distinct functionalities: performance visualization, error analysis, fairness assessment, and performance summary as illustrated in Figure \ref{fig:quality}. To the best of our knowledge, this taxonomy is novel, as its sub-fields have traditionally been studied independently.

Performance visualization presents the model’s performance in a visual format. 
As indicated in Table \ref{tab:model_quality_taxonomy} its role is to aid data analysts and scientists. Error analysis helps all users identify model limitations, preventing misuse and facilitating debugging. Fairness assessment detects and contextualizes bias, supporting domain experts in real-world applications and data scientists in mitigating bias during training. Performance summary provides a concise evaluation of model performance, enabling data analysts and scientists to quickly validate results across different metrics and thresholds. Each of these is described in more detail below.

\noindent \textbf{Performance Visualization}
This functionality encompasses graphical tools such as the Receiver Operating Characteristic (ROC) curve and the Precision-Recall (PR) curve, which illustrate model performance and enable users to quickly glean critical insights. Specifically, the ROC curve, a widely adopted visualization, plots the true positive rate (TPR) against the false positive rate (FPR) across various decision thresholds \citep{FAWCETT2006861}. However, in scenarios with highly imbalanced datasets, the ROC curve may present a biased perspective \citep{miller2024review}. 
For such tasks, the PR curve is often more informative, depicting precision versus recall to better reflect performance under imbalance. Additionally, a calibration curve (or reliability diagram) provides a visual representation of the model’s probability estimates, illustrating how closely predicted probabilities align with empirical outcomes. This plot reveals whether the model exhibits overconfidence or underconfidence in its predictions \citep{wang2023calibration}. While visualizations generally enhance interpretability, these performance graphs typically require machine learning expertise for full comprehension, limiting their accessibility to non-specialists.

\noindent \textbf{Error Analysis} This functionality includes methods for analyzing the errors of the model. Error analysis in machine learning has gained significant attention in recent years as a crucial step to diagnose and improve model performance by analyzing mistakes that aggregate metrics often hide. 
This section provides a comprehensive overview, addressing under-performing metrics, common error types, patterns in misclassified samples, feature ranges with poor performance, feature categories linked to higher error rates, and clustering errors into groups. 
This functionality appeals to all user categories. Data scientists can use Error Analysis to debug the ML process and improve the performance of the model. Similarly, domain experts and data analysts can detect cases where they shouldn't rely on the model.  

Overall scores, such as accuracy, ROC-AUC and F1-score, don’t reveal the key features or patterns behind a model’s mistakes \citep{singla2021understanding,van2023barriers}. They also hide the types of errors the model makes. Common errors, like false positives (FP) and false negatives (FN), can be shown with a confusion matrix \citep{talbot2009ensemblematrix}. This tool highlights where the model gets things wrong, helping users decide which errors to focus on. Plus, to spot patterns in these mistakes, you can look at the features causing them. A feature importance method can be applied to misclassified samples or groups of samples to give insights into which features lead to misclassifications \citep{ResponsibleAIToolbox}.

Commercial models for gender classification often struggle with certain subgroups in data \citep{buolamwini2018gender}. For tabular data, we can identify these weaker groups, or "slices," by using clustering algorithms, picking specific feature values by hand \citep{chung2019automated}, or applying a decision tree (DT) to spot slices with higher error rates \citep{chen2004failure,ResponsibleAIToolbox}. However, finding these underperforming slices in data like images or audio isn’t so straightforward \citep{eyuboglu2022domino}. 

To tackle this challenge, a growing field called "Data Slicing" has emerged, focusing on ways to pinpoint data subsets where models don’t perform well \citep{johnson2023does}. The goal is to create slices that meet two key needs: (a) they contain samples with lots of errors, and (b) they group together samples that make sense as a set. Several approaches have been developed. Some involve interactive tools to explore how models behave across different slices, such as Zeno \citep{cabrera2023zeno}, Errudite \citep{wu2019errudite}, Kaleidoscope \citep{suresh2023kaleidoscope}, and SliceFinder \citep{chung2019automated}. Others rely on representing data through embeddings, like Domino \citep{eyuboglu2022domino} and Spotlight \citep{d2022spotlight}, or use techniques like dimensionality reduction with PlaneSpot \citep{plumb2022towards} and linear algebra with Sliceline \citep{sagadeeva2021sliceline}.

\noindent \textbf{Fairness Assessment} Fairness has emerged as a critical area of study in AI, with sources of unfairness traced to either data or algorithmic design \citep{biasSurvey}. Fairness is typically evaluated across two dimensions \citep{kheya2024pursuit}: individual fairness, which assesses whether a model provides similar predictions for similar individuals, and group fairness, which examines equitable treatment across distinct demographic groups. 

The bulk of fairness research focuses on classification and regression tasks, for which a variety of metrics have been developed. For group fairness in classification, prominent metrics include disparate treatment and disparate impact, as proposed by \citep{aghaei2019learning}, alongside equalized odds, equal opportunity \citep{NIPS2016_9d268236}, and statistical parity \citep{dwork2012fairness}. In regression, the "price of fairness" (POF) stands out as a key metric bridging individual and group fairness \citep{berk2017convex}, complemented by statistical parity measures tailored to regression \citep{agarwal2019fair}. Similarly, \citep{calders2013controlling} introduced two fairness metrics for regression, equal means and balanced residuals. Additionally, traditional machine learning metrics have been adapted to assess fairness by comparing performance differences between sensitive and non-sensitive groups \citep{das2021fairness}. 

Despite significant progress, selecting an appropriate fairness metric remains an unresolved challenge \citep{weerts2024can}. Fairness research is also challenged with practical trade-offs, such as the trade-off between fairness and accuracy \citep{10.1145/3368089.3409697} and the difficulty of jointly ensuring fairness and calibration \citep{pleiss2017fairness}. Fairness assessments can further incorporate calibration errors per group, a concept termed Equal Calibration \citep{doi:10.1089/big.2016.0047,kheya2024pursuit}. Statistical tests, often paired with k-fold cross-validation, have also been proposed to evaluate model fairness \citep{uddin2024novel}.

\begin{table}[htb]
\small
\centering
\renewcommand{\arraystretch}{3} % Adjust row spacing
\begin{tabular}{llll} % Four columns, with wrapping for 2nd, 3rd, 4th
\toprule
\textbf{Functionality} & \textbf{Description} & \textbf{Examples} & \textbf{User} \\
\midrule
\makecell[l]{Performance \\ Visualization} & \makecell[l]{Graphical representation \\ of model performance} & \makecell[l]{ROC curve \\ PR curves \\ Calibration curves} & \makecell[l]{Data analysts \\ Data scientists} \\

\makecell[l]{Error \\ Analysis} & \makecell[l]{Understand and \\ analyze model errors} & \makecell[l]{Confusion matrices \\ Error rates per category} & \makecell[l]{All} \\ 

\makecell[l]{Fairness \\ Assessment} & \makecell[l]{Evaluation of fairness \\ across groups} & \makecell[l]{Fairness metrics} & \makecell[l]{Domain experts \\ Data scientists} \\

\makecell[l]{Performance \\ Summary} & \makecell[l]{Comprehensive overview \\ of model performance \\ and metrics} & \makecell[l]{Metrics \\ Confidence intervals \\ Model comparison} & \makecell[l]{Data Analysts \\ Data Scientists} \\ 
\bottomrule
\end{tabular}
\caption{Model Quality Explainability Taxonomy}
\label{tab:model_quality_taxonomy}
\end{table}

\noindent \textbf{Performance Summary} This functionality outlines methods for summarizing a model’s performance. A thorough performance summary combines various metrics and approaches to provide a clear overall view of how the model performs.

First, the Performance Summary covers key metrics for classification and regression. For classification, common metrics include Accuracy, AUC, and F1, while regression relies on metrics like R-Squared, Root Mean Square Error (RMSE), and Mean Square Error (MSE), each highlighting different aspects of model performance \citep{10.1007/978-3-031-35314-7_2}. For instance, the F1 score works well in imbalanced tasks where the positive class matters more, but it can be hard for non-experts to understand. Accuracy, on the other hand, is simple and widely used, yet it falls short for problems like imbalanced datasets \citep{miller2024review}. It’s worth noting that most machine learning metrics are not easy to grasp for domain experts with limited ML knowledge.

To assess the uncertainty in performance estimates and evaluate model robustness, confidence intervals (CIs) can be computed using cross-validation or bootstrapping \citep{Bates_2023,BBC-CV,willmott1985statistics}. Overlapping confidence intervals suggest that two models perform similarly, while statistical significance tests can help detect meaningful differences in performance \citep{demvsar2006statistical,dror2018hitchhiker}. 

Selecting an appropriate statistical test is nontrivial, as multiple options exist for comparing model performance, including the modified paired t-test, analysis of variance (ANOVA), and McNemar’s test \citep{6790639,demvsar2006statistical}. When conducting multiple statistical tests, applying a correction method such as the Bonferroni correction \citep{neyman1928use} is crucial to mitigate false positives from multiple comparisons \citep{dror2017replicability,farcomeni2008review,10.5555/39892}.

Threshold optimization is a crucial aspect of performance evaluation in classification. Many machine learning models output probabilities rather than discrete class labels, requiring the selection of an appropriate threshold for decision-making. The default threshold of 0.5 is rarely optimal, as the best threshold varies depending on the data, task, and evaluation metric. Common approaches for identifying optimal thresholds include analyzing the Receiver Operating Characteristic (ROC) curve and the Precision-Recall (PR) curve. More advanced methods, such as mixed-integer linear programming \citep{koseoglu2024otlp} and brute-force search \citep{zou2016finding}, can also be employed.

Another critical aspect of performance evaluation is model calibration. Ideally, a model should not only achieve high predictive accuracy but also produce well-calibrated probability estimates \citep{wang2023calibration}. A comprehensive calibration assessment typically includes metrics such as Expected Calibration Error (ECE), Maximum Calibration Error (MCE), Classwise ECE (CECE), and the Brier Score \citep{naeini2015obtaining, kull2019beyond, calibration2}.

Finally, performance evaluation often involves \textit{comparative analysis}. Understanding a model’s effectiveness frequently requires benchmarking against other models—such as comparing a new model to a previous version or assessing different configurations from an AutoML process. AutoML platforms typically provide dashboards with leaderboards and detailed performance summaries, enabling users to systematically compare models and identify their relative strengths and weaknesses.

\subsection{Learning Process Explainability Taxonomy}
\label{relatedwork:ixAutoML}

% Specify cash problem, hyper-parameter tuning is needed. etc

\begin{figure}[ht!]
    \centering
    \includegraphics[width=0.8\textwidth]{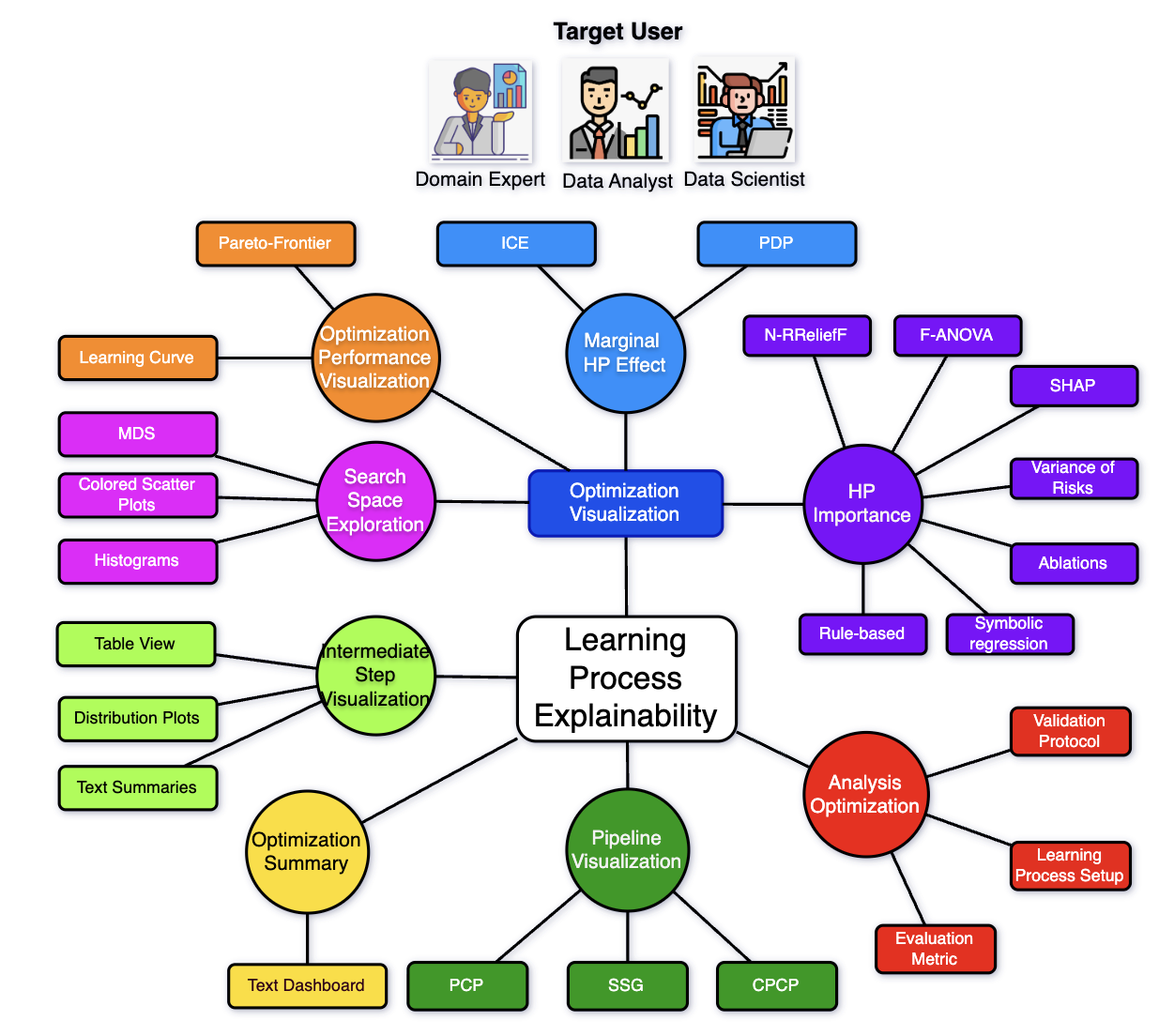} 
    \caption{This figure presents the proposed learning process explainability Taxonomy.}
    \label{fig:learningprocess-Taxonomy} % Reference label
\end{figure}

The goal of learning process explainability is to provide insights into how a machine learning model is trained. Increased transparency in this process is crucial for ML experts and developers, as it enables model debugging and improvement. To address this need, the field of Interactive and Explainable AutoML (ixAutoML) has recently gained traction \citep{humancenteredautoml}, building upon the principles of human-centered AutoML \citep{pfisterer2019towards}. ixAutoML goes beyond explaining model outputs, extending to the interpretability of the learning algorithm’s performance. The significance of learning process explainability is multifaceted. Figure \ref{fig:learningprocess-Taxonomy} presents the main pillars of Learning process explainability. Table \ref{tab:learning_process_taxonomy} provides a short overview of each functionality of Learning process explainability. Functionalities included in learning process explainability are ones concerned with visualization of the optimization process helping explain how the optimization progresses, pipeline visualization that provides an overview of the steps and options of the ML process, optimization summary that provides a quick summarization of key learning process details, and finally the intermediate visualization that shows the output and statistics of the intermediate pipeline steps before data are fed to the predictive  model. Mostly, data scientists are concerned with this component, as they look to debug and improve the learning process. However, data analysts can benefit from optimization summary and pipeline visualization to verify that the ML process is correct.

\subsubsection{Optimization Visualization}
In terms of \textbf{Optimization visualization}, four main approaches exist; (a) visualizing the hyper-parameter importance, taking into account all of the hyper-parameters, (b) visualizing the individual (marginal) effect of hyper-parameters, (c) the search space exploration, and (d) the visualization of the optimization performance. The explanation methods discussed in this functionality are generally applicable across a wide range of optimization techniques. In cases involving specific procedures such as meta-heuristic optimization, a more detailed overview can be found in the systematic review by \citep{almeida2025systematic}.

\noindent \textbf{Hyper-Parameter Importance}
This sub-functionality focuses on the role of hyperparameters in model training.  Identifying the most influential hyperparameters enables the pruning of less important ones, allowing the model to be trained within a smaller, more fine-grained search space. This process helps data scientists and analysts maximize model performance while significantly reducing computational costs. Several prominent methods have been developed to assess hyperparameter importance, including functional ANOVA (F-ANOVA) \citep{hutter2014efficient}, surrogate-based ablations \citep{biedenkapp2017efficient}, SHAP \citep{mu2024shrinkhpo}, forward selection \citep{forwardselection}, rule-based systems \citep{chakraborty2024explainable}, N-RReliefF \citep{Sun2019HyperparameterIA}, variance of risks \citep{jin2022hyperparameter}, and even symbolic regression \citep{segel2023symbolic}.

\noindent \textbf{Marginal Effects of Hyperparameters}
This sub-functionality examines the marginal effect of tuned hyperparameters during the learning process, on the predictive performance of machine learning models. Understanding the marginal effect of individual hyperparameters on model performance allows for dynamic adaptation of the configuration space once a sufficient confidence threshold is reached \citep{moosbauer2022improving}. One approach involves adapting Partial Dependence (PDP) plots to visualize the impact of hyperparameters on model performance \citep{moosbauer2021explaining, moosbauer2022improving}. Similarly, Individual Conditional Expectation (ICE) plots have been used alongside PDPs to provide a more detailed view of hyperparameter effects \citep{zoller2023xautoml}.

\noindent \textbf{Search space exploration}
To assess whether the sampling strategy and the number of tested configurations during the learning process are adequate, a search space coverage methodology has been proposed. One approach utilizes multi-dimensional scaling (MDS) to generate a 2D projection of the configuration space, allowing for a visual representation of search-space coverage \citep{sass2022deepcave, zoller2023xautoml}. This visualization often includes a heatmap that highlights estimated performance. Additionally, search space exploration over time has been depicted using colored scatter plots, which illustrate how different configurations are explored throughout the learning process \citep{zoller2023xautoml}. In case of algorithm performance, histograms have been viewed to show an overview of the configurations performance of specific algorithms \citep{wang2019atmseer}.

\noindent \textbf{Optimization Performance Visualization}
This sub-functionality includes methods that show the performance of the system. Learning curves are adopted, to show the progression of the systems predictive performance over time \citep{park2020hypertendril}. The learning curve allows the data scientists to find errors or breakthrough points by finding plateau or sudden drops in the curve, and matching them to specific search regions. Similarly, pareto-frontier can provide an overview of the optimal combination of two metrics \citep{sass2022deepcave}. In some experiments, we may optimize two objectives at the same time, instead of a single one \citep{Sharma2022}. Through the pareto-frontier plot, a domain experts can visually understand the trade-off between two metrics for their model. Similarly, data analysts and scientists can use the trade-offs to improve the optimization loop (e.g objective function refinement), improving the final model performance.

\subsubsection{Pipeline Visualization}

Another functionality of learning process explainability is \textbf{pipeline visualization and overview}, which involves generating visual representations of the machine learning pipeline. ML pipelines typically consist of multiple steps and hyperparameters, making visualization challenging due to their complexity and conditional dependencies. As mentioned before, a configuration is a single instance of the ML pipeline.
Initially, parallel coordinate plots (PCP) were used to visualize all of the configurations \citep{wang2019atmseer}, but they lacked the ability to represent conditional relationships. To address this limitation, Conditional Parallel Coordinate Plots (CPCP) were introduced and later adopted in subsequent studies for configuration visualization, as they naturally support conditional dependencies \citep{Weidele2019ConditionalPC, weidele2020autoaiviz, zoller2023xautoml, park2020hypertendril}. 
Additionally, to provide a higher-level overview of the pipeline structure, a structure search graph (SSG) has been proposed. This method aggregates all the different configurations explored during the search process into a single graph, offering a comprehensive overview of the algorithms and hyperparameters considered \citep{zoller2023xautoml}. In this case, many ixAutoML tools offer dashboards, that allow side by side comparison of individual pipelines/configurations.

\subsubsection{Optimization Summary}

The Optimization Summary functionality contains dashboards that allow easy digestion of information concerning the optimization. Simple visualizations are used to convey important information such as the best model's type, hyper-parameters, and performance through text \citep{wang2019atmseer,zoller2023xautoml}. These dashboards, also present the objective metric, potential secondary optional metric to optimize, and information about failed runs \citep{sass2022deepcave,zoller2023xautoml,park2020hypertendril}.

\subsubsection{Intermediate step visualization}

We propose this unforeseen functionality as part of our taxonomy, inspired by the pipeline visualizations of \citep{DataRobot}. This functionality, Intermediate Step Visualization, aims to provide an interactive and visual representation of the outputs at each stage of the machine learning pipeline. Specifically, it includes visualizations of preprocessing, feature selection, and feature engineering steps, those that transform the data before model training.

These visualizations primarily take the form of tables, but can also include distribution plots and summary statistics. Users can explore these by interacting with the graphical representation of the trained pipeline. Clicking on each step reveals a preview of the data transformation, along with a comparison to the original, unprocessed data.

Intermediate step visualization is helpful to all actors, as it provides transparency to the internal processes of the system. Data Analysts and developers, in particular, can leverage this capability to detect issues and debug the pipeline.

\begin{table}[htb]
\small
\centering
\renewcommand{\arraystretch}{3} % Adjust row spacing
\begin{tabular}{llll} % Four columns, with wrapping for 2nd, 3rd, 4th
\toprule
\textbf{Functionality} & \textbf{Description} & \textbf{Examples} & \textbf{User} \\
\midrule
\makecell[l]{Optimization \\ Visualization} & \makecell[l]{Graphical representation \\ of hyperparameter effects \\ and optimization progress} & \makecell[l]{PDP plots \\ Learning curves \\ Pareto-frontier plots} & \makecell[l]{Data Scientists} \\

\makecell[l]{Pipeline \\ Visualization} & \makecell[l]{Visual overview of the \\ machine learning pipeline \\ structure and dependencies} & \makecell[l]{CPCP plots \\ Structure search graphs \\ Dashboards} & \makecell[l]{Data Analysts \\ Data Scientists} \\

\makecell[l]{Optimization \\ Summary} & \makecell[l]{Condensed overview of \\ optimization results and \\ key metrics} & \makecell[l]{Best model stats \\ Objective metrics \\ Failed run reports} & \makecell[l]{Data Analysts \\ Data Scientists} \\

\makecell[l]{Intermediate \\ Visualization} & \makecell[l]{Interactive display of \\ pipeline step results and \\ transformations} & \makecell[l]{Table visualizations \\ Distribution plots} & \makecell[l]{Data Scientists} \\
\bottomrule
\end{tabular}
\caption{This table presents the functionalities of Learning Process Explainability Taxonomy.}
\label{tab:learning_process_taxonomy}
\end{table}

\section{Open Challenges}

In this section, we describe related tools, identify their gaps by consolidating the proposed Question Bank, propose potential solutions that bridge the current gaps, and suggest open directions and challenges for future research on HXAI.

\subsection{Available Software}

\begin{figure}[ht!]
    \centering
    \includegraphics[width=0.6\textwidth]{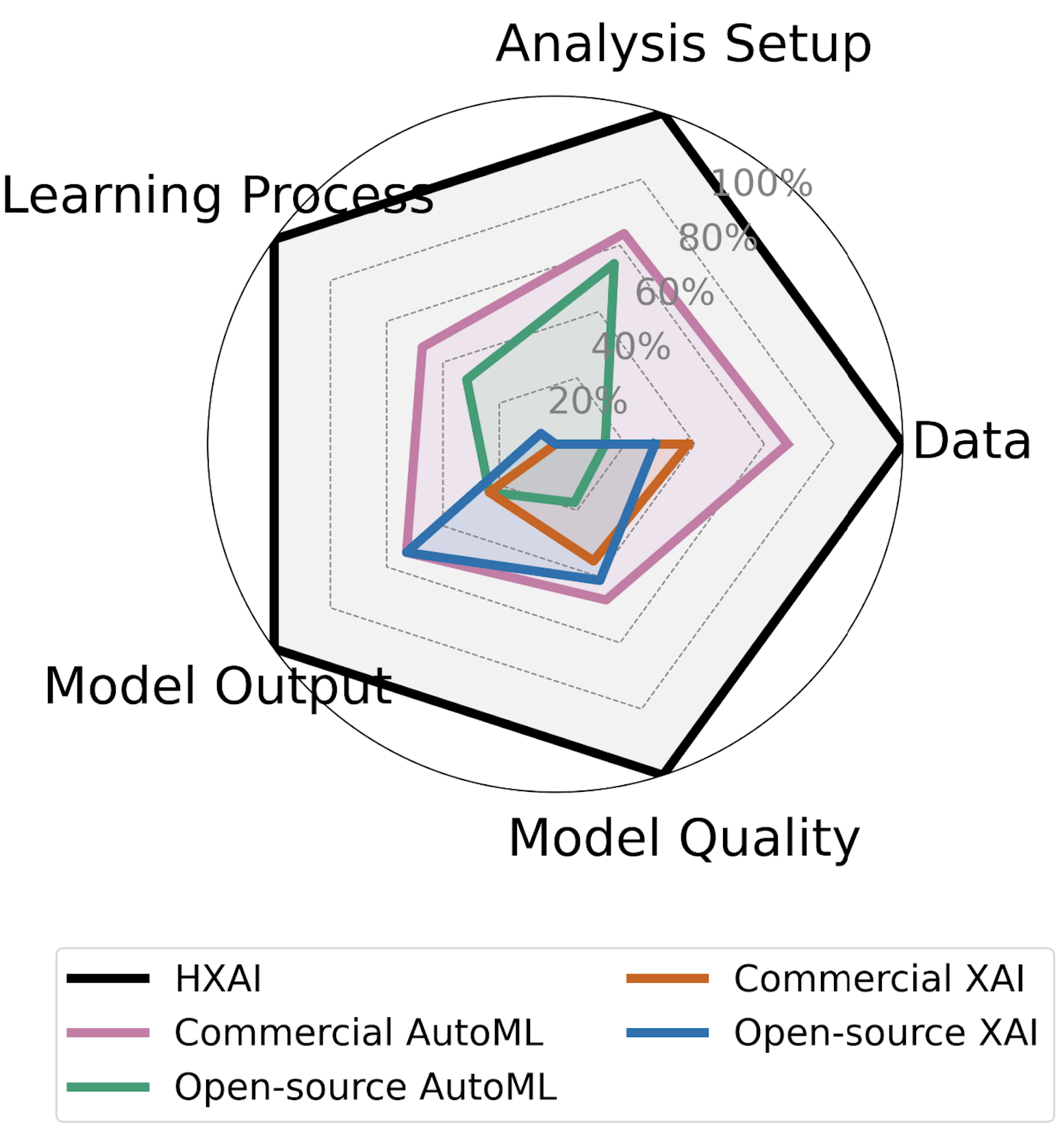} 
    \caption{This figure summarizes the performance of the four categories of AI software: open-source and commercial tools for both AutoML and XAI. We compare against the proposed HXAI solution. Explainability performance is measured as the percentage of HXAI questions answered within each HXAI component. To ensure a fair comparison, the highest score per functionality and HXAI component is used. Even under these optimistic conditions, the figure reveals that coverage across the HXAI components remains minimal across all software types.}
    \label{fig:Software-Spider} % Reference label
\end{figure}

Numerous software solutions, both open-source and commercial, provide methods for XAI. However, none of them fully addresses the proposed, broader, scope of HXAI. Figure \ref{fig:Software-Spider} depicts how existing AI software perform across each HXAI explainability component, based on the percentage of questions they are capable of answering from the proposed question bank. This percentage reflects the readiness of each tool, calculated by the proportion of component-specific questions it can address. Tables \ref{data-tbl}, \ref{analysis-setup-tbl}, \ref{model-predictions-tbl}, \ref{model-performance-tbl}, \ref{learning-process-tbl}, \ref{unified-expl-tbl} in the Appendix provide a more detailed breakdown of this readiness across the HXAI components.

Given that no existing tool performs well across all HXAI components, Figure \ref{fig:Software-Spider} illustrates an optimistic scenario representing the current state-of-the-art. For each component, the best-performing tool within a given software category is selected. Even under these favorable assumptions, the portion of HXAI questions answered is low demonstrating that existing tools fail to provide an adequate coverage of the HXAI framework. We should note here that, however, that these tools are dedicated to specific components, and are expected to under-perform in others. For example, we don't expect XAI tools to have good performance in learning process or analysis setup, as this is not their specialization.

Specifically, Open-source XAI tools predominantly focus on model output explanations, but they often fall short in areas such as data, learning process, and Quality Explainability \citep{intelXAI, wenzhuo2022omnixai, SpinnerEtAl2020, dalex, captum, Alibi, aix360, pyreal, ResponsibleAIToolbox}. On the other hand, commercial XAI tools tend to offer better support for data explainability \citep{ArizeAI, CensiusAI, WhyLabs, FiddlerAI}.

In the context of AutoML, open-source tools primarily emphasize result explainability \citep{autosklearn, H2OAutoML20, mljar, faml, autogluon}, whereas both open-source and commercial AutoML tools generally lack transparency in their learning processes (see Table \ref{learning-process-tbl}). Only commercial AutoML tools provide support for data explainability \citep{JADBio, H2ODriverlessAI, DataRobot, SageMakerAutopilot}.

There are some notable exceptions to these trends. For example, OmniXAI, a leading open-source XAI tool, offers limited capabilities for data and Quality Explainability. However, it incorporates LLM-based explanations for predictions \citep{wenzhuo2022omnixai}. Microsoft's Responsible AI toolbox offers error analysis, fairness methods and data explainability \citep{ResponsibleAIToolbox}.

A significant limitation of the aforementioned tools is their inability to provide actionable suggestions for improving data quality, model performance, or the learning process as seen in Table \ref{unified-expl-tbl} and Figure \ref{fig:Software-Spider}. This shortcoming reduces their chances for adoption by data scientists. Additionally, while the explanation methods employed by these tools are somewhat interpretable to experts, they still remain inaccessible to users without a background in ML. To the best of our knowledge, only a few of these tools have integrated LLMs. However, none of the tools provide automated guidance to help users diagnose potential issues in the data, model, or learning process. Instead, they rely on manual inspection of results, often presented in complex and unintuitive user interfaces.

\subsection{Future Directions to advance HXAI}

The field of HXAI faces several open challenges and promising future directions, all striving towards the ultimate goal of trustworthy AI. These challenges span from fundamental model design to user-centric evaluation and real-world implementation. Pursuing these future directions will be critical for realizing the full potential of HXAI and building truly trustworthy AI systems.

\noindent \textbf{Addressing the Performance-Explainability Trade-off}: Current research often focuses on improving the performance of inherently interpretable models or developing methods to explain black-box models. However, the trade-off between performance and explainability persists. A crucial future direction lies in the creation of self-explaining model architectures, eliminating the need for post-hoc explanations.

\noindent \textbf{Standardizing Explainability Evaluation}: While research exists on user studies and the desired properties of explanations, such as the Effective Explanation Qualities described in 
\ref{tab:HXAI-Effective-AIGents}, a standardized approach for conducting user studies and subjectively measuring explainability is still lacking. This leads to conflicting results across studies, hindering progress. Future research needs to develop standard methodologies for evaluating explainability, considering the diverse needs and goals of different user profiles. For instance, explanations suitable for machine learning experts may be ineffective for business executives, necessitating tailored evaluation approaches. To this direction, a widely accepted question bank, such as the one proposed here, could serve as the basis for a quantitative evaluation of explainability from any user perspective.

\noindent  \textbf{Measuring Trustworthiness Holistically}: Trustworthiness includes a combination of factors, such as stability, fairness, and explainability. Selecting the appropriate metrics for each factor and developing a comprehensive method for measuring overall system trustworthiness remains an open problem. Research is needed to define and quantify these dimensions of trust and to create standardized frameworks for assessing trustworthiness.

\noindent  \textbf{Achieving Real-Time Explainability}: Many existing XAI methodologies are computationally expensive, posing a significant challenge for real-time applications. Implementing real-time explanation systems requires the development of computationally efficient explanation methods that can enhance inference-side decision-making without introducing major delays. Research into optimized explainability algorithms and systems is necessary.

\noindent  \textbf{Advancing Interactive and Explainable Agents}: The development of interactive explanations and explainable AI agents is a new but promising area. To facilitate the widespread adoption of AI agents in real-world scenarios, there is a need for the (a) creation of datasets and tool-sets specifically designed for training and evaluating explainable agents, (b) comprehensive user studies to assess the robustness of AI agents against hallucination and their overall effectiveness in providing clear and informative explanations.

\noindent \textbf{Modular Multi-Agentic Systems} The current AI agent system functions as a single monolithic agent, acting as the intermediary between the AI system and the end-user. This approach can be enhanced by adopting a modular, multi-agentic architecture, where multiple specialized agents are each responsible for explaining a specific HXAI component. These agents would communicate with a central HXAI agent, which focuses exclusively on managing the interaction with the end-user.

\subsection{Limitations}

This work has several limitations. First, it does not present an in-depth, systematic review of the XAI literature as seen in other reviews \citep{almeida2025systematic}. Instead, we highlight key contributions identified through prior review papers and searches across major research repositories. Second, our analysis primarily focuses on two data modalities, tabular and image data. Although many of the methods discussed could be extended to other modalities (such as text), an extended approach of all possible data types lies beyond the scope of this paper. Third, the proposed framework is centered on predictive modeling within the supervised learning setting. While our modular, plug-and-play HXAI proposal may be adaptable to other learning paradigms, such as unsupervised learning, this would require further improvements and additions tailored to those tasks. We do not provide a practical, code-based implementation of the HXAI framework. Instead, our aim is to introduce a novel perspective that addresses foundational challenges in the current XAI field. Finally, the proposed approach has not yet been validated through user studies or surveys. Rather, it builds upon insights from prior literature to propose a novel, but yet untested, direction for XAI research.

\section{Conclusion}
In this paper, we introduced Holistic Explainable AI (HXAI), a novel approach that broadens the scope of explainability in AI systems. Unlike traditional XAI methods that primarily focus on model outputs, HXAI extends explainability across six additional components: data, learning process, analysis setup, model quality, model output, and explainability agent. 

We proposed a new taxonomy, defining the key components of HXAI and illustrating how it seamlessly integrates into the machine learning workflow. To support practical implementation, we introduced a structured question bank designed to guide users in applying HXAI principles, while also identifying critical gaps in existing software tools. Our contributions include the characterization of what constitutes effective explanations and a definition of the key parts that build trust in AI, seen as the main goal of HXAI. 

Furthermore, we proposed the concept of an HXAI agent capable of automating and tailoring explanations for diverse user needs, paving the way for more accessible and adaptive explainability. Finally, we outlined future research directions to further develop the field. By moving beyond traditional post-hoc methods, HXAI enables more transparent, interpretable, and user-centric AI systems, better addressing the diverse needs of stakeholders across domains.

%\bmhead{Acknowledgments}

\section*{Declarations}

% https://www.springer.com/gp/editorial-policies/competing-interests
% Some journals require declarations to be submitted in a standardised format. Please check the Instructions for Authors of the journal to which you are submitting to see if you need to complete this section. If yes, your manuscript must contain the following sections under the heading `Declarations':

\begin{itemize}
\item Funding: The project was partly founded by Honda Research Institute Europe GmbH.
\item Competing interests: The authors declare that they have no known competing financial interests or personal
relationships that could have appeared to influence the work reported in this paper.
\item Data availability: No datasets were analysed or created during the study.
\item Code availability: No code was written during the study.
\item Author contribution: George Paterakis structured and wrote the manuscript. Andrea Castellani helped in the literature review as well as the overall writing and structuring of the manuscript. George Papoutsoglou offered guidance in structuring and writing the manuscript. 
Tobias Rodemann and Ioannis Tsamardinos supervised the project, ensuring the quality and accuracy of the final manuscript.
All authors reviewed the manuscript.
%\item Ethics approval and consent to participate
%\item Consent for publication
%\item Data availability 
\end{itemize}

\clearpage
\begin{appendices}

\section{Definitions}
\label{app:def}

In this section, we present an extra set of definitions for HXAI for completeness.

\noindent \textbf{Definition 1} \textit{(Understandability)}. A property of a model that allows humans to understand how it functions without knowing its internal functionality \citep{MONTAVON}.

\noindent \textbf{Definition 2} \textit{(Trust)}. The extent to which users are willing to rely on an AI system's outputs. \citep{MERSHA2024128111}.

\noindent \textbf{Definition 3} \textit{(Explanation)}. The answer to a "why" or "why-should" question. A good explanation requires both interpretability and completeness \citep{gilpin}.

\noindent \textbf{Definition 4} \textit{(Explicability)}. A property stating that something potentially can be explained \citep{10.3389/frai.2020.507973}.

\noindent \textbf{Definition 5} \textit{(Transparency)}. The ability to understand the mechanism of how the model works \citep{Mythos-Lipton}. Transparent models can be categorized into three types according to \citep{Mythos-Lipton}.

\noindent \textbf{Definition 6} \textit{(Confidence)}. Denotes the probability of an event happening and measures the trust in a model's decision \citep{rojat2021explainable,kailkhura2019reliable}.

\noindent \textbf{Definition 7} \textit{(Explanandum)}. The model that is being explained, usually an AI model or an ML pipeline \citep{miller2019explanation}.

\noindent \textbf{Definition 8} \textit{(Explanator)}. The system that generates or provides the explanations \citep{miller2019explanation}.

\noindent \textbf{Definition 9} \textit{(Explainee)}. The recipient of the explanations provided by the explanation system \citep{miller2019explanation}.

\noindent \textbf{Definition 10} \textit{(Interpretable Models)}. Machine learning methods that do not require an explanator to be understood \citep{Schwalbe2024}. Also referred to as transparent models \citep{Mythos-Lipton}.

\noindent \textbf{Definition 11} \textit{(Black-box Models)}. Machine learning models that are not interpretable by humans \citep{GUIDOTTI}, as their internals are either unknown or known but not interpretable.

\noindent \textbf{Definition 12} \textit{(Interpretable Machine Learning (iML))}. The research area focused on developing interpretable models \citep{Schwalbe2024}.

\section{HXAI Question Bank}
\label{app:qb}

\subsection{Data Explainability}

Data Explainability focuses on making data understandable to users. Domain experts often pose broad questions such as, "What does my data contain?" or "Is my data reliable?" However, current Data Explainability methods struggle to address such vague inquiries. Instead, we provide techniques for answering more specific questions that Data Analysts and Data Scientists commonly ask. In the future, a Large Language Model could leverage these specific questions and their corresponding answers to generate clear and comprehensive explanations for domain experts. The following questions are inspired by the field of Exploratory Data Analysis \citep{tukey77}. Table \ref{tbl:data-questions} presents the questions, their functionality in the HXAI taxonomy, the user that will benefit from the answer, and the methods that can answer the question.

\begin{small}
\renewcommand{\arraystretch}{3}
\begin{longtable}{llll}
    \caption{Data Questions \label{tbl:data-questions}}\\
    \toprule
    \textbf{Questions} & \textbf{Functionality} & \textbf{User} & \textbf{Methods} \\

        \makecell[l]{How are the \\ features and target \\ distributed? } & \makecell[l]{Data \\ Visualization} & \makecell[l]{DA, DS} & \makecell[l]{Histogram \\ KDE Plot \\ Bar Plot} \\ 
    
        %\makecell[l]{For categorical features,\\  how are the  \\ categories distributed? } & \makecell[l]{Data \\ Visualization} & \makecell[l]{DA, DS} & \makecell[l]{Histogram \\ KDE Plot} \\ 
        
        %\makecell[l]{For numerical features,\\ how are continuous \\ values distributed? } & \makecell[l]{Data \\ Visualization} & \makecell[l]{DA, DS} & \makecell[l]{Histogram \\ Bar Plot} \\ 
        
        \makecell[l]{Are outliers visualized?} & \makecell[l]{Data \\ Visualization} & \makecell[l]{DA, DS} & \makecell[l]{Histogram \\ Bar Plot} \\ 
        
        %\makecell[l]{How is the \\ target variable \\ distributed? } & \makecell[l]{Data \\ Visualization} & \makecell[l]{DA, DS} & \makecell[l]{Histogram \\ Bar Plot} \\ 
        
        \makecell[l]{How do pairs \\ of features/target \\ relate visually?} & \makecell[l]{Data \\ Visualization} & \makecell[l]{DA, DS} & \makecell[l]{Scatter \\ Heatmap \\ Bar plots} \\ 
        
        %\makecell[l]{Are there any \\ visual patterns \\ between the target \\ and key features? } & \makecell[l]{Data \\ Visualization} & \makecell[l]{All} & \makecell[l]{Scatter \\ Heatmap \\ Bar plots} \\
        
        \makecell[l]{How features \\ are interacting? } & \makecell[l]{Data \\ Visualization} & \makecell[l]{DA, DS} & \makecell[l]{Heatmap \\ PCP \\ \citep{PCPOrder}} \\ 
        
        \makecell[l]{Is the data \\ separable on \\ a latent space?} & \makecell[l]{Data \\ Visualization} & \makecell[l]{DS} & \makecell[l]{PCA. \\ \citep {PCA} \\ UMAP \\ \citep{mcinnes2018umap}} \\ 
        
        \makecell[l]{What are the \\ dimensions of \\ the dataset? } & \makecell[l]{Data \\ Summary} & \makecell[l]{DA, DS} & \makecell[l]{Text} \\ 
        
        \makecell[l]{What are the \\ data types of \\ each feature? } & \makecell[l]{Data \\ Summary} & \makecell[l]{DA, DS} & \makecell[l]{Text} \\ 
        
        \makecell[l]{Is the data \\ containing sensitive \\ information?} & \makecell[l]{Data \\ Summary} & \makecell[l]{All} & \makecell[l]{Text} \\ 
        
        \makecell[l]{What are the \\ distributions of \\ each feature/target? } & \makecell[l]{Data \\ Summary} & \makecell[l]{DA, DS} & \makecell[l]{Statistical \\ Meta-Features \\ \citep{RIVOLLI2022108101}} \\ 
        
        \makecell[l]{Which features \\ have missing values, \\ and what percentage is \\ missing for each?} & \makecell[l]{Data \\ Summary} & \makecell[l]{DA, DS} & \makecell[l]{Simple Meta-Features \\ \citep{RIVOLLI2022108101}} \\ 
        
        %\makecell[l]{What are general \\ statistics of numerical \\ features by \\ categorical groups?} & \makecell[l]{Data \\ Summary} & \makecell[l]{DA, DS} & \makecell[l]{Text} \\ 
        
        % \makecell[l]{How do certain \\ features vary across \\ different values in \\ the target variable? } & \makecell[l]{Data \\ Summary} & \makecell[l]{DA, DS} & \makecell[l]{Text} \\ 
        
        %\makecell[l]{What is the  \\ distribution of the \\ target variable? } & \makecell[l]{Data \\ Summary} & \makecell[l]{DA, DS} & \makecell[l]{Simple \\ Meta-Features \\ \citep{RIVOLLI2022108101}} \\ 
        
        \makecell[l]{How are missing \\ values distributed? } & \makecell[l]{Data \\ Quality} & \makecell[l]{DA, DS} & \makecell[l]{Simple \\ Meta-Features \\ \citep{RIVOLLI2022108101} }\\ 
        
        \makecell[l]{Are there any \\ patterns in the \\ missing data? } & \makecell[l]{Data \\ Quality} & \makecell[l]{DA, DS} & \makecell[l]{Text \\ Little's Test \\ \citep{Little01121988} } \\ 
        
        \makecell[l]{Is the target \\ variable imbalanced? } & \makecell[l]{Data \\ Quality} & \makecell[l]{DA, DS} & \makecell[l]{Meta-Features} \\ 
        
        \makecell[l]{Are there \\ under-represented populations? } & \makecell[l]{Data \\ Quality} & \makecell[l]{All} & \makecell[l]{Text} \\ 
        
        \makecell[l]{Are there duplicate \\ or highly correlated \\ samples/ features? } & \makecell[l]{Data \\ Quality} & \makecell[l]{DA, DS} & \makecell[l]{Data \\ Deduplication \\ \citep{wang2023sudowoodo} \\ \citep{jin2021deep}} \\ 
        
        %\makecell[l]{Are there highly correlated \\ samples or features? } & \makecell[l]{Data \\ Quality} & \makecell[l]{DA, DS} & \makecell[l]{Data \\ Deduplication} \\ 
        
        \makecell[l]{Are there any \\ noticeable outliers?} & \makecell[l]{Data \\ Quality} & \makecell[l]{DA, DS} & \makecell[l]{Isolation Forest \\ \citep{liu2008isolation} \\ VAE \\ \citep{pmlr-v108-eduardo20a}} \\ 
        
        \makecell[l]{Are features \\ linearly or non-linearly \\ correlated?} & \makecell[l]{Data \\ Relationships} & \makecell[l]{DA, DS} & \makecell[l]{Pearson  \\ \\ Spearman \\ } \\ 
        
        \makecell[l]{Are there meaningful \\ transformations that could make \\ relationships more effective \\ or interpretable? } & \makecell[l]{Data \\ Relationships} & \makecell[l]{DA, DS} & \makecell[l]{Log \\ Polynomial \\  Ratios \\ Differences } \\ 
        
        \makecell[l]{Do any observed correlations \\ suggest potential causation?} & \makecell[l]{Data \\ Relationships} & \makecell[l]{DS} & \makecell[l]{Causal Discovery \\ \citep{tsamardinos2006max} \\ \citep{biza2024automatedcausaldiscoverycase}} \\ 
        
        %\makecell[l]{Are there derived features \\ that capture relationships \\ more effectively?} & \makecell[l]{Data \\ Relationships} & \makecell[l]{DS} & \makecell[l]{Ratios \\ Differences} \\ 
        
        \makecell[l]{Can the data be \\ split into clusters? } & \makecell[l]{Data \\ Relationships} & \makecell[l]{DA, DS} & \makecell[l]{K-means \\ \citep{kmeans} \\ DBScan \\ \citep{ester1996density} } \\ 
        
        \makecell[l]{Are there any \\ consistent patterns within \\ specific populations?} & \makecell[l]{Data \\ Relationships} & \makecell[l]{All} & \makecell[l]{Clustering and \\ Feature Importance  \\ \citep{fisher2019all}} \\ 
    \midrule
\end{longtable}
\end{small}

\subsection{Analysis Setup Explainability}

Analysis Setup Explainability can help explain the setup choices to the user. This component is novel, and methodology is mostly based on domain knowledge. In case the ML process is done through an AutoML tool, meta-learning and rule-based setup choices are used. Table \ref{tbl:analysis-setup-questions} presents the questions corresponding to this HXAI component.

\begin{table}[htb!]
    
    \caption{Analysis Setup Questions \label{tbl:analysis-setup-questions}}
    \small
    \renewcommand{\arraystretch}{3} % Adjust row spacing
    
    \begin{tabular}{llll}
         \toprule
        \textbf{Questions} & \textbf{Functionality} & \textbf{User} & \textbf{Methods} \\ 
        \midrule
        
        \makecell[l]{What is \\ the type of \\ the problem?} 
        & \makecell[l]{Problem \\ Formulation}
        & \makecell[l] {DE}
        & \makecell[l]{Text} \\ 
        
        \makecell[l]{What output \\ can I expect \\ from the model?} 
        & \makecell[l]{Problem \\ Formulation}
        & \makecell[l] {DE} 
        & \makecell[l]{Text} \\ 
        
        \makecell[l]{Which/Why did  this \\ validation protocol \\ get selected?} 
        & \makecell[l]{ Analysis \\ Optimization }
        & \makecell[l]{DA \\ DS} 
        & \makecell[l]{Rules-Based \\ Meta-Learning \\ \citep{autosklearn}} \\ 
        
        \makecell[l]{How does the \\ validation protocol \\ works visually?} 
        & \makecell[l]{ Analysis \\ Optimization }
        & \makecell[l] {DE} 
        & \makecell[l]{Visual animation} \\ 
        
        \makecell[l]{What is the metric \\ to be optimized \\ for this problem?} 
        & \makecell[l]{ Analysis \\ Optimization }
        & \makecell[l] {All}
        & \makecell[l]{ Text } \\ 
        
        \makecell[l]{Why is this metric \\ suitable for this problem \\ and what it \\ focuses on?} 
        & \makecell[l]{ Analysis \\ Optimization }
        & \makecell[l] {DE}  
        & \makecell[l]{AI Explanation} \\ 
        
        \makecell[l]{Is the metric \\ aligned with \\ business goals?} 
        & \makecell[l]{ Analysis \\ Optimization }
        & \makecell[l] {All} 
        & \makecell[l]{User input \\ AI Explanation} \\ 
        
        \makecell[l]{What is the \\ success criteria \\ for the task?} 
        & \makecell[l]{ Analysis \\ Optimization }
        & \makecell[l] {DE} 
        & \makecell[l]{Trivial Model Score} \\ 
        
        \makecell[l]{Which steps \\ will be included  \\ in the pipeline \\ and why?} 
        & \makecell[l]{ Analysis \\ Optimization }
        & \makecell[l]{DA \\ DS}  
        & \makecell[l]{SSG \citep{zoller2023xautoml} \\ AI Explanation} \\ 
        
        \makecell[l]{Which algorithms \\ will be included \\ in each step \\ and why?} 
        & \makecell[l]{ Analysis \\ Optimization } 
        & \makecell[l]{DS}  
        & \makecell[l]{ MetaLearning 
                        \\ \citep{borboudakis2023meta} 
                        \\  \citep{brazdil2003ranking} 
                        \\ + AI Explanation } \\ 
        
        \makecell[l]{Which hyper-parameters \\ will be tuned for each \\ algorithm and why?} 
        & \makecell[l]{ Analysis \\ Optimization }  
        & \makecell[l]{DS} 
        & \makecell[l]{MetaLearning 
                        \\ \citep{tunability}
                        \\ \citep{van2018hyperparameter} 
                        \\ AI Explanation} \\ 
        \bottomrule
    \end{tabular}
\end{table}

\subsection{Results Explainability}

Results explainability is concerned with explaining the results of the analysis to the user. It can be further split into two categories, (a) "Output" that explains the models predictions/output and (b) "Quality" that explains the model's performance.

\subsubsection{Output Explainability}

The following set of questions is primarily derived from established XAI literature. Example-based methods, which provide explanations through specific instances from the dataset, and cohort-based methods, which offer insights based on groups of similar instances, have been less prominent in the literature. Table \ref{tbl:model-output-questions} presents the questions, functionality, users, and methods in this component.

\begin{small}[htb!]
\renewcommand{\arraystretch}{3}
\begin{longtable}{llll}
    \caption{Model Output Questions \label{tbl:model-output-questions}}\\
    \toprule
        \textbf{Questions} & \textbf{Functionality} & \textbf{User} & \textbf{Methods} \\ 
        \midrule
        \makecell[l]{Which training instances \\ most influenced  \\  a specific prediction?} & \makecell[l]{Post-Hoc \\ Example} & \makecell[l]{DS} & \makecell[l]{Influence \\  \citep{koh2017understanding} \\ \citep{krishnan2017palm}  } \\ 
        
        \makecell[l]{Were influencing instances  \\  similar or dissimilar  \\  to the test instance? } & \makecell[l]{Post-Hoc \\ Example} & \makecell[l]{DS} & \makecell[l]{Influence \\  \citep{koh2017understanding} \\ \citep{krishnan2017palm} } \\ 
        
        \makecell[l]{How does the  model  \\ perform on “prototypes”  \\ or  “criticisms”?} & \makecell[l]{Post-Hoc \\ Example} & \makecell[l]{DS} & \makecell[l]{MMD-Critic\\ \citep{kim2016examples} \\ Protodash\\ \citep{gurumoorthy2019efficient}} \\ 
        
        %\makecell[l]{How does the model \\ output  vary on    \\  “criticisms” that  \\ represent  edge cases? } & \makecell[l]{Post-Hoc \\ Example} & \makecell[l]{DS} & \makecell[l]{MMD-Critic\\ \citep{kim2016examples}} \\ 
        
        %\makecell[l]{How does the  \\  model  output  \\  vary across  \\  similar instances?} & \makecell[l]{Post-Hoc \\ Example} & \makecell[l]{DE \\ DA} & \makecell[l]{KNN similarity} \\ 
        
        \makecell[l]{What small, \\  plausible changes   to an \\ instance would   lead to  \\ a different prediction?} & \makecell[l]{Post-Hoc \\ Example} & \makecell[l]{DE \\ DA} & \makecell[l]{Counterfactuals \\ \citep{wachter2017counterfactual} \\ \citep{mothilal2020explaining} } \\ 
        
        \makecell[l]{Is the model  \\ very sensitive to   \\ input changes?} & \makecell[l]{Post-Hoc \\ Example-Based} & \makecell[l]{DS} & \makecell[l]{Adversarial Attacks \\  \citep{szegedy2013intriguing}} \\ 
        
        \makecell[l]{What changes  \\ can I make without   \\ changing my prediction?} & \makecell[l]{Post-Hoc \\ Local} & \makecell[l]{DE \\ DA} & \makecell[l]{Anchors \\ \citep{ribeiro2018anchors}} \\ 
        
        \makecell[l]{For a given prediction  \\ which features influenced   \\ the outcome   \\ and how much?} & \makecell[l]{Post-Hoc \\ Local} & \makecell[l] {DE} & \makecell[l]{LIME\\ \citep{LIME} \\ SHAP\\ \citep{scott2017unified} \\ GRAD-CAM \\\citep{jiang2021layercam} \\ RISE \\ \citep{petsiuk2018rise}} \\ 
        
        \makecell[l]{How does the  \\ local feature importance   \\ compare with the  \\ global feature importance?} & \makecell[l]{Post-Hoc \\ Local} & \makecell[l]{DA \\ DS} & \makecell[l]{Dashboard} \\ 
        
        \makecell[l]{How does a  \\ prediction change when  \\ we change the values   \\of each feature?} & \makecell[l]{Post-Hoc \\ Local} & \makecell[l]{DE \\ DA} & \makecell[l]{ICE \\ \citep{ICE}} \\ 
        
        %\makecell[l]{Which pixels or regions \\ contributed most to  \\ a specific prediction?} & \makecell[l]{Post-Hoc \\ Local} & \makecell[l]{DE \\ DA} & \makecell[l]{GRAD-CAM \\\citep{jiang2021layercam} \\ RISE \\ \citep{petsiuk2018rise} \\ Expected Gradients \\ \citep{erion2021improving}} \\ 
        
        \makecell[l]{How does feature  \\ importance change across  \\ different subgroups?} & \makecell[l]{Post-Hoc \\ Global} & \makecell[l] {All} & \makecell[l]{Dashboard} \\ 
        
        \makecell[l]{Do feature interactions\\ within a cohort significantly\\ impact predictions?} & 
        \makecell[l]{Post-Hoc \\ Global} & \makecell[l] {All} & \makecell[l]{Feature Interaction \\ \citep{friedman2008predictive}} \\ 
        
        \makecell[l]{Are certain features \\ more important for \\ predicting specific classes \\ than others?} & \makecell[l]{Post-Hoc \\ Global} & \makecell[l] {All} & \makecell[l]{Feature Importance \\  \citep{fisher2019all} \\ TCAV \citep{kim2018interpretability} \\ ACE \citep{ghorbani2019towards} \\ SpRay \\ \citep{Lapuschkin2019}\\ GAM\\ \citep{ibrahim2019global}} \\ 
        
        %\makecell[l]{Are there visual attributes \\ that affect predictions \\ differently across cohorts?} & \makecell[l]{Post-Hoc \\ Global} & \makecell[l] {All} & \makecell[l]{TCAV \citep{kim2018interpretability} \\ ACE \citep{ghorbani2019towards}} \\ 
        
        %\makecell[l]{Does the model focus \\ on different parts \\ of the image for \\  different cohorts?} & \makecell[l]{Post-Hoc \\ Global} & \makecell[l] {All} & \makecell[l]{SpRay \\ \citep{Lapuschkin2019}\\ GAM\\ \citep{ibrahim2019global}} \\ 
        
        \makecell[l]{Which features have \\ the most influence on \\ the model’s overall \\ predictions?} & \makecell[l]{Post-Hoc \\ Global} & \makecell[l] {All} & \makecell[l]{Feature Importance\\  \citep{fisher2019all} \\ SHAP\\ \citep{lundberg2020local} \\ GAM \citep{ibrahim2019global} } \\ 
        
        \makecell[l]{Are there specific \\features interactions \\ that significantly \\ impact predictions?} & \makecell[l]{Post-Hoc \\ Global} & \makecell[l] {All} & \makecell[l]{Feature Interaction\\ \citep{friedman2008predictive}} \\
        
        \makecell[l]{How does the average \\  prediction of a model \\ change when the values \\ of each feature change?} & \makecell[l]{Post-Hoc \\ Global} & \makecell[l] {All} & \makecell[l]{PDP\\ \citep{friedman2001greedy} \\ ALE\\ \citep{apley2020visualizing}} \\

        \makecell[l]{How consistent \\ are the important features \\ across different models? } & \makecell[l]{Post-Hoc \\ Global} & \makecell[l]{DA \\ DS} & \makecell[l]{Dashboard} \\ 
        
        \makecell[l]{What is a simple \\ visualization of my model?} & \makecell[l]{Post-Hoc \\ Global } & \makecell[l] {DE} & \makecell[l]{Surrogate models \\ \citep{bastani2017interpreting} \\ \citep{lakkaraju2017interpretable}} \\
        
        %\makecell[l]{In which part \\ of the images does \\ the model focus to \\ make overall predictions? } & \makecell[l]{Post-Hoc \\ Global} & \makecell[l] {All} & \makecell[l]{GAM \citep{ibrahim2019global}} \\ 

    \bottomrule
\end{longtable}
\end{small}

\subsubsection{Quality Explainability}

Quality Explainability is concerned with explaining the performance of the model to the user. The questions presented below are inspired by the commercial AutoMLs' performance reports as well as tools on fairness and error analysis such as Microsoft's Responsible AI toolbox \citep{DataRobot,H2ODriverlessAI,ResponsibleAIToolbox}. Table \ref{tbl:model-quality-questions} presents the questions, functionality, users, and methods in this component.

\begin{small}
\renewcommand{\arraystretch}{3}
\begin{longtable}{llll}
    \caption{Model Quality Questions \label{tbl:model-quality-questions}}\\
    \toprule
        \textbf{Questions} & \textbf{Functionality} & \textbf{User} & \textbf{Methods} \\ 
        \makecell[l]{How does the\\ performance look like?} & \makecell[l]{Performance \\ Visualization} & \makecell[l]{DA \\ DS} & \makecell[l]{ROC \\ PR Curve} \\ 
        
        \makecell[l]{What does the\\ calibration curve\\ look like?} & \makecell[l]{Performance \\ Visualization} & \makecell[l]{DS} & \makecell[l]{Calibration curve \\ Visualization} \\ 
        
        \makecell[l]{In which metrics\\ does the model\\ underperform?} & \makecell[l]{Performance \\ Visualization} & \makecell[l]{DS} & \makecell[l]{Text} \\ 
        
        \makecell[l]{What are the most \\ common types of errors \\ and which classes are \\ mostly affected?} & \makecell[l]{Error \\ Analysis} & \makecell[l]{DA \\ DS} & \makecell[l]{Confusion Matrix} \\ 
        
        \makecell[l]{For each error type,\\ can specific patterns \\ in the misclassified samples \\ be identified?} & \makecell[l]{Error \\ Analysis} & \makecell[l]{DS} & \makecell[l]{Feature Importance  \\ \citep{fisher2019all} \\ SHAP \\  \citep{lundberg2020local}} \\ 
        
        \makecell[l]{Are there certain ranges/values \\ of features \\ where the model \\consistently performs poorly?} & \makecell[l]{Error \\ Analysis} & \makecell[l]{DS} & \makecell[l]{Manual \\ \citep{chung2019automated} \\ DT \\ \citep{chen2004failure} \\ Data Slicing \\ \citep{cabrera2023zeno} } \\ 
        
        %\makecell[l]{Are there feature categories \\ that are disproportionately \\ associated with \\ higher error rates?} & \makecell[l]{Error \\ Analysis} & \makecell[l]{DS} & \makecell[l]{Manual \\ \citep{chung2019automated} \\ DT \\ \citep{chen2004failure} \\ Data Slicing \\ \citep{cabrera2023zeno} } \\ 
        
        \makecell[l]{Can errors be \\ clustered into groups \\ with similar characteristics?} & \makecell[l]{Error \\ Analysis} & \makecell[l]{DS} & \makecell[l]{Data Slicing \\ \citep {plumb2022towards} \\ \citep{eyuboglu2022domino} \\ \citep{d2022spotlight} } \\ 
        
        \makecell[l]{How does model \\ performance vary \\ across different groups? } & \makecell[l]{Fairness \\ Assessment} & \makecell[l] {All} & \makecell[l]{Dashboard} \\ 
        
        \makecell[l]{Is the model’s \\ probabilities calibrated \\ across different groups?} & 
        \makecell[l]{Fairness \\ Assessment} & \makecell[l]{DS} & \makecell[l]{Equal Calibration \\ \citep{doi:10.1089/big.2016.0047} } \\ 
        
        \makecell[l]{Are there specific \\ groups with consistently \\ lower performance?} & \makecell[l]{Fairness \\ Assessment} & \makecell[l] {All} & \makecell[l]{Statistical Tests \\ \citep{uddin2024novel}} \\ 
        
        \makecell[l]{How does the \\ model perform across \\ various fairness metrics?} & \makecell[l]{Fairness \\ Assessment} & \makecell[l]{DA \\ DS} & \makecell[l]{Fairness Metrics \\ \citep{aghaei2019learning} \\ \citep{NIPS2016_9d268236} \\ \citep{dwork2012fairness} \\ \citep{berk2017convex}} \\ 
        
        \makecell[l]{How does model \\ perform across metrics?} & \makecell[l]{Performance \\ Summary} & \makecell[l]{DA \\ DS} & \makecell[l]{Dashboard} \\ 

        \makecell[l]{How metric values  \\ vary across classes?} & \makecell[l]{Performance \\ Summary} & \makecell[l]{DS} & \makecell[l]{Dashboard} \\ 
        
        \makecell[l]{What is the best \\ threshold for each metric?} & \makecell[l]{Performance \\ Summary} & \makecell[l]{DS} & \makecell[l]{Threshold Optimization \\ \citep{koseoglu2024otlp} \\ \citep{zou2016finding} } \\ 
        
        \makecell[l]{Does the model \\ produce well-calibrated \\ probabilities?} & \makecell[l]{Performance \\ Summary} & \makecell[l]{DS} & \makecell[l]{Calibration Metrics\\  \citep{naeini2015obtaining} \\ \citep{kull2019beyond} \\ \citep{calibration2}} \\ 
        
        \makecell[l]{What are the \\confidence intervals \\ for each metric?} & \makecell[l]{Performance \\ Summary} & \makecell[l]{DA \\ DS} & \makecell[l]{Resampling \\ \citep{willmott1985statistics}  \\ Boostrapping \\ \citep{Tsamardinos2018}} \\ 
        
        \makecell[l]{How does the performance \\ of a (the best) model compare \\ against another model?} & \makecell[l]{Performance \\ Summary} & \makecell[l]{DS} & \makecell[l]{Statistical Tests\\  \citep{demvsar2006statistical} \\ \citep{6790639}} \\ 
\end{longtable}
\end{small}

\subsection{Learning Process Explainability}

The Learning Process Explainability is crucial for data scientists aiming to enhance their machine learning workflows. The questions in this section are inspired by previous works on the ixAutoML field and the related tools \citep{zoller2023xautoml,park2020hypertendril,humancenteredautoml,sass2022deepcave}. Table \ref{tbl:learning-process-questions} presents the questions, functionality, users, and methods in this component.

\begin{small}
\renewcommand{\arraystretch}{3}
\begin{longtable}{llll}
    \caption{Learning Process Questions \label{tbl:learning-process-questions}}\\
    \toprule
    \textbf{Questions} & \textbf{Functionality} & \textbf{User} & \textbf{Methods} \\
    \midrule
    
        \makecell[l]{How does the \\ ML system explore \\ the search space \\ over time?} & \makecell[l]{Optimization \\ Visualization} & \makecell[l]{DS} & \makecell[l]{Colored Scatter plots \\ \citep{zoller2023xautoml}, } \\ 
        
        \makecell[l]{Does the ML \\ process demonstrate  \\ a preference for \\specific algorithms? } & \makecell[l]{Optimization \\ Visualization} & \makecell[l]{DS} & \makecell[l]{Histogram \\ \citep{wang2019atmseer}} \\ 
        
        \makecell[l]{How does the \\ loss function change \\ over training epochs?} & \makecell[l]{Optimization \\ Visualization} & \makecell[l]{DS} & \makecell[l]{Learning Curve\\ \citep{park2020hypertendril}} \\ 
        
        %\makecell[l]{Do the training \\ and validation \\ learning curves \\ suggest difficulty \\in optimization? } & \makecell[l]{Optimization \\ Visualization} & \makecell[l]{DS} & \makecell[l]{Learning Curve\\ \citep{park2020hypertendril}} \\ 
        
        %\makecell[l]{Are there any \\ plateau periods or \\ sudden drops in loss?} & \makecell[l]{Optimization \\ Visualization} & \makecell[l]{DS} & \makecell[l]{Learning Curve\\ \citep{park2020hypertendril}} \\ 
        
        %\makecell[l]{How does the \\ cost evolve through \\ the optimization process? } & \makecell[l]{Optimization \\ Visualization} & \makecell[l]{DS} & \makecell[l]{Learning Curve\\ \citep{park2020hypertendril}} \\ 
        
        \makecell[l]{Which hyper-parameters \\ are the most important?} & \makecell[l]{Optimization \\ Visualization} & \makecell[l]{DS} & \makecell[l]{F-Anova \\ \citep{hutter2014efficient} \\ forward selection \\ \citep{forwardselection} \\ N-RReliefF \\ \citep{Sun2019HyperparameterIA} } \\ 
        
        \makecell[l]{Which algorithms \\ were trained \\ on each step? } & \makecell[l]{Optimization \\ Visualization} & \makecell[l]{DA \\ DS} & \makecell[l]{SSG \citep{zoller2023xautoml} } \\ 
        
        \makecell[l]{Which hyperparameters \\ were tuned, what is \\ their distribution \\ and ranges?} & \makecell[l]{Optimization \\ Visualization} & \makecell[l]{DS} & \makecell[l]{MDS \\ \citep{sass2022deepcave} \\ Scatter Plots\\ \citep{zoller2023xautoml}} \\ 
        
        \makecell[l]{How does one \\ hyper-parameter \\ affect the performance result?} & \makecell[l]{Optimization \\ Visualization} & \makecell[l]{DS} & \makecell[l]{PDP\\  \citep{moosbauer2021explaining} \\ ICE \\ \citep{zoller2023xautoml} } \\ 
        
        \makecell[l]{What model type \\, hyperparameters, and \\ feature transformations did \\ ML select as optimal?} & \makecell[l]{Optimization \\ Summary} & \makecell[l]{DA \\ DS} & \makecell[l]{Text} \\ 
        
        \makecell[l]{What were the final \\ performance metrics, \\ what is the training \\ prediction time, \\ the pipeline size?} & \makecell[l]{Optimization \\ Summary} & \makecell[l]{DA \\ DS} & \makecell[l]{Dashboard} \\ 
        
        \makecell[l]{How does the \\ best model’s summary \\ compare against others? } & \makecell[l]{Optimization \\ Summary} & \makecell[l]{DA \\ DS} & \makecell[l]{Dashboard} \\ 
        
        \makecell[l]{What was the \\ training objective?} & \makecell[l]{Optimization \\ Summary} & \makecell[l]{DA \\ DS} & \makecell[l]{Text} \\ 

        \makecell[l]{Were there any errors \\ during optimization? } & \makecell[l]{Optimization \\ Summary} & \makecell[l]{DS} & \makecell[l]{Status Heatmap\\ \citep{sass2022deepcave}} \\ 
        
        \makecell[l]{What is the \\ output of each \\ preprocessing method?} & \makecell[l]{Intermediate \\ Visualization} & \makecell[l]{DS} & \makecell[l]{Table  \\ Visualization} \\ 
        
        \makecell[l]{How does the \\ preprocessed distributions \\ compare against the \\ original data?} & \makecell[l]{Intermediate \\ Visualization} & \makecell[l]{DS} & \makecell[l]{Distribution \\ Plots} \\ 
        
        \makecell[l]{What is the \\ output of each  \\ feature engineering method?} & \makecell[l]{Intermediate \\ Visualization} & \makecell[l]{DS} & \makecell[l]{Table \\ Visualization} \\ 
        
        \makecell[l]{What do the \\ distributions of the new \\ features compare to \\ the original features?} & \makecell[l]{Intermediate \\ Visualization} & \makecell[l]{DS} & \makecell[l]{Distribution \\ Plots} \\ 
        
        \makecell[l]{What does the \\ pipeline look like \\ in general?} & \makecell[l]{Pipeline \\ Visualization} & \makecell[l]{DS} & \makecell[l]{SSG \\ \citep{zoller2023xautoml} } \\ 
        
        \makecell[l]{What does the \\ best pipeline look like?} & \makecell[l]{Pipeline \\ Visualization} & \makecell[l]{DS} & \makecell[l]{CPCP \\ \citep{weidele2020autoaiviz} \\ \citep{park2020hypertendril} } \\ 
        
        \makecell[l]{How does the \\ best pipeline compare \\ to other pipelines?} & \makecell[l]{Pipeline \\ Visualization} & \makecell[l]{DS} & \makecell[l]{Dashboard \\ with CPCPs} \\ 
    \toprule

\end{longtable}
\end{small}

\subsection{HXAI Agent's Information Aggregation}

This component identifies the root cause of errors and provides actionable advice to improve the data, model and learning process. The information  aggregation uses explainability to allow for improved ML performance. The main question that this component answers is: \emph{``Are errors predominantly due to model choice, data issues, or training flaws?''}.  Table \ref{tbl:unified-questions} summarizes the questions in this component.

\begin{small}
\renewcommand{\arraystretch}{3}
\begin{longtable}{llll}
    \caption{HXAI Agent based questions \label{tbl:unified-questions}}\\
    \toprule
    \textbf{Questions} & \textbf{Functionality} & \textbf{User} & \textbf{Methods} \\
        \makecell[l]{Is the data imbalance \\ handled by the pipeline?} & \makecell[l]{Data } & \makecell[l]{DA \\ DS} & \makecell[l]{AI explanation \\ using data,learning process,\\ and performance explanations} \\ 
        
        \makecell[l]{How can the \\ imbalance be handled \\ by the pipeline?} & \makecell[l]{Data } & \makecell[l]{DS} & \makecell[l]{ AI explanation \\ SMOTE \\ \citep{chawla2002smote} \\ image augmentation \\ \citep{bloice2017augmentor} \\ re-weighting } \\ 
        
        \makecell[l]{Is sample collection \\ needed for subpopulations?} & \makecell[l]{Data } & \makecell[l]{DA \\ DS} & \makecell[l]{AI explanation \\ using learning process \\ performance quality explanations} \\ 
        
        \makecell[l]{Is the missing data \\ handled by the pipeline?} & \makecell[l]{Data} & \makecell[l]{DA \\ DS} & \makecell[l]{AI explanation \\ using data,learning process,\\ and model quality explanations} \\ 
        
        \makecell[l]{How can missing \\ data be fixed?} & \makecell[l]{Data} & \makecell[l]{DS} & \makecell[l]{AI explanation \\ suggesting imputation methods \\  using data explanations} \\ 
        
        \makecell[l]{Are outliers handled\\  by the pipeline?} & \makecell[l]{Data} & \makecell[l]{DA \\ DS} & \makecell[l]{AI explanation \\ using learning process \\ model quality explanations} \\ 
        
        \makecell[l]{How can outliers be \\ handled by the pipeline?} & \makecell[l]{Data} & \makecell[l]{DS} & \makecell[l]{AI explanation \\ suggesting OD methods \\  using data explanations} \\ 
        
        \makecell[l]{Are mislabeled data \\ handled by the pipeline?} & \makecell[l]{Data} & \makecell[l]{DA \\ DS} & \makecell[l]{AI explanation \\ using learning process \\ model output \\ model quality explanations} \\ 
        
        \makecell[l]{How can mislabeled\\ data be handled?} & \makecell[l]{Data} & \makecell[l]{DS} & \makecell[l]{AI explanation \\ suggesting Mislabel \\ correction methods} \\ 
        
        \makecell[l]{Are there redundant \\ or highly correlated \\ features beneficial?} & \makecell[l]{Data} & \makecell[l]{DS} & \makecell[l]{AI explanation \\ using learning process \\ model output \\ model quality explanations} \\ 
        
        \makecell[l]{Are redundant features\\  handled by the pipeline?} & \makecell[l]{Data} & \makecell[l]{DA \\ DS} & \makecell[l]{AI explanation \\ using learning process \\ model output explanations} \\ 
        
        \makecell[l]{Do additional features \\ need to be collected?} & \makecell[l]{Data} & \makecell[l]{DA \\ DS} & \makecell[l]{AI explanation \\ using learning process \\ model quality explanations} \\ 
        
        \makecell[l]{Can the objective \\ function  be refined to \\ improve performance?} & \makecell[l]{Analysis \\ Setup.} & \makecell[l]{DS} & \makecell[l]{AI explanation \\ using learning process \\ analysis setup explanations} \\ 
        
        \makecell[l]{Is the current \\ metric appropriate?} & \makecell[l]{Analysis \\ Setup.} & \makecell[l]{All} & \makecell[l]{AI explanation \\ using data \\ analysis setup explanations} \\ 
        
        \makecell[l]{Is the estimation \\ protocol appropriate?} & \makecell[l]{Analysis \\ Setup.} & \makecell[l]{DA \\ DS} & \makecell[l]{AI explanation \\ using data \\ analysis setup explanations} \\ 
        
        \makecell[l]{What are points of \\ failure of the model?} & \makecell[l]{Model \\ Quality} & \makecell[l]{DA \\ DS} & \makecell[l]{AI explanation \\ using model output \\ model quality  explanations} \\ 
        
        \makecell[l]{Are fairness concerns\\  handled by the pipeline?} & \makecell[l]{Model \\ Quality} & \makecell[l]{DA \\ DS} & \makecell[l]{AI explanation \\ using learning process \\ model quality explanations} \\ 
        
        \makecell[l]{How can fairness \\ problems be mitigated?} & \makecell[l]{Model \\ Quality} & \makecell[l]{DS} & \makecell[l]{AI explanation \\ suggesting fairness \\ mitigation methods \\ \citep{bdcc7010015}} \\ 
        
        \makecell[l]{Is the model \\ overfitting?} & \makecell[l]{Model \\ Quality} & \makecell[l]{DA \\ DS} & \makecell[l]{AI explanation \\ using learning process \\ model quality explanations} \\ 
        
        \makecell[l]{Are feature \\ transformations beneficial?} & \makecell[l]{Model \\ Quality} & \makecell[l]{DS} & \makecell[l]{AI explanation \\ using learning process \\ model quality explanations} \\ 
        
        \makecell[l]{Can Feature Selection \\ simplify the model without \\ sacrificing accuracy?} & \makecell[l]{Model \\ Quality} & \makecell[l]{DA \\ DS} & \makecell[l]{AI explanation \\ using learning process \\ model quality explanations} \\ 
        
        \makecell[l]{Which feature engineering \\ techniques can improve \\ the performance?} & \makecell[l]{Model \\ Quality} & \makecell[l]{DS} & \makecell[l]{AI explanation \\ suggesting feature \\ transformation methods \\ \citep{khalid2014survey}} \\ 
        
        \makecell[l]{Does search space \\ need refinement?} & \makecell[l]{Learning \\ Process} & \makecell[l]{DS} & \makecell[l]{AI explanation \\ using learning process \\ analysis setup explanations} \\ 
        
        \makecell[l]{Do more predictive \\ models needed?} & \makecell[l]{Learning \\ Process} & \makecell[l]{DS} & \makecell[l]{AI explanation \\ using learning process \\ model quality explanations} \\
        
        \makecell[l]{Are important \\ hyper-parameters missing?} & \makecell[l]{Learning \\ Process} & \makecell[l]{DS} & \makecell[l]{AI explanation \\ using learning process \\ analysis setup explanations} \\ 
        
        \makecell[l]{Are hyper-parameters \\ wrongly optimized?} & \makecell[l]{Learning \\ Process} & \makecell[l]{DS} & \makecell[l]{AI explanation \\ using learning process \\ analysis setup explanations} \\ 
        
        \makecell[l]{How can the optimization \\ algorithm improve?} & \makecell[l]{Learning \\ Process} & \makecell[l]{DS} & \makecell[l]{AI explanation \\ suggesting alternative \\ optimization processes} \\ 
\end{longtable}
\end{small}

\end{appendices}

\section{Software tools}

In this section, we include the tables that match questions from the Question bank against the capabilities of currently available tools. See Tables \ref{data-tbl},  \ref{analysis-setup-tbl}, \ref{model-predictions-tbl}, \ref{model-performance-tbl}, \ref{learning-process-tbl}, \ref{unified-expl-tbl}.

\begin{sidewaystable}[!htbp]
    \centering
    \tiny
    \caption{Table that presents whether each data question is answered by software. \label{data-tbl}}

    \renewcommand{\arraystretch}{1} % Adjust row spacing
    \setlength{\tabcolsep}{1pt}       % Adjust column spacing
    %\scalebox{0.7}{ % Scale down by 80%
    %\begin{adjustbox}{max width=0.5\textwidth}
    %\resizebox{\textwidth}{!}{ % Scales to fit within text width
    %\adjustbox{valign=c, scale=0.6}{
    \begin{tabular*}{\textwidth}{@{\extracolsep\fill}l*{23}{c}}
    \toprule
        \multirow{2}{*}{\textbf{Question}} 
         & \multicolumn{10}{c}{\textbf{Open XAI}} 
         & \multicolumn{4}{c}{\textbf{Comm. XAI}} 
         & \multicolumn{5}{c}{\textbf{Open AutoML}} 
         & \multicolumn{4}{c}{\textbf{Comm. AutoML}} \\
        \cmidrule(lr){2-11} \cmidrule(lr){12-15} \cmidrule(lr){16-20} \cmidrule(lr){21-24}
         & \rotatebox{90}{Intel-XAI-Tools} 
         & \rotatebox{90}{EthicalML's XAI} 
         & \rotatebox{90}{OmniXAI} 
         & \rotatebox{90}{Explainer} 
         & \rotatebox{90}{Responsible AI Toolbox} 
         & \rotatebox{90}{Pyreal} 
         & \rotatebox{90}{AI Explainability 360} 
         & \rotatebox{90}{Alibi} 
         & \rotatebox{90}{Captum} 
         & \rotatebox{90}{Dalex} 
         & \rotatebox{90}{ArizeAI} 
         & \rotatebox{90}{CensiousAI} 
         & \rotatebox{90}{WhyLabs} 
         & \rotatebox{90}{FiddlerAI} 
         & \rotatebox{90}{MLJAR} 
         & \rotatebox{90}{AutoSklearn} 
         & \rotatebox{90}{H20 Automl} 
         & \rotatebox{90}{FLAML} 
         & \rotatebox{90}{AutoGluon} 
         & \rotatebox{90}{DriverlessAI} 
         & \rotatebox{90}{DataRobot} 
         & \rotatebox{90}{JADBio} 
         & \rotatebox{90}{Amazon Sagemaker AutoPilot} \\
         \midrule

        How are the features and target distributed? 
        & \xmark & \checkmark & \checkmark & \xmark & \checkmark & \xmark & \xmark & \xmark & \xmark & \xmark 
         & \checkmark & \checkmark & \checkmark & \checkmark 
         & \xmark & \xmark & \xmark & \xmark & \xmark 
         & \checkmark & \checkmark & \checkmark & \checkmark \\[2mm]
    
         Are outliers visualized?  
         & \xmark & \xmark & \xmark & \xmark & \xmark & \xmark & \xmark & \xmark & \xmark & \xmark 
         & \xmark & \xmark & \checkmark & \xmark 
         & \xmark & \xmark & \xmark & \xmark & \xmark 
         & \checkmark & \checkmark & \checkmark & \xmark \\[2mm]

         How do pairs of features/target relate visually?
         & \xmark & \xmark & \checkmark & \xmark & \checkmark & \xmark & \xmark & \xmark & \xmark & \xmark 
         & \xmark & \xmark & \checkmark & \xmark 
         & \xmark & \xmark & \xmark & \xmark & \xmark 
         & \checkmark & \checkmark & \xmark & \xmark \\[2mm]
    
         How features are interacting? 
         & \xmark & \xmark & \xmark & \xmark & \xmark & \xmark & \xmark & \xmark & \xmark & \xmark 
         & \xmark & \xmark & \xmark & \xmark 
         & \xmark & \xmark & \xmark & \xmark & \xmark 
         & \checkmark & \checkmark & \xmark & \xmark \\[2mm]

         Is the data separable on a latent space? 
         & \xmark & \xmark & \xmark & \xmark & \xmark & \xmark & \xmark & \xmark & \xmark & \xmark 
         & \xmark & \xmark & \xmark & \xmark 
         & \xmark & \xmark & \xmark & \xmark & \xmark 
         & \xmark & \xmark & \xmark & \xmark \\[2mm]
         
         What are the dimensions of the dataset? 
         & \xmark & \xmark & \xmark & \xmark & \checkmark & \xmark & \xmark & \xmark & \xmark & \checkmark 
         & \checkmark & \checkmark & \checkmark & \checkmark 
         & \xmark & \xmark & \xmark & \xmark & \checkmark 
         & \checkmark & \checkmark & \checkmark & \checkmark \\[2mm]
         
         What are the data types of each feature? 
         & \xmark & \xmark & \xmark & \xmark & \xmark & \xmark & \xmark & \xmark & \xmark & \xmark 
         & \checkmark & \checkmark & \checkmark & \checkmark 
         & \xmark & \xmark & \xmark & \xmark & \checkmark 
         & \checkmark & \checkmark & \checkmark & \checkmark \\[2mm]
         
         Is the data containing sensitive information? 
         & \xmark & \xmark & \xmark & \xmark & \xmark & \xmark & \xmark & \xmark & \xmark & \xmark 
         & \xmark & \xmark & \xmark & \xmark 
         & \xmark & \xmark & \xmark & \xmark & \xmark 
         & \checkmark & \xmark & \xmark & \xmark \\[2mm]

         What are the distributions of each feature/target?
         & \xmark & \xmark & \xmark & \xmark & \xmark & \xmark & \xmark & \xmark & \xmark & \xmark 
         & \checkmark & \checkmark & \checkmark & \checkmark 
         & \xmark & \xmark & \xmark & \xmark & \xmark 
         & \checkmark & \checkmark & \xmark & \checkmark \\[2mm]

        Which features have missing values, and what percentage is missing for each? 
         & \xmark & \xmark & \xmark & \xmark & \xmark & \xmark & \xmark & \xmark & \xmark & \xmark 
         & \checkmark & \checkmark & \checkmark & \checkmark 
         & \xmark & \xmark & \xmark & \xmark & \xmark 
         & \checkmark & \checkmark & \checkmark & \checkmark \\[2mm]
         
         How are missing values distributed? 
         & \xmark & \xmark & \xmark & \xmark & \xmark & \xmark & \xmark & \xmark & \xmark & \xmark 
         & \checkmark & \checkmark & \xmark & \xmark 
         & \xmark & \xmark & \xmark & \xmark & \xmark 
         & \checkmark & \checkmark & \xmark & \checkmark \\[2mm]
         
         Are there any patterns in the missing data? 
         & \xmark & \xmark & \xmark & \xmark & \xmark & \xmark & \xmark & \xmark & \xmark & \xmark 
         & \xmark & \xmark & \xmark & \xmark 
         & \xmark & \xmark & \xmark & \xmark & \xmark 
         & \xmark & \xmark & \xmark & \xmark \\[2mm]
         
         Is the target variable imbalanced? 
         & \xmark & \checkmark & \checkmark & \xmark & \xmark & \xmark & \xmark & \xmark & \xmark & \xmark 
         & \xmark & \xmark & \xmark & \xmark 
         & \xmark & \xmark & \xmark & \xmark & \xmark 
         & \checkmark & \xmark & \xmark & \checkmark \\[2mm]
         
         Are there under-represented populations? 
         & \xmark & \checkmark & \xmark & \xmark & \checkmark & \xmark & \xmark & \xmark & \xmark & \xmark 
         & \xmark & \xmark & \xmark & \xmark 
         & \xmark & \xmark & \xmark & \xmark & \xmark 
         & \xmark & \xmark & \xmark & \xmark \\[2mm]

         Are there duplicate or highly correlated samples/ features?
         & \xmark & \xmark & \xmark & \xmark & \xmark & \xmark & \xmark & \xmark & \xmark & \xmark 
         & \xmark & \xmark & \xmark & \xmark 
         & \xmark & \xmark & \xmark & \xmark 
         & \checkmark & \checkmark & \checkmark & \checkmark \\[2mm]

         Are there any noticeable outliers? 
         & \xmark & \xmark & \xmark & \xmark & \xmark & \xmark & \xmark & \xmark & \xmark & \xmark 
         & \checkmark & \checkmark & \checkmark & \xmark 
         & \xmark & \xmark & \xmark & \xmark & \xmark 
         & \checkmark & \xmark & \xmark & \checkmark \\[2mm]

         Are features linearly or non-linearly correlated? 
         & \xmark & \xmark & \xmark & \xmark & \xmark & \xmark & \xmark & \xmark & \xmark & \xmark 
         & \xmark & \xmark & \xmark & \xmark 
         & \xmark & \xmark & \xmark & \xmark & \xmark 
         & \xmark & \xmark & \xmark & \checkmark \\[2mm]
         
         Are there meaningful transformations that could make relationships more effective or interpretable? 
         & \xmark & \xmark & \xmark & \xmark & \xmark & \xmark & \xmark & \xmark & \xmark & \xmark 
         & \xmark & \xmark & \xmark & \xmark 
         & \xmark & \xmark & \xmark & \xmark & \xmark 
         & \xmark & \checkmark & \xmark & \xmark \\[2mm]

         Do any observed correlations suggest potential causation? 
         & \xmark & \xmark & \xmark & \xmark & \checkmark & \xmark & \xmark & \xmark & \xmark & \xmark 
         & \xmark & \xmark & \xmark & \xmark 
         & \xmark & \xmark & \xmark & \xmark & \xmark 
         & \xmark & \xmark & \xmark & \xmark \\[2mm]

         Can the data be split into clusters? 
         & \xmark & \xmark & \xmark & \xmark & \xmark & \xmark & \xmark & \xmark & \xmark & \xmark 
         & \xmark & \xmark & \xmark & \xmark 
         & \xmark & \xmark & \xmark & \xmark & \xmark 
         & \checkmark & \checkmark & \xmark & \xmark \\[2mm]
         
         Are there any consistent patterns within specific populations? 
         & \xmark & \xmark & \xmark & \xmark & \checkmark & \xmark & \xmark & \xmark & \xmark & \xmark 
         & \xmark & \xmark & \xmark & \xmark 
         & \xmark & \xmark & \xmark & \xmark & \xmark 
         & \xmark & \xmark & \xmark & \xmark \\
        \bottomrule
    \end{tabular*}
    %}
    %}\
    %}
    %\end{adjustbox}
\end{sidewaystable}

\begin{sidewaystable}[!htbp]
    \centering
    \tiny
    \caption{Table that presents whether each analysis setup question is answered by software. \label{analysis-setup-tbl}}
    
    \renewcommand{\arraystretch}{1} % Adjust row spacing
    \setlength{\tabcolsep}{1pt}       % Adjust column spacing
    %\scalebox{0.7}{ % Scale down by 80%
    %\begin{adjustbox}{max width=0.5\textwidth}
    %\resizebox{\textwidth}{!}{ % Scales to fit within text width
    %\adjustbox{valign=c, scale=0.6}{
    \begin{tabular*}{\textwidth}{@{\extracolsep\fill}l*{23}{c}}
    \toprule
        \multirow{2}{*}{\textbf{Question}} 
         & \multicolumn{10}{c}{\textbf{Open XAI}} 
         & \multicolumn{4}{c}{\textbf{Comm. XAI}} 
         & \multicolumn{5}{c}{\textbf{Open AutoML}} 
         & \multicolumn{4}{c}{\textbf{Comm. AutoML}} \\
        \cmidrule(lr){2-11} \cmidrule(lr){12-15} \cmidrule(lr){16-20} \cmidrule(lr){21-24}
         & \rotatebox{90}{Intel-XAI-Tools} 
         & \rotatebox{90}{EthicalML's XAI} 
         & \rotatebox{90}{OmniXAI} 
         & \rotatebox{90}{Explainer} 
         & \rotatebox{90}{Responsible AI Toolbox} 
         & \rotatebox{90}{Pyreal} 
         & \rotatebox{90}{AI Explainability 360} 
         & \rotatebox{90}{Alibi} 
         & \rotatebox{90}{Captum} 
         & \rotatebox{90}{Dalex} 
         & \rotatebox{90}{ArizeAI} 
         & \rotatebox{90}{CensiousAI} 
         & \rotatebox{90}{WhyLabs} 
         & \rotatebox{90}{FiddlerAI} 
         & \rotatebox{90}{MLJAR} 
         & \rotatebox{90}{AutoSklearn} 
         & \rotatebox{90}{H20 Automl} 
         & \rotatebox{90}{FLAML} 
         & \rotatebox{90}{AutoGluon} 
         & \rotatebox{90}{DriverlessAI} 
         & \rotatebox{90}{DataRobot} 
         & \rotatebox{90}{JADBio} 
         & \rotatebox{90}{Amazon Sagemaker AutoPilot} \\
         \midrule
        What is the type of problem? & \xmark & \xmark & \xmark & \xmark & \xmark & \xmark & \xmark & \xmark & \xmark & \xmark & \xmark & \xmark & \xmark & \xmark & \xmark & \xmark & \checkmark & \xmark & \checkmark & \checkmark & \checkmark & \checkmark & \checkmark \\  [2mm]
        What output can I expect from the model? & \xmark & \xmark & \xmark & \xmark & \xmark & \xmark & \xmark & \xmark & \xmark & \xmark & \xmark & \xmark & \xmark & \xmark & \xmark & \xmark & \xmark & \xmark & \xmark & \xmark & \xmark & \xmark & \xmark \\  [2mm]
        Which/Why did this validation protocol get selected? & \xmark & \xmark & \xmark & \xmark & \xmark & \xmark & \xmark & \xmark & \xmark & \xmark & \xmark & \xmark & \xmark & \xmark & \xmark & \xmark & \xmark & \xmark & \checkmark & \xmark & \xmark & \checkmark & \checkmark \\  [2mm]
        How does validation protocol works visually? & \xmark & \xmark & \xmark & \xmark & \xmark & \xmark & \xmark & \xmark & \xmark & \xmark & \xmark & \xmark & \xmark & \xmark & \xmark & \xmark & \xmark & \xmark & \xmark & \xmark & \xmark & \xmark & \xmark \\  [2mm]
        What is the metric to be optimized for this problem? & \xmark & \xmark & \xmark & \xmark & \xmark & \xmark & \xmark & \xmark & \xmark & \xmark & \xmark & \xmark & \xmark & \xmark & \xmark & \xmark & \xmark & \xmark & \checkmark & \checkmark & \checkmark & \checkmark & \checkmark \\  [2mm]
        Why is this metric suitable for this problem and what it focuses on? & \xmark & \xmark & \xmark & \xmark & \xmark & \xmark & \xmark & \xmark & \xmark & \xmark & \xmark & \xmark & \xmark & \xmark & \xmark & \xmark & \xmark & \xmark & \xmark & \xmark & \xmark & \checkmark & \checkmark \\  [2mm]
        Is the metric aligned with business goals? & \xmark & \xmark & \xmark & \xmark & \xmark & \xmark & \xmark & \xmark & \xmark & \xmark & \xmark & \xmark & \xmark & \xmark & \xmark & \xmark & \xmark & \xmark & \xmark & \xmark & \xmark & \xmark & \xmark \\ [2mm] 
        What is the success criteria for the task? & \xmark & \xmark & \xmark & \xmark & \xmark & \xmark & \xmark & \xmark & \xmark & \xmark & \xmark & \xmark & \xmark & \xmark & \xmark & \xmark & \xmark & \xmark & \xmark & \xmark & \xmark & \xmark & \xmark \\  [2mm]
        Which steps will be included in the pipeline and why? & \xmark & \xmark & \xmark & \xmark & \xmark & \xmark & \xmark & \xmark & \xmark & \xmark & \xmark & \xmark & \xmark & \xmark & \xmark & \xmark & \xmark & \xmark & \checkmark & \xmark & \checkmark & \checkmark & \checkmark \\  [2mm]
        Which algorithms will be included in each step and why? & \xmark & \xmark & \xmark & \xmark & \xmark & \xmark & \xmark & \xmark & \xmark & \xmark & \xmark & \xmark & \xmark & \xmark & \xmark & \xmark & \xmark & \xmark & \checkmark & \xmark & \checkmark & \checkmark & \checkmark \\  [2mm]
        Which hyper-parameters will be tuned for each algorithm and why? & \xmark & \xmark & \xmark & \xmark & \xmark & \xmark & \xmark & \xmark & \xmark & \xmark & \xmark & \xmark & \xmark & \xmark & \xmark & \xmark & \xmark & \xmark & \checkmark & \xmark & \xmark & \checkmark & \checkmark \\  [2mm]   
        \bottomrule
    \end{tabular*}
    %}
    %}\
    %}
    %\end{adjustbox}
    
\end{sidewaystable}

\begin{sidewaystable}[!h]
    \centering
    \tiny
    \caption{Table that presents whether each model outputs question is answered by software. \label{model-predictions-tbl}}
    
    \renewcommand{\arraystretch}{1} % Adjust row spacing
    \setlength{\tabcolsep}{1pt}       % Adjust column spacing
    %\scalebox{0.7}{ % Scale down by 80%
    %\begin{adjustbox}{max width=0.5\textwidth}
    %\resizebox{\textwidth}{!}{ % Scales to fit within text width
    %\adjustbox{valign=c, scale=0.6}{
    \begin{tabular*}{\textwidth}{@{\extracolsep\fill}l*{23}{c}}
    \toprule
        \multirow{2}{*}{\textbf{Question}} 
         & \multicolumn{10}{c}{\textbf{Open XAI}} 
         & \multicolumn{4}{c}{\textbf{Comm. XAI}} 
         & \multicolumn{5}{c}{\textbf{Open AutoML}} 
         & \multicolumn{4}{c}{\textbf{Comm. AutoML}} \\
        \cmidrule(lr){2-11} \cmidrule(lr){12-15} \cmidrule(lr){16-20} \cmidrule(lr){21-24}
         & \rotatebox{90}{Intel-XAI-Tools} 
         & \rotatebox{90}{EthicalML's XAI} 
         & \rotatebox{90}{OmniXAI} 
         & \rotatebox{90}{Explainer} 
         & \rotatebox{90}{Responsible AI Toolbox} 
         & \rotatebox{90}{Pyreal} 
         & \rotatebox{90}{AI Explainability 360} 
         & \rotatebox{90}{Alibi} 
         & \rotatebox{90}{Captum} 
         & \rotatebox{90}{Dalex} 
         & \rotatebox{90}{ArizeAI} 
         & \rotatebox{90}{CensiousAI} 
         & \rotatebox{90}{WhyLabs} 
         & \rotatebox{90}{FiddlerAI} 
         & \rotatebox{90}{MLJAR} 
         & \rotatebox{90}{AutoSklearn} 
         & \rotatebox{90}{H20 Automl} 
         & \rotatebox{90}{FLAML} 
         & \rotatebox{90}{AutoGluon} 
         & \rotatebox{90}{DriverlessAI} 
         & \rotatebox{90}{DataRobot} 
         & \rotatebox{90}{JADBio} 
         & \rotatebox{90}{Amazon Sagemaker AutoPilot} \\
        \midrule
         Which training instances most influenced a specific prediction? & \xmark & \xmark & \xmark & \xmark & \xmark & \xmark & \xmark & \xmark & \checkmark & \xmark & \xmark & \xmark & \xmark & \xmark & \xmark & \xmark & \xmark & \xmark & \xmark & \xmark & \xmark & \xmark & \xmark \\  [2mm]
        Were influencing instances similar or dissimilar to the test instance? & \xmark & \xmark & \xmark & \xmark & \xmark & \xmark & \xmark & \xmark & \xmark & \xmark & \xmark & \xmark & \xmark & \xmark & \xmark & \xmark & \xmark & \xmark & \xmark & \xmark & \xmark & \xmark & \xmark \\  [2mm]
        How does the model perform on “prototypes” or or “criticisms”? & \xmark & \xmark & \xmark & \xmark & \xmark & \xmark & \xmark & \xmark & \xmark & \xmark & \xmark & \xmark & \xmark & \xmark & \xmark & \xmark & \xmark & \xmark & \xmark & \xmark & \xmark & \xmark & \xmark \\  [2mm]
        What small, plausible changes to an instance would lead to a different prediction? & \xmark & \xmark & \checkmark & \xmark & \checkmark & \xmark & \checkmark & \checkmark & \xmark & \xmark & \xmark & \xmark & \xmark & \xmark & \xmark & \xmark & \xmark & \xmark & \xmark & \xmark & \checkmark & \checkmark & \xmark \\  [2mm]
        Is the model very sensitive to input changes? & \xmark & \xmark & \xmark & \xmark & \xmark & \xmark & \xmark & \xmark & \xmark & \xmark & \xmark & \xmark & \xmark & \xmark & \xmark & \xmark & \xmark & \xmark & \xmark & \xmark & \checkmark & \xmark & \xmark \\  [2mm]
        What changes can I make, without changing my prediction? & \xmark & \xmark & \xmark & \checkmark & \xmark & \xmark & \checkmark & \checkmark & \xmark & \xmark & \xmark & \xmark & \xmark & \xmark & \xmark & \xmark & \xmark & \xmark & \xmark & \xmark & \xmark & \xmark & \xmark \\  [2mm]
        
        For a given prediction, which features influenced the outcome and how much? & \checkmark & \xmark & \checkmark & \checkmark & \checkmark & \checkmark & \checkmark & \checkmark & \checkmark & \checkmark & \checkmark & \checkmark & \xmark & \checkmark & \xmark & \xmark & \checkmark & \xmark & \xmark & \checkmark & \checkmark & \xmark & \checkmark \\  [2mm]
        How does the local feature importance compare with the global feature importance? & \xmark & \xmark & \xmark & \xmark & \xmark & \xmark & \xmark & \xmark & \xmark & \xmark & \xmark & \xmark & \xmark & \xmark & \xmark & \xmark & \xmark & \xmark & \xmark & \xmark & \xmark & \xmark & \xmark \\ [2mm]
        How does a prediction change when we change the values of each feature? & \xmark & \xmark & \checkmark & \checkmark & \checkmark & \xmark & \xmark & \checkmark & \xmark & \checkmark & \xmark & \xmark & \xmark & \xmark & \checkmark & \checkmark * & \checkmark & \xmark & \xmark & \checkmark & \checkmark & \checkmark & \xmark \\  [2mm]
        How does feature importance change across different subgroups? & \xmark & \xmark & \xmark & \xmark & \checkmark & \xmark & \xmark & \xmark & \xmark & \xmark & \checkmark & \checkmark & \xmark & \xmark & \xmark & \xmark & \xmark & \xmark & \xmark & \checkmark & \checkmark & \xmark & \xmark \\  [2mm]
        Do feature interactions within a cohort significantly impact predictions? & \xmark & \xmark & \xmark & \xmark & \checkmark & \xmark & \xmark & \xmark & \xmark & \xmark & \xmark & \xmark & \xmark & \xmark & \xmark & \xmark & \xmark & \xmark & \xmark & \xmark & \xmark & \xmark & \xmark \\  [2mm]
        Are certain features more important for predicting specific classes than others? & \xmark & \xmark & \xmark & \xmark & \checkmark & \xmark & \xmark & \xmark & \xmark & \xmark & \xmark & \xmark & \xmark & \xmark & \xmark & \xmark & \xmark & \xmark & \xmark & \checkmark & \xmark & \xmark & \xmark \\  [2mm]

        Which features have the most influence on the model’s overall predictions? & \checkmark & \checkmark & \checkmark & \checkmark & \checkmark & \checkmark & \checkmark & \checkmark & \xmark & \checkmark & \checkmark & \checkmark & \checkmark & \checkmark & \checkmark & \checkmark * & \checkmark & \checkmark & \checkmark & \checkmark & \checkmark & \checkmark & \checkmark \\  [2mm]
        Are there specific interactions between features that significantly impact predictions? & \xmark & \xmark & \xmark & \xmark & \xmark & \xmark & \xmark & \xmark & \xmark & \xmark & \xmark & \xmark & \xmark & \xmark & \xmark & \xmark & \xmark & \xmark & \xmark & \xmark & \checkmark & \xmark & \xmark \\  [2mm]
        How does the average prediction of a model change when the values of each feature change? & \xmark & \xmark & \checkmark & \xmark & \checkmark & \checkmark & \xmark & \checkmark & \xmark & \xmark & \xmark & \xmark & \xmark & \checkmark & \checkmark & \checkmark * & \checkmark & \xmark & \xmark & \checkmark & \checkmark & \checkmark & \checkmark \\  [2mm]
        How consistent are the important features across different models? & \xmark & \xmark & \xmark & \xmark & \xmark & \xmark & \xmark & \xmark & \xmark & \xmark & \xmark & \xmark & \xmark & \xmark & \xmark & \xmark & \xmark & \xmark & \xmark & \xmark & \xmark & \xmark & \xmark \\  [2mm]
        What is a simple visualization of my model? & \xmark & \xmark & \xmark & \checkmark & \checkmark & \xmark & \xmark & \checkmark & \xmark & \checkmark & \checkmark & \xmark & \xmark & \checkmark & \checkmark & \xmark & \xmark & \xmark & \xmark & \xmark & \checkmark & \xmark & \xmark \\  [2mm] 
    \bottomrule
    \end{tabular*}
    %}
    %}\
    %}
    %\end{adjustbox}
    
\end{sidewaystable}

\begin{sidewaystable}[h]
    \centering
    \tiny
    \caption{Table that presents whether each model performance question is answered by software. \label{model-performance-tbl}}
    \renewcommand{\arraystretch}{1} % Adjust row spacing
    \setlength{\tabcolsep}{1pt}       % Adjust column spacing
    %\scalebox{0.7}{ % Scale down by 80%
    %\begin{adjustbox}{max width=0.5\textwidth}
    %\resizebox{\textwidth}{!}{ % Scales to fit within text width
    %\adjustbox{valign=c, scale=0.6}{
    \begin{tabular*}{\textwidth}{@{\extracolsep\fill}l*{23}{c}}
    \toprule
        \multirow{2}{*}{\textbf{Question}} 
         & \multicolumn{10}{c}{\textbf{Open XAI}} 
         & \multicolumn{4}{c}{\textbf{Comm. XAI}} 
         & \multicolumn{5}{c}{\textbf{Open AutoML}} 
         & \multicolumn{4}{c}{\textbf{Comm. AutoML}} \\
        \cmidrule(lr){2-11} \cmidrule(lr){12-15} \cmidrule(lr){16-20} \cmidrule(lr){21-24}
         & \rotatebox{90}{Intel-XAI-Tools} 
         & \rotatebox{90}{EthicalML's XAI} 
         & \rotatebox{90}{OmniXAI} 
         & \rotatebox{90}{Explainer} 
         & \rotatebox{90}{Responsible AI Toolbox} 
         & \rotatebox{90}{Pyreal} 
         & \rotatebox{90}{AI Explainability 360} 
         & \rotatebox{90}{Alibi} 
         & \rotatebox{90}{Captum} 
         & \rotatebox{90}{Dalex} 
         & \rotatebox{90}{ArizeAI} 
         & \rotatebox{90}{CensiousAI} 
         & \rotatebox{90}{WhyLabs} 
         & \rotatebox{90}{FiddlerAI} 
         & \rotatebox{90}{MLJAR} 
         & \rotatebox{90}{AutoSklearn} 
         & \rotatebox{90}{H20 Automl} 
         & \rotatebox{90}{FLAML} 
         & \rotatebox{90}{AutoGluon} 
         & \rotatebox{90}{DriverlessAI} 
         & \rotatebox{90}{DataRobot} 
         & \rotatebox{90}{JADBio} 
         & \rotatebox{90}{Amazon Sagemaker AutoPilot} \\
    \midrule
         How does the performance curves look like? & \checkmark & \checkmark & \checkmark & \xmark & \checkmark & \xmark & \xmark & \xmark & \xmark & \checkmark & \xmark & \xmark & \checkmark & \checkmark & \xmark & \xmark & \xmark & \xmark & \xmark & \checkmark & \checkmark & \checkmark & \checkmark \\  [2mm]

        What does the calibration curve look like? & \xmark & \xmark & \xmark & \xmark & \xmark & \xmark & \xmark & \checkmark & \xmark & \xmark & \xmark & \xmark & \xmark & \checkmark & \xmark & \xmark & \xmark & \xmark & \xmark & \xmark & \xmark & \xmark & \xmark \\  [2mm]
        
        In which metrics does the model underperform? & \xmark & \xmark & \xmark & \xmark & \xmark & \xmark & \xmark & \xmark & \xmark & \xmark & \xmark & \xmark & \xmark & \xmark & \xmark & \xmark & \xmark & \xmark & \xmark & \xmark & \xmark & \xmark & \xmark \\ [2mm]
        
        What are the most common types of errors and which classes are most affected? & \checkmark & \xmark & \checkmark & \xmark & \checkmark & \xmark & \xmark & \xmark & \xmark & \xmark & \checkmark & \xmark & \checkmark & \checkmark & \checkmark & \xmark & \checkmark & \xmark & \xmark & \checkmark & \checkmark & \xmark & \checkmark \\  [2mm]
        
        For each error type, can specific patterns or attributes in the misclassified samples be identified? & \xmark & \xmark & \xmark & \xmark & \xmark & \xmark & \xmark & \xmark & \xmark & \xmark & \checkmark & \xmark & \xmark & \xmark & \xmark & \xmark & \xmark & \xmark & \xmark & \xmark & \xmark & \xmark & \xmark \\  [2mm]
        
        Are there certain ranges/values of features where the model consistently performs poorly? 
        & \xmark & \xmark & \xmark & \xmark & \checkmark & \xmark & \xmark & \xmark & \xmark & \xmark & \checkmark & \xmark & \xmark & \xmark & \xmark & \xmark & \xmark & \xmark & \xmark & \xmark & \xmark & \xmark & \xmark \\ [2mm]

        Can errors be clustered into groups with similar characteristics, and if so, what patterns define these groups? & \xmark & \xmark & \xmark & \xmark & \xmark & \xmark & \xmark & \xmark & \xmark & \xmark & \xmark & \xmark & \xmark & \xmark & \xmark & \xmark & \xmark & \xmark & \xmark & \xmark & \xmark & \xmark & \xmark \\ [2mm]
        
        How does model performance vary across different groups? & \xmark & \checkmark & \checkmark & \xmark & \checkmark & \xmark & \xmark & \xmark & \xmark & \checkmark & \checkmark & \xmark & \checkmark & \checkmark & \xmark & \xmark & \xmark & \xmark & \xmark & \checkmark & \checkmark & \xmark & \xmark \\ [2mm]
        
        Is the model’s probabilities calibrated across different groups? & \xmark & \xmark & \xmark & \xmark & \xmark & \xmark & \xmark & \xmark & \xmark & \xmark & \xmark & \xmark & \xmark & \xmark & \xmark & \xmark & \xmark & \xmark & \xmark & \xmark & \xmark & \xmark & \xmark \\  [2mm]
        
        Are there specific groups with consistently lower or higher performance? & \xmark & \xmark & \xmark & \xmark & \checkmark & \xmark & \xmark & \xmark & \xmark & \xmark & \xmark & \xmark & \xmark & \xmark & \xmark & \xmark & \xmark & \xmark & \xmark & \xmark & \xmark & \xmark & \xmark \\  [2mm]
        
        How does the model perform across various fairness metrics? & \xmark & \checkmark & \checkmark & \xmark & \checkmark & \xmark & \xmark & \xmark & \xmark & \checkmark & \checkmark & \xmark & \xmark & \checkmark & \checkmark & \xmark & \xmark & \xmark & \xmark & \checkmark & \checkmark & \xmark & \xmark \\ [2mm]
        
        How does models perform on different metrics? & \checkmark & \xmark & \checkmark & \xmark & \checkmark & \xmark & \xmark & \xmark & \xmark & \xmark & \checkmark & \checkmark & \checkmark & \xmark & \xmark & \xmark & \checkmark & \xmark & \checkmark & \checkmark & \checkmark & \checkmark & \checkmark \\ [2mm]
        
        How metric values vary across classes? & \checkmark & \xmark & \xmark & \xmark & \xmark & \xmark & \xmark & \xmark & \xmark & \xmark & \xmark & \xmark & \xmark & \xmark & \xmark & \xmark & \checkmark & \xmark & \xmark & \checkmark & \checkmark & \checkmark & \checkmark \\  [2mm]
        
        What is the best threshold for each metric? & \xmark & \xmark & \xmark & \xmark & \xmark & \xmark & \xmark & \xmark & \xmark & \xmark & \xmark & \xmark & \xmark & \xmark & \checkmark & \xmark & \xmark & \xmark & \xmark & \checkmark & \checkmark & \checkmark & \xmark \\  [2mm]
        
        Does the model produce well-calibrated probabilities? & \xmark & \xmark & \xmark & \xmark & \xmark & \xmark & \xmark & \xmark & \xmark & \xmark & \xmark & \xmark & \xmark & \xmark & \xmark & \xmark & \xmark & \xmark & \xmark & \xmark & \xmark & \checkmark & \xmark \\  [2mm]
        
        What are the confidence intervals for each metric? & \xmark & \xmark & \xmark & \xmark & \xmark & \xmark & \xmark & \xmark & \xmark & \xmark & \xmark & \xmark & \xmark & \xmark & \xmark & \xmark & \xmark & \xmark & \xmark & \xmark & \checkmark & \checkmark & \checkmark \\  [2mm]
        
        How does the performance of a (the best) model compare against another’s model performance? & \xmark & \xmark & \xmark & \xmark & \xmark & \xmark & \xmark & \xmark & \xmark & \xmark & \xmark & \xmark & \xmark & \xmark & \xmark & \xmark & \xmark & \xmark & \checkmark & \checkmark & \xmark & \checkmark & \checkmark \\
    \bottomrule
    \end{tabular*}
    %}
    %}\
    %}
    
    %\end{adjustbox}
\end{sidewaystable}

\begin{sidewaystable}[h]
    \centering
    \tiny
    \caption{Table that presents whether each learning process question is answered by software. \label{learning-process-tbl}}
    \renewcommand{\arraystretch}{1} % Adjust row spacing
    \setlength{\tabcolsep}{1pt}       % Adjust column spacing
    %\scalebox{0.7}{ % Scale down by 80%
    %\begin{adjustbox}{max width=0.5\textwidth}
    %\resizebox{\textwidth}{!}{ % Scales to fit within text width
    %\adjustbox{valign=c, scale=0.6}{
    
    \begin{tabular*}{\textwidth}{@{\extracolsep\fill}l*{23}{c}}
    \toprule
        \multirow{2}{*}{\textbf{Question}} 
         & \multicolumn{10}{c}{\textbf{Open XAI}} 
         & \multicolumn{4}{c}{\textbf{Comm. XAI}} 
         & \multicolumn{5}{c}{\textbf{Open AutoML}} 
         & \multicolumn{4}{c}{\textbf{Comm. AutoML}} \\
        \cmidrule(lr){2-11} \cmidrule(lr){12-15} \cmidrule(lr){16-20} \cmidrule(lr){21-24}
         & \rotatebox{90}{Intel-XAI-Tools} 
         & \rotatebox{90}{EthicalML's XAI} 
         & \rotatebox{90}{OmniXAI} 
         & \rotatebox{90}{Explainer} 
         & \rotatebox{90}{Responsible AI Toolbox} 
         & \rotatebox{90}{Pyreal} 
         & \rotatebox{90}{AI Explainability 360} 
         & \rotatebox{90}{Alibi} 
         & \rotatebox{90}{Captum} 
         & \rotatebox{90}{Dalex} 
         & \rotatebox{90}{ArizeAI} 
         & \rotatebox{90}{CensiousAI} 
         & \rotatebox{90}{WhyLabs} 
         & \rotatebox{90}{FiddlerAI} 
         & \rotatebox{90}{MLJAR} 
         & \rotatebox{90}{AutoSklearn} 
         & \rotatebox{90}{H20 Automl} 
         & \rotatebox{90}{FLAML} 
         & \rotatebox{90}{AutoGluon} 
         & \rotatebox{90}{DriverlessAI} 
         & \rotatebox{90}{DataRobot} 
         & \rotatebox{90}{JADBio} 
         & \rotatebox{90}{Amazon Sagemaker AutoPilot} \\
        \midrule
         How does the ML system explore the search space over time? & \xmark & \xmark & \xmark & \xmark & \xmark & \xmark & \xmark & \xmark & \xmark & \xmark & \xmark & \xmark & \xmark & \xmark & \xmark & \xmark & \xmark & \xmark & \xmark & \xmark & \xmark & \xmark & \xmark \\  [2mm]
         
        Does the ML process demonstrate a preference for specific algorithms? & \xmark & \xmark & \xmark & \xmark & \xmark & \xmark & \xmark & \xmark & \xmark & \xmark & \xmark & \xmark & \xmark & \xmark & \xmark & \xmark & \xmark & \xmark & \xmark & \xmark & \xmark & \xmark & \xmark \\  [2mm]
        
        How does the loss function change over training epochs? & \xmark & \xmark & \xmark & \xmark & \xmark & \xmark & \xmark & \xmark & \xmark & \xmark & \xmark & \xmark & \xmark & \xmark & \checkmark & \checkmark & \checkmark & \checkmark & \checkmark & \checkmark & \checkmark & \checkmark & \xmark \\  [2mm]
        
        Which hyper-parameters are the most important? & \xmark & \xmark & \xmark & \xmark & \xmark & \xmark & \xmark & \xmark & \xmark & \xmark & \xmark & \xmark & \xmark & \xmark & \xmark & \xmark & \xmark & \xmark & \xmark & \xmark & \xmark & \xmark & \xmark \\  [2mm]
        
        Which algorithms were trained on each step? & \xmark & \xmark & \xmark & \xmark & \xmark & \xmark & \xmark & \xmark & \xmark & \xmark & \xmark & \xmark & \xmark & \xmark & \xmark & \xmark & \xmark & \xmark & \xmark & \xmark & \xmark & \checkmark & \xmark \\  [2mm]
        
        Which hyperparameters were tuned, what is their distribution and ranges? & \xmark & \xmark & \xmark & \xmark & \xmark & \xmark & \xmark & \xmark & \xmark & \xmark & \xmark & \xmark & \xmark & \xmark & \xmark & \xmark & \xmark & \xmark & \xmark & \xmark & \xmark & \checkmark & \xmark \\  [2mm]
        
        How does one hyper-parameter value affect the performance result? & \xmark & \xmark & \xmark & \xmark & \xmark & \xmark & \xmark & \xmark & \xmark & \xmark & \xmark & \xmark & \xmark & \xmark & \xmark & \xmark & \xmark & \xmark & \xmark & \xmark & \xmark & \xmark & \xmark \\  [2mm]
        What model type, hyperparameters, and feature transformations did ML select as optimal? & \xmark & \xmark & \xmark & \xmark & \xmark & \xmark & \xmark & \xmark & \xmark & \xmark & \xmark & \xmark & \xmark & \xmark & \checkmark & \xmark & \xmark & \checkmark & \checkmark & \checkmark & \checkmark & \checkmark & \checkmark \\  [2mm]
        
        What were the final performance metrics, what is the training/prediction time, the pipeline size? & \xmark & \xmark & \xmark & \xmark & \xmark & \xmark & \xmark & \xmark & \xmark & \xmark & \xmark & \xmark & \xmark & \xmark & \checkmark & \xmark & \checkmark & \checkmark & \checkmark & \checkmark & \checkmark & \checkmark & \checkmark \\  [2mm]
        
        How does the best model’s summary compare against others? & \xmark & \xmark & \xmark & \xmark & \xmark & \xmark & \xmark & \xmark & \xmark & \xmark & \xmark & \xmark & \xmark & \xmark & \checkmark & \checkmark & \checkmark & \xmark & \checkmark & \checkmark & \checkmark & \checkmark & \checkmark \\  [2mm]
        
        What was the training objective? & \xmark & \xmark & \xmark & \xmark & \xmark & \xmark & \xmark & \xmark & \xmark & \xmark & \xmark & \xmark & \xmark & \xmark & \checkmark & \checkmark & \xmark & \xmark & \xmark & \checkmark & \checkmark & \checkmark & \checkmark \\  [2mm]
        
        Were there any errors during optimization? & \xmark & \xmark & \xmark & \xmark & \xmark & \xmark & \xmark & \xmark & \xmark & \xmark & \xmark & \xmark & \xmark & \xmark & \xmark & \checkmark & \xmark & \xmark & \xmark & \xmark & \xmark & \xmark & \xmark \\ [2mm]
        
        What is the output of each preprocessing method? & \xmark & \xmark & \xmark & \xmark & \xmark & \xmark & \xmark & \xmark & \xmark & \xmark & \xmark & \xmark & \xmark & \xmark & \xmark & \xmark & \xmark & \xmark & \xmark & \xmark & \checkmark & \xmark & \xmark \\  [2mm]
        
        How does the preprocessed distributions compare against the original data? & \xmark & \xmark & \xmark & \xmark & \xmark & \xmark & \xmark & \xmark & \xmark & \xmark & \xmark & \xmark & \xmark & \xmark & \xmark & \xmark & \xmark & \xmark & \xmark & \xmark & \checkmark & \xmark & \xmark \\ [2mm]
        
        What is the output of each feature selection/engineering method? & \xmark & \xmark & \xmark & \xmark & \xmark & \xmark & \xmark & \xmark & \xmark & \xmark & \xmark & \xmark & \xmark & \xmark & \xmark & \xmark & \xmark & \xmark & \xmark & \xmark & \checkmark & \xmark & \xmark \\  [2mm]
        
        What do the distributions of the new features look like compared to the original features? & \xmark & \xmark & \xmark & \xmark & \xmark & \xmark & \xmark & \xmark & \xmark & \xmark & \xmark & \xmark & \xmark & \xmark & \xmark & \xmark & \xmark & \xmark & \xmark & \xmark & \checkmark & \xmark & \xmark \\ [2mm]
        
        What does the pipeline look like in general? & \xmark & \xmark & \xmark & \checkmark & \xmark & \xmark & \xmark & \xmark & \xmark & \xmark & \xmark & \xmark & \xmark & \xmark & \checkmark & \xmark & \xmark & \xmark & \xmark & \xmark & \xmark & \checkmark & \xmark \\ [2mm]
        
        What does the best pipeline look like? & \xmark & \xmark & \xmark & \xmark & \xmark & \xmark & \xmark & \xmark & \xmark & \xmark & \xmark & \xmark & \xmark & \xmark & \xmark & \checkmark & \xmark & \xmark & \xmark & \checkmark & \xmark & \checkmark & \xmark \\  [2mm]
        
        How does the best pipeline compare to others? & \xmark & \xmark & \xmark & \xmark & \xmark & \xmark & \xmark & \xmark & \xmark & \xmark & \xmark & \xmark & \xmark & \xmark & \xmark & \checkmark & \xmark & \xmark & \xmark & \xmark & \xmark & \xmark & \xmark \\  
        \bottomrule
    \end{tabular*}
    %}
    %}\
    %}
    %\end{adjustbox}
    
\end{sidewaystable}

\begin{sidewaystable}[h]
    \centering
    \tiny
    \caption{Table that presents whether each question from  HXAI Agent's information aggregation is answered by software. \label{unified-expl-tbl}}

    \renewcommand{\arraystretch}{1} % Adjust row spacing
    \setlength{\tabcolsep}{1pt}       % Adjust column spacing
    %\scalebox{0.7}{ % Scale down by 80%
    %\begin{adjustbox}{max width=0.5\textwidth}
    %\resizebox{\textwidth}{!}{ % Scales to fit within text width
    %\adjustbox{valign=c, scale=0.6}{
    \begin{tabular*}{\textwidth}{@{\extracolsep\fill}l*{23}{c}}
    \toprule
        \multirow{2}{*}{\textbf{Question}} 
         & \multicolumn{10}{c}{\textbf{Open XAI}} 
         & \multicolumn{4}{c}{\textbf{Comm. XAI}} 
         & \multicolumn{5}{c}{\textbf{Open AutoML}} 
         & \multicolumn{4}{c}{\textbf{Comm. AutoML}} \\
        \cmidrule(lr){2-11} \cmidrule(lr){12-15} \cmidrule(lr){16-20} \cmidrule(lr){21-24}
         & \rotatebox{90}{Intel-XAI-Tools} 
         & \rotatebox{90}{EthicalML's XAI} 
         & \rotatebox{90}{OmniXAI} 
         & \rotatebox{90}{Explainer} 
         & \rotatebox{90}{Responsible AI Toolbox} 
         & \rotatebox{90}{Pyreal} 
         & \rotatebox{90}{AI Explainability 360} 
         & \rotatebox{90}{Alibi} 
         & \rotatebox{90}{Captum} 
         & \rotatebox{90}{Dalex} 
         & \rotatebox{90}{ArizeAI} 
         & \rotatebox{90}{CensiousAI} 
         & \rotatebox{90}{WhyLabs} 
         & \rotatebox{90}{FiddlerAI} 
         & \rotatebox{90}{MLJAR} 
         & \rotatebox{90}{AutoSklearn} 
         & \rotatebox{90}{H20 Automl} 
         & \rotatebox{90}{FLAML} 
         & \rotatebox{90}{AutoGluon} 
         & \rotatebox{90}{DriverlessAI} 
         & \rotatebox{90}{DataRobot} 
         & \rotatebox{90}{JADBio} 
         & \rotatebox{90}{Amazon Sagemaker AutoPilot} \\
        \midrule
        Is the data imbalance handled by the pipeline? & \xmark & \xmark & \xmark & \xmark & \xmark & \xmark & \xmark & \xmark & \xmark & \xmark & \xmark & \xmark & \xmark & \xmark & \xmark & \xmark & \xmark & \xmark & \xmark & \xmark & \xmark & \xmark & \xmark \\  [2mm]
        
        How can the imbalance be handled by the pipeline? & \xmark & \xmark & \xmark & \xmark & \xmark & \xmark & \xmark & \xmark & \xmark & \xmark & \xmark & \xmark & \xmark & \xmark & \xmark & \xmark & \xmark & \xmark & \xmark & \xmark & \xmark & \xmark & \xmark \\  [2mm]
        
        Is sample collection needed for subpopulations? & \xmark & \xmark & \xmark & \xmark & \xmark & \xmark & \xmark & \xmark & \xmark & \xmark & \xmark & \xmark & \xmark & \xmark & \xmark & \xmark & \xmark & \xmark & \xmark & \xmark & \xmark & \xmark & \xmark \\  [2mm]
        
        Is the missing data handled by the pipeline? & \xmark & \xmark & \xmark & \xmark & \xmark & \xmark & \xmark & \xmark & \xmark & \xmark & \xmark & \xmark & \xmark & \xmark & \xmark & \xmark & \xmark & \xmark & \xmark & \xmark & \xmark & \xmark & \xmark \\  [2mm]
        
        %Are missing values predictive of the target? & \xmark & \xmark & \xmark & \xmark & \xmark & \xmark & \xmark & \xmark & \xmark & \xmark & \xmark & \xmark & \xmark & \xmark & \xmark & \xmark & \xmark & \xmark & \xmark & \xmark & \xmark & \xmark & \xmark \\  [2mm]
        
        How can missing data be fixed? & \xmark & \xmark & \xmark & \xmark & \xmark & \xmark & \xmark & \xmark & \xmark & \xmark & \xmark & \xmark & \xmark & \xmark & \xmark & \xmark & \xmark & \xmark & \xmark & \xmark & \xmark & \xmark & \xmark \\  [2mm]
        
        Are outliers handled by the pipeline? & \xmark & \xmark & \xmark & \xmark & \xmark & \xmark & \xmark & \xmark & \xmark & \xmark & \xmark & \xmark & \xmark & \xmark & \xmark & \xmark & \xmark & \xmark & \xmark & \xmark & \xmark & \xmark & \xmark \\  [2mm]
        
        How can outliers be handled by the pipeline? & \xmark & \xmark & \xmark & \xmark & \xmark & \xmark & \xmark & \xmark & \xmark & \xmark & \xmark & \xmark & \xmark & \xmark & \xmark & \xmark & \xmark & \xmark & \xmark & \xmark & \xmark & \xmark & \xmark \\  [2mm]
        
        Are mislabeled data handled by the pipeline? & \xmark & \xmark & \xmark & \xmark & \xmark & \xmark & \xmark & \xmark & \xmark & \xmark & \xmark & \xmark & \xmark & \xmark & \xmark & \xmark & \xmark & \xmark & \xmark & \xmark & \xmark & \xmark & \xmark \\  [2mm]
        
        How can mislabeled data be handled? & \xmark & \xmark & \xmark & \xmark & \xmark & \xmark & \xmark & \xmark & \xmark & \xmark & \xmark & \xmark & \xmark & \xmark & \xmark & \xmark & \xmark & \xmark & \xmark & \xmark & \xmark & \xmark & \xmark \\  [2mm]
        
        Are there redundant or highly correlated features beneficial? & \xmark & \xmark & \xmark & \xmark & \xmark & \xmark & \xmark & \xmark & \xmark & \xmark & \xmark & \xmark & \xmark & \xmark & \xmark & \xmark & \xmark & \xmark & \xmark & \xmark & \xmark & \xmark & \xmark \\  [2mm]
        
        Are redundant features handled by the pipeline? & \xmark & \xmark & \xmark & \xmark & \xmark & \xmark & \xmark & \xmark & \xmark & \xmark & \xmark & \xmark & \xmark & \xmark & \xmark & \xmark & \xmark & \xmark & \xmark & \xmark & \xmark & \xmark & \xmark \\  [2mm]
        
        Do additional features need to be collected? & \xmark & \xmark & \xmark & \xmark & \xmark & \xmark & \xmark & \xmark & \xmark & \xmark & \xmark & \xmark & \xmark & \xmark & \xmark & \xmark & \xmark & \xmark & \xmark & \xmark & \xmark & \xmark & \xmark \\  [2mm]
        
        What are points of failure of the model? & \xmark & \xmark & \xmark & \xmark & \xmark & \xmark & \xmark & \xmark & \xmark & \xmark & \xmark & \xmark & \xmark & \xmark & \xmark & \xmark & \xmark & \xmark & \xmark & \xmark & \xmark & \xmark & \xmark \\  [2mm]
        
        Are fairness concerns handled by the pipeline? & \xmark & \xmark & \xmark & \xmark & \xmark & \xmark & \xmark & \xmark & \xmark & \xmark & \xmark & \xmark & \xmark & \xmark & \xmark & \xmark & \xmark & \xmark & \xmark & \xmark & \xmark & \xmark & \xmark \\  [2mm]
        
        How can fairness problems be mitigated? & \xmark & \xmark & \xmark & \xmark & \xmark & \xmark & \xmark & \xmark & \xmark & \xmark & \xmark & \xmark & \xmark & \xmark & \xmark & \xmark & \xmark & \xmark & \xmark & \xmark & \xmark & \xmark & \xmark \\  [2mm]
        
        Can the objective function be refined to improve performance? & \xmark & \xmark & \xmark & \xmark & \xmark & \xmark & \xmark & \xmark & \xmark & \xmark & \xmark & \xmark & \xmark & \xmark & \xmark & \xmark & \xmark & \xmark & \xmark & \xmark & \xmark & \xmark & \xmark \\  [2mm]
        
        Is the current metric appropriate? & \xmark & \xmark & \xmark & \xmark & \xmark & \xmark & \xmark & \xmark & \xmark & \xmark & \xmark & \xmark & \xmark & \xmark & \xmark & \xmark & \xmark & \xmark & \xmark & \xmark & \xmark & \xmark & \xmark \\  [2mm]
        
        Is the estimation protocol appropriate? & \xmark & \xmark & \xmark & \xmark & \xmark & \xmark & \xmark & \xmark & \xmark & \xmark & \xmark & \xmark & \xmark & \xmark & \xmark & \xmark & \xmark & \xmark & \xmark & \xmark & \xmark & \xmark & \xmark \\  [2mm]
        
        Is the model overfitting? & \xmark & \xmark & \xmark & \xmark & \xmark & \xmark & \xmark & \xmark & \xmark & \xmark & \xmark & \xmark & \xmark & \xmark & \xmark & \xmark & \xmark & \xmark & \xmark & \xmark & \xmark & \xmark & \xmark \\  [2mm]
        
        %Which features have a negative impact the model’s performance? & \xmark & \xmark & \xmark & \xmark & \xmark & \xmark & \xmark & \xmark & \xmark & \xmark & \xmark & \xmark & \xmark & \xmark & \xmark & \xmark & \xmark & \xmark & \xmark & \xmark & \xmark & \xmark & \xmark \\  [2mm]
        
        %Which samples have a negative impact the model’s performance? & \xmark & \xmark & \xmark & \xmark & \xmark & \xmark & \xmark & \xmark & \xmark & \xmark & \xmark & \xmark & \xmark & \xmark & \xmark & \xmark & \xmark & \xmark & \xmark & \xmark & \xmark & \xmark & \xmark \\  [2mm]
        
        Are feature transformations beneficial? & \xmark & \xmark & \xmark & \xmark & \xmark & \xmark & \xmark & \xmark & \xmark & \xmark & \xmark & \xmark & \xmark & \xmark & \xmark & \xmark & \xmark & \xmark & \xmark & \xmark & \xmark & \xmark & \xmark \\  [2mm]
        
        Can Feature Selection simplify the model without sacrificing accuracy? & \xmark & \xmark & \xmark & \xmark & \xmark & \xmark & \xmark & \xmark & \xmark & \xmark & \xmark & \xmark & \xmark & \xmark & \xmark & \xmark & \xmark & \xmark & \xmark & \xmark & \xmark & \xmark & \xmark \\  [2mm]
        
        Which feature engineering techniques can improve the performance? & \xmark & \xmark & \xmark & \xmark & \xmark & \xmark & \xmark & \xmark & \xmark & \xmark & \xmark & \xmark & \xmark & \xmark & \xmark & \xmark & \xmark & \xmark & \xmark & \xmark & \xmark & \xmark & \xmark \\  [2mm]
        
        Does search space need refinement? & \xmark & \xmark & \xmark & \xmark & \xmark & \xmark & \xmark & \xmark & \xmark & \xmark & \xmark & \xmark & \xmark & \xmark & \xmark & \xmark & \xmark & \xmark & \xmark & \xmark & \xmark & \xmark & \xmark \\  [2mm]
        
        Do more predictive models needed? & \xmark & \xmark & \xmark & \xmark & \xmark & \xmark & \xmark & \xmark & \xmark & \xmark & \xmark & \xmark & \xmark & \xmark & \xmark & \xmark & \xmark & \xmark & \xmark & \xmark & \xmark & \xmark & \xmark \\  [2mm]
        
        Are important hyper-parameters missing? & \xmark & \xmark & \xmark & \xmark & \xmark & \xmark & \xmark & \xmark & \xmark & \xmark & \xmark & \xmark & \xmark & \xmark & \xmark & \xmark & \xmark & \xmark & \xmark & \xmark & \xmark & \xmark & \xmark \\  [2mm]
        
        Are hyper-parameters wrongly optimized? & \xmark & \xmark & \xmark & \xmark & \xmark & \xmark & \xmark & \xmark & \xmark & \xmark & \xmark & \xmark & \xmark & \xmark & \xmark & \xmark & \xmark & \xmark & \xmark & \xmark & \xmark & \xmark & \xmark \\  [2mm]
        
        How can the optimization algorithm improve?  & \xmark & \xmark & \xmark & \xmark & \xmark & \xmark & \xmark & \xmark & \xmark & \xmark & \xmark & \xmark & \xmark & \xmark & \xmark & \xmark & \xmark & \xmark & \xmark & \xmark & \xmark & \xmark & \xmark \\ 
        \bottomrule
    \end{tabular*}
    %}
    %}\
    %}
    %\end{adjustbox}
\end{sidewaystable}

\clearpage
\bibliography{sn-bibliography}% common bib file

\end{document}